\documentclass[lettersize,journal]{IEEEtran}
\usepackage{amsmath,amsfonts}
\usepackage{algorithmic}
\usepackage{algorithm}
\usepackage{array}
\usepackage[caption=false,font=normalsize,labelfont=sf,textfont=sf]{subfig}
\usepackage{textcomp}
\usepackage{stfloats}
\usepackage{url}
\usepackage{verbatim}
\usepackage{graphicx}
\usepackage{cite}
\usepackage{booktabs} 
\usepackage{array} 
\usepackage{multirow}

\usepackage{hyperref}
\usepackage{tabularx}

\usepackage[capitalise]{cleveref}

\crefname{figure}{Fig.}{Figs.}
\Crefname{figure}{Fig.}{Figs.}

\crefname{table}{TABLE}{TABLES}
\Crefname{table}{TABLE}{TABLES}

\crefname{section}{Section}{Sections}
\Crefname{section}{Section}{Sections}

\crefname{equation}{Eq.}{Eqs.}
\Crefname{equation}{Equation}{Equations}

\usepackage{amssymb} 
\usepackage{pifont}

\newcommand{\cmark}{\ding{51}} 
\newcommand{\xmark}{\ding{55}} 

\begin{document}

\title{ReSAGE-PAR: Representational Similarity Assessment for Generative Expansion in Pedestrian Attribute Recognition}

\author{Pablo Ayuso-Albizu, Pablo Carballeira, Juan C. SanMiguel, Paula Moral, \textit{Universidad Autónoma de Madrid, Madrid, Spain,
pablo.ayuso@estudiante.uam.es, juancarlos.sanmiguel@uam.es, pablo.carballeira@uam.es, paula.moral@uam.es}
\thanks{This work has been submitted to the IEEE for possible publication. Copyright may be transferred without notice, after which this version may no longer be accessible. This work has been partially supported by the Regional Government of Madrid of Spain (grant TEC 2024/COM-322).}
}

\markboth{}%
{Ayuso-Albizu \MakeLowercase{\textit{et al.}}: ReSAGE-PAR}


\IEEEpubidadjcol
\maketitle

\begin{abstract}

To address the limited diversity and data scarcity in Pedestrian Attribute Recognition (PAR), we explore image synthesis using diffusion models guided by attribute-based prompts. While this enables the controlled generation of pedestrian images, it faces two critical challenges: (i) the domain gap between high-quality pre-training data and low-resolution, non-standard surveillance crops, and (ii) the need for reliable attribute verification to prevent generative hallucinations. In this paper, we introduce a robust generate-score-autolabel pipeline called ReSAGE-PAR (REpresentational Similarity Assessment for Generative Expansion in PAR) that bridges this domain gap and enables scalable, high-fidelity dataset expansion. First, we adapt pre-trained diffusion models to native PAR resolutions using a tailored LoRA-based Image-to-Image approach. Second, we extract vision-language alignment scores between the generated images and their conditioning prompts, utilizing a comprehensive prompting strategy that includes label-consistent and inconsistent complements. Finally, we formulate a Bayesian classifier that converts these continuous scores into reliable binary pseudo-labels. Extensive evaluations demonstrate the effectiveness of ReSAGE-PAR in preserving spatial priors and verifying attributes. When integrated into PAR training, ReSAGE-PAR consistently yields significant improvements—achieving gains of up to 8.7\% on standard backbones and pushing state-of-the-art frameworks to new performance levels. This proves its value as an architecture-agnostic solution for scalable PAR enhancement. The complete codebase for ReSAGE-PAR is publicly available at \url{http://www-vpu.eps.uam.es/publications/ReSAGE-PAR}.

\end{abstract}


\section{Introduction}
\label{sec:introduction}

Pedestrian Attribute Recognition (PAR) is a critical component in surveillance and re-identification systems, aiming to predict semantic properties like clothing, accessories, or viewpoint from a single image. However, this task remains inherently challenging because attributes are typically small, occluded, long-tailed, and captured under diverse illumination conditions~\cite{human_attribute_survey}. While benchmarks like \textit{PETA}~\cite{peta}, \textit{RAP}~\cite{RAPv1}, and \textit{RAPv2}~\cite{RAPv2} provide essential data, their reliance on low-resolution, noisy surveillance frames intrinsically imposes a severe barrier. Furthermore, existing data splits suffer from identity overlap, motivating zero-shot variants like \textit{PA100K}~\cite{pa100k}, \textit{PETAzs}, and \textit{RAPzs}~\cite{rethinking}, where disjoint identities make generalization significantly harder (e.g., mA drops of 9--17\%~\cite{rethinking}).

Collectively, characteristics such as low resolution, occlusion, and viewpoint variations severely hinder standard recognition, amplifying the inherent problem of data scarcity in surveillance datasets. To tackle this scarcity, generative models are increasingly employed to synthesize additional pedestrian samples. While early approaches relied on Generative Adversarial Networks (GANs) \cite{dgnet2019, spgan2018}, their limited diversity and lack of photorealism \cite{diffusionbeatgans} have paved the way for Text-to-Image Diffusion Models as a superior alternative \cite{method_dafusion}. Yet, naively applying these powerful diffusion models to PAR remains extremely difficult. The drastic visual gap between the high-quality web images these models were originally trained on and the noisy surveillance domain introduces two major unmet challenges largely ignored by previous literature.

The first major challenge arises from the profound domain gap existing between the high-quality images used to pre-train foundation models and the specific constraints of surveillance data. Foundation models like Stable Diffusion are pretrained on high-quality images (e.g., LAION \cite{laion}), which differs drastically from the low-resolution, blurred, and oddly-angled surveillance frames in PAR datasets. Simply prompting a standard model yields realistic portraits that fail to match the PAR target distribution.

The second obstacle involves the inherent uncertainty of generative labeling, where relying exclusively on text prompts proves insufficient due to the tendency of diffusion models to hallucinate or omit specific attributes. Generating images is not enough; training a supervised PAR model requires accurate labels. Relying solely on the generation prompt is risky because diffusion models can generate unprompted features or ignore parts of complex prompts (e.g., generating a ``backpack'' when requested, but missing ``glasses''). Manual annotation is unscalable for synthetic expansion. This creates a need for an automated verification mechanism.

In this paper, we propose ReSAGE-PAR, a novel generate-score-autolabel framework tailored for PAR. To address the aforementioned challenges, we structure ReSAGE-PAR by explicitly separating the dataset-specific generative adaptation from the score-driven label verification. First, we leverage Low-Rank Adaptation (LoRA) to bridge the visual gap between generic text-to-image models and the surveillance domain. By injecting trainable rank-decomposition matrices into the frozen backbone, LoRA efficiently captures the specific 'surveillance style' and resolution of each PAR dataset. Second, we introduce a Bayesian verification stage that converts continuous vision-language alignment scores into reliable pseudo-labels. This mechanism filters generative hallucinations by explicitly evaluating the semantic consistency between the synthetic image and its conditioning prompt. We demonstrate that ReSAGE-PAR serves as a model-agnostic data augmentation module, delivering consistent improvements across diverse backbones and state-of-the-art (SOTA) frameworks.

Our main contributions are summarized as follows:
\begin{itemize}
    \item \textit{Domain-Specific Generative Adaptation:} We present the first application of LoRA for Image-to-Image (img2img) diffusion in the PAR domain. Unlike previous generative baselines, our approach effectively bridges the dataset domain gap by adapting to the non-standard resolutions and noise profiles of surveillance while efficiently preserving the rich spatial priors of the original model.
  
  \item \textit{Score-Driven Autolabeling and Metric Evaluation:} We provide the first comprehensive evaluation of multiple Vision-Language metrics (e.g., CLIP, BLIP, ImageReward) for fine-grained pedestrian attribute verification. We identify BLIPScore as the most robust signal for detecting prompt-image misalignment in low-resolution contexts.

  \item\textit{Model-Agnostic Performance and Synthetic Scaling:} We introduce a principled Bayesian framework that transforms continuous alignment scores into discrete pseudo-labels. This method enables effective dataset scaling by prioritizing label precision over raw volume, achieving SOTA results through verified synthetic expansion.
\end{itemize}

The rest of this article is structured as follows. \cref{sec:related_work} reviews the related literature in text-image representational similarity metrics, generative augmentation, and automatic annotation. In \cref{sec:method}, we introduce our proposed ReSAGE-PAR method for domain-aware image generation and score-driven autolabeling. \cref{sec:experimental_results} explores the internal stages of our framework, validating the intrinsic accuracy and robustness of the Bayesian filter as a standalone autolabeling mechanism. \cref{sec:application_par} evaluates the application of our verified synthetic data to downstream PAR, providing an extensive empirical analysis that demonstrates the superiority of our generative augmentation, its scalability across diverse backbones, and its ability to push SOTA frameworks to new performance levels. Finally, conclusions are discussed in \cref{sec:conclusions}, followed by limitations and future work in \cref{sec:limitations_future}.

\section{Related Work}
\label{sec:related_work}

Our research sits at the intersection of multimodal evaluation, generative data synthesis, and automated supervision. First, we review \textit{Text-Image Representational Similarity Metrics} (\cref{subsec:related_work_promptfidelity}), which serve as the core mechanism for our verification strategy.
Second, we discuss the evolution of \textit{Data Augmentation for Pedestrian-Centric Recognition} (\cref{subsec:related_work_syn_generation}), positioning our diffusion-based approach against classical augmentation and earlier generative methods.
Finally, we examine \textit{Automatic Annotation} strategies (\cref{subsec:related_work_automatic_labels}), highlighting the shift towards scalable, model-assisted labeling to mitigate the scarcity of annotated surveillance data.


\subsection{Text-image Representational similarity metrics}
\label{subsec:related_work_promptfidelity}

Recent research on text-image alignment provides scores to verify whether a prompt is faithfully expressed in an image.
CLIPScore \cite{clipscore} builds on CLIP and measures cosine similarity between jointly trained image/text embeddings. It is widely adopted for filtering synthetic data in recent works \cite{method_dafusion, avss_zs}, despite being biased toward broad topical matches rather than fine details.
BLIPScore \cite{blip} utilizes caption-style pretraining, making it potentially more sensitive to attribute visibility.
Preference-based metrics like ImageReward \cite{imgreward} and HPSv2Score \cite{hpsv2score} rank images based on human aesthetic judgment, capturing plausibility.
More recently, decompositional approaches like VQAScore \cite{vqascore} and DSG \cite{dsg} leverage visual question answering for fine-grained compositional constraints.

Crucially, while these metrics have been benchmarked on general high-quality generic images, their comparative effectiveness for verifying fine-grained attributes in the low-resolution, noisy surveillance domain remains unexplored. Most existing PAR augmentation pipelines naively adopt CLIP without evaluating whether it accurately perceives attributes like "glasses" or "backpacks" in surveillance frames.
In this work, we bridge this gap by systematically evaluating these metrics to identify the most reliable autolabeling signal for the PAR domain.

\subsection{Data augmentation for Pedestrian-Centric Recognition}
\label{subsec:related_work_syn_generation}

Data augmentation is a critical yet challenging necessity in pedestrian-centric recognition, where long-tailed attribute distributions, frequent occlusions, and surveillance-specific domain shifts severely limit generalization \cite{human_attribute_survey}. While traditional pipelines rely heavily on label-preserving transformations (e.g., random erasing, cropping) to improve invariance, these methods are fundamentally limited to the information present in the original pixels and cannot increase diversity along high-level semantic axes. Consequently, diffusion-based generative augmentation represents an emerging and under-explored frontier in PAR. It offers a complementary mechanism to synthesize realistic pedestrians with controllable variation (appearance, pose, accessories), rebalancing scarce concepts beyond what hand-crafted policies can achieve.

\subsubsection{Classical augmentation policies}
\label{subsubsec:clasaugpol}
 
Automatic policy methods such as AutoAugment \cite{autoaug}, RandAugment \cite{randaug}, and TrivialAugment \cite{trivaug} select stronger transform combinations beyond hand-crafted pipelines. 
To simulate occlusions and encourage part-based reasoning, region-dropout strategies like Cutout \cite{cutout_dataaug} and Random Erasing \cite{random_erasing_dataaug} remove random patches during training. 
Mixing-based regularizers (Mixup \cite{mixup}, CutMix \cite{cutmix}) further reduce overfitting by blending samples or regions. 
While effective, these techniques mainly preserve the original semantics and therefore provide limited control over rare compositional attribute combinations, motivating complementary generative augmentation.

\subsubsection{GANs}
\label{subsubsec:gans}
Prior to the rise of diffusion models, GANs were the dominant paradigm for generative augmentation. Seminal works like DG-Net~\cite{dgnet2019} and SPGAN~\cite{spgan2018} utilized adversarial training to disentangle identity from appearance or transfer domain styles, yielding improvements in re-identification. Despite these advances, GAN-based synthesis suffers from inherent training instability and, crucially, lacks fine-grained semantic controllability. Unlike text-driven models, GANs struggle to generate specific, complex attribute combinations on demand, a limitation that paved the way for the flexible conditioning of diffusion models~\cite{diffusionbeatgans}.

\subsubsection{Diffusion-based augmentation}
\label{subsubsec:diffaug}

\paragraph{Diffusion-based synthetic augmentation for pedestrian analysis}
Recent work leverages diffusion models as realistic pedestrian generators to augment training data in pedestrian-centric recognition. MALS \cite{mals_dataset} synthesizes large-scale pedestrian images with text-to-image diffusion and derives attribute vectors from captions, showing that synthetic pedestrians can pre-train retrieval models at competitive levels on real benchmarks, albeit with labels tied to a fixed attribute vocabulary. In a broader context, \textit{DAFUSION}~\cite{method_dafusion} introduced a general framework for fine-tuning diffusion models for effective data augmentation. This strategy was subsequently adapted to the PAR domain by \cite{avss_zs}, who utilized it to generate surveillance-style images conditioned on attributes to improve zero-shot PAR performance. Other data-centric approaches \cite{method_avss_alonso} tackle the domain gap by generating high-resolution synthetic images to maximize attribute fidelity, followed by explicit degradation transformations (e.g., blurring, downsampling) to simulate surveillance quality. Unlike these multi-stage degradation heuristics, ReSAGE-PAR utilizes LoRA to learn and generate directly within the target dataset's native resolution and style. Related evidence from person re-identification further supports diffusion as a controllable data augmenter, where synthesis can preserve identity and vary pose or appearance (e.g., Diffusion-ReID \cite{reid_diffusion_syn} and Pose-dIVE \cite{reid_diffusion_syn_pose}), suggesting diffusion-based augmentation is broadly useful for downstream pedestrian understanding.

\paragraph{Prompting and controllable conditioning for attribute fidelity} beyond generating more samples, diffusion augmentation is only effective if synthetic images faithfully realize fine-grained and compositional attributes. To improve controllability, composable/grounded conditioning enriches prompts with additional structure and spatial cues (e.g., multi-condition adapters and grounded generation) \cite{mou2024t2i} \cite{huang2023composer} \cite{li2023gligen}. Complementary, prompt editing and faithfulness methods manipulate cross-attention to enforce attribute presence and reduce omissions \cite{hertz2022prompt} \cite{chefer2023attend}, while automatic prompt optimization rewrites templates into more attribute-complete prompts without changing semantics \cite{hao2023optimizing}. These directions are compatible with PAR-style synthesis, where structured, dataset-aware templates can improve coverage of clothing, accessories, and actions. 

\paragraph{Dataset/domain adaptation via parameter-efficient fine-tuning} a practical challenge is that off-the-shelf diffusion models are not specialized for surveillance images or dataset-specific attribute taxonomies. Parameter-efficient adaptation \cite{pefts} enables tailoring the generator to the target domain while keeping training lightweight, and can be combined with structured prompts to reduce the surveillance domain gap and increase coverage of rare attribute combinations. In our work, we follow this motivation and adopt dataset-aware diffusion adaptation together with structured prompts designed for PAR.

\subsection{Automatic annotation of synthetic data}
\label{subsec:related_work_automatic_labels}

Autolabeling has been explored across modalities to mitigate data scarcity.
In 3D settings, differentiable rendering of Signed Distance Function (SDF) priors aligns synthetic predictions to real RGB/LiDAR for detector supervision \cite{zakharov2020sdflabel}, while Vision Language Model (VLM)-assisted pipelines reduce human effort in point cloud annotation \cite{vespa2025}. In 2D domains, \textit{AutoLabel} methods learn confidence-aware soft labels for strongly augmented samples \cite{qin2023autolabel}. More recently, the rise of Multimodal LLMs (MLLMs) has shifted the paradigm towards \textit{zero-shot} annotation, where models like LLaVA or SharedGPT-4V generate fine-grained descriptions directly from raw images \cite{sharegpt4v, llava}. Complementary lines study synthetic data generation itself, including controllable diffusion models for distribution-aligned synthesis \cite{difflm2024}, self-improving training loops \cite{sims2024}, and the use of AI-labeled synthetic data for evaluation \cite{autoeval2025}. ReSAGE-PAR differs from these heavy MLLM-based approaches by prioritizing \textit{computational efficiency} alongside accuracy. Instead of relying on resource-intensive Large Models for inference on every sample, we (i) fine-tune diffusion with LoRA to generate \textit{dataset-aware} pedestrian samples, and (ii) assign pseudo-labels using a lightweight ensemble of scoring functions.
This allows for scalable high-throughput data generation while avoiding the latency and compute costs associated with large-scale MLLM deployment.

\section{Method}
\label{sec:method}

\begin{figure*}[!t]
 \centering
\includegraphics[width=\linewidth, height=5.5cm, keepaspectratio=false, trim=16 6 16 6, clip]{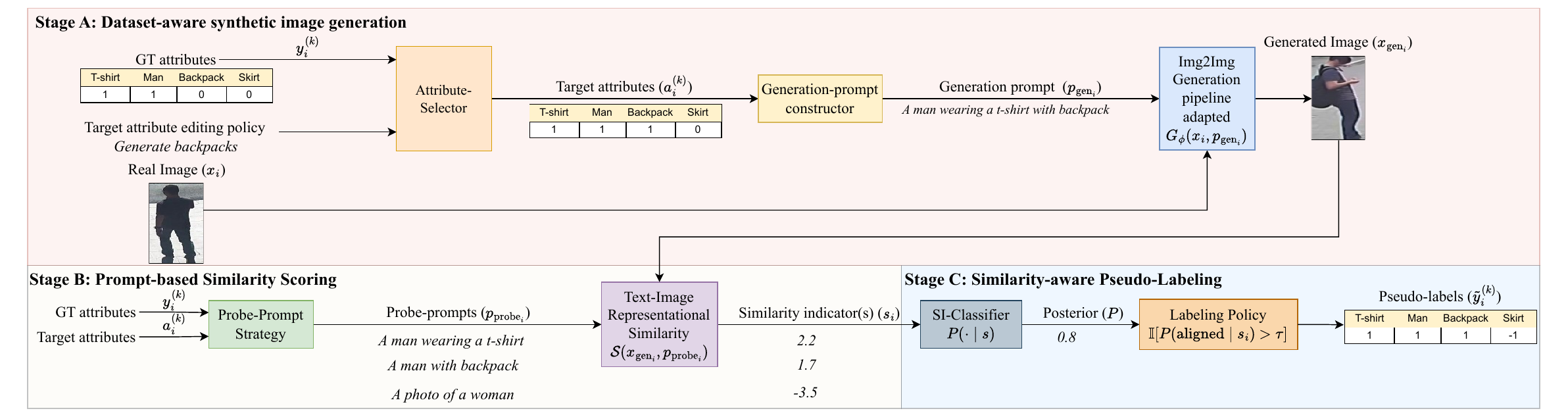}
 \caption{ReSAGE-PAR overview.
\emph{Stage A-dataset-aware synthetic image generation:} Given a real image $x_i$ and a target attribute-editing policy, this stage generates a synthetic image $x_{\text{gen},i}$ that preserves the coarse spatial layout of the original sample while enforcing the presence of target attributes $a_i$.
\emph{Stage B-Prompt-based Similarity Scoring:} Given the target attributes and the real labels $y_i$, this stage constructs probe prompts $p_{\text{probe},i}$ and utilizes a text--image similarity metric $\mathcal{S}$ to produce similarity indicator(s) $s_i$ that quantify the alignment between the synthetic image and the desired attributes. 
\emph{Stage C-Similarity-aware Pseudo-Labeling:} Using the similarity indicators, a Bayesian classifier estimates the posterior probability $P(\cdot|s_i)$ of alignment. Finally, a labeling policy applies a threshold $\tau$ to convert these probabilities into a definitive attribute-level pseudo-label vector $\tilde{y}_i$, masking out uncertain or non-target attributes.}

 \label{fig:sec_method_overview}
\end{figure*}

The general objective of ReSAGE-PAR is to generate synthetic images similar to real data and automatically label them for PAR. To achieve this, the framework is divided into three stages as shown in \cref{fig:sec_method_overview}: (a) dataset-aware synthetic image generation, (b) prompt-based similarity scoring, and (c) similarity-aware pseudo-labeling. We name our method ReSAGE-PAR (REpresentational Similarity Assessment for Generative Expansion in PAR).

\begin{figure}[h]
 
\includegraphics[width=0.9\linewidth, trim=9cm 4cm 9cm 4cm, clip]{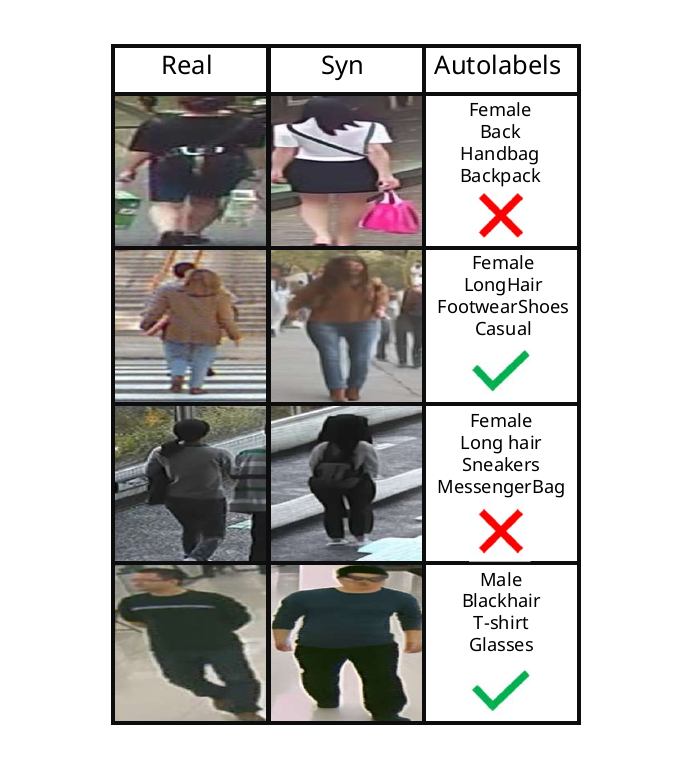}
\centering
 \caption{Qualitative results of ReSAGE-PAR. Each triplet shows a real image (Real), its dataset-aware synthetic counterpart (Syn) generated with Stable Diffusion + LoRA at the dataset resolution, and the prompt attributes (Autolabels). Green terms are confirmed by our score-based autolabeler; red terms are flagged as mismatched. The \cmark / \xmark summarizes whether the whole attribute set is consistent with the image.}

 \label{fig:figureabstract}
\end{figure}

\subsection{Stage A: Dataset-aware synthetic image generation}
\label{subsec:method_stage_a}

Let $\mathcal{D} = \{(x_i, y_i)\}_{i=1}^{N}$ be the dataset of PAR real images, where $x_i \in \mathcal{X}$ denotes a real image and $y_i \in \{0, 1\}^K$ is its associated ground-truth label vector with $K$ attributes. This first stage aims to generate synthetic samples from the target dataset by conditioning the generative process on a set of target attributes, while using the original sample $x_i$ as a structural prior to maintain the coarse spatial layout. We decompose this stage into three modules:

\paragraph{Attribute Selector} 
This module guides how the original image is semantically edited to generate the synthetic data. This requires defining an editing policy that determines the specific visual characteristics we want to add or modify (e.g., adding a "backpack"). Specifically, the selector takes the original ground-truth labels $y_i$ and applies this policy to yield the \textit{target attribute vector} $a_i \in \{0,1\}^K$, which explicitly represents the final set of attributes requested for the new image. In this work, as a first approach, we adopt an identity policy where no modifications are introduced, meaning the target attributes strictly match the original ground-truth, i.e., $a_i = y_i$. While the semantic labels remain constant, the diffusion prior inherently introduces critical diversity in pedestrian pose, background, and illumination that traditional pixel-level augmentations cannot achieve. A key strength of ReSAGE-PAR is the high structural diversity maintained across all domains. As qualitatively demonstrated in \cref{fig:figureabstract}, the generative prior introduces massive non-trivial spatial variations beyond simple pixel perturbations, highlighting how our LoRA-based adaptation effectively overrides high-definition priors to generate diverse surveillance-style samples.

\paragraph{Generation-Prompt Constructor}
This module defines as a mapping $\mathcal{T}: \{0,1\}^K \rightarrow \mathcal{P}$ to translate the target attribute vector into a \textit{generation prompt} $p_{\text{gen}_i} = \mathcal{T}(a_i)$. 
In this work, we implement $\mathcal{T}$ as a \textit{deterministic, template-based function} to ensure strict semantic adherence to the target attributes. 
The exact templates are summarized in \cref{tab:prompt_templates_compact}. Further implementation details and qualitative examples are detailed in the accompanying supplementary material.

\paragraph{Img2Img Generation Pipeline} 
To synthesize new pedestrian samples, this module leverages the real image $x_i$ as an initial structural anchor, preserving the overall layout, background, and pedestrian pose. Guided by the prompt $p_{\text{gen}_i}$, a conditional generator $G_{\phi}$ iteratively denoises the latent representation to sample a synthetic image $x_{\text{gen}_i}$ that reflects the target attributes:
\begin{equation}
    x_{\text{gen}_i} = G_{\phi}(x_i, p_{\text{gen}_i}) = G_{\phi}(x_i, \mathcal{T}(a_i)).
\end{equation}

To instantiate this conditional generator $G_{\phi}$, we combine the strong natural image priors of a base text-to-image diffusion model (e.g., Stable Diffusion) with a domain-specific adaptation via LoRA. While the base model provides the fundamental synthesis capabilities and text comprehension, standard diffusion models struggle with the unique characteristics of surveillance footage (such as low resolution, motion blur, and specific lighting). Therefore, the LoRA parameters within $G_{\phi}$ are fine-tuned exclusively on real PAR images. In this work, we apply LoRA for each dataset adaptation to effectively bridge the visual domain gap, as detailed in \cref{subsec:experimental_lora}.

\begin{table}[!t]
\centering
\caption{Dataset-specific prompt templates. Double underscores (\_\_) indicate attributes and prefixes emitted only if present in the target vector. The \texttt{start} and \texttt{gender} tokens are foundational. Abbrev.: \texttt{hs}=hair style, \texttt{ub}=upper body, \texttt{lb}=lower body.}
\label{tab:prompt_templates_compact}
\scriptsize
\setlength{\tabcolsep}{3pt}
\renewcommand{\arraystretch}{1.3} 
\begin{tabular}{l p{0.68\linewidth}}
\toprule
\textbf{Dataset} & \textbf{Template sketch}  \\
\midrule
RAPv1 / v2 / zs & \texttt{start gender \_\_with hs\_\_ \_\_is action\_\_ \_\_att\_\_ \_\_wearing ub\_\_ \_\_lb\_\_ \_\_shoes\_\_ \_\_attachment\_\_} \\
Example of RAPv1 & there is a man with black hair sinlge shoulder bag other wearing t-shirt jacket jeans \\

\midrule
PETA / PETAzs & \texttt{start gender \_\_with hair-color\_\_ \_\_with hair\_\_ \_\_carrying item\_\_ \_\_with accessory\_\_ \_\_wearing ub-color ub\_\_ \_\_lb-color lb\_\_ \_\_shoe-color shoe\_\_} \\

Example of PETA & a man with black short hair carrying backpack with nothing wearing black casual grey casual trousers grey shoes \\

\midrule
PA100k & \texttt{start gender \_\_from view\_\_ \_\_with attr\_\_ \_\_wearing upper-wear/a upper-type\_\_ \_\_lower-wear/a lower-type\_\_ \_\_boots\_\_} \\

Example of PA100k & there is a woman from front wearing short sleeve a plaid skirt or dress \\

\bottomrule
\end{tabular}
\end{table}

\subsection{Stage B: Prompt-based Similarity Scoring}
\label{subsec:method_stage_b}

This stage quantifies the alignment between the generated image $x_{\text{gen}_i}$ and its target attributes $a_i$.

\paragraph{Probe-Prompt Builder}
This module constructs a set of $N$ probe prompts $\mathcal{P}_{\text{probe}_i} = \{p_{\text{probe}_i}^{(1)}, \dots, p_{\text{probe}_i}^{(j)}\} \subset \mathcal{P}$ designed to verify the visual content of the synthetic image.

\paragraph{Text--Image Similarity Computing.}
The core objective of this module is to explicitly quantify the semantic alignment between the synthesized pedestrian image and its corresponding textual description. This step is crucial for detecting potential prompt-image misalignments, ensuring that the generative model successfully rendered the requested attributes. To achieve this, let $\mathcal{S}: \mathcal{X} \times \mathcal{P} \rightarrow \mathbb{R}$ be a Text-Image Representational Similarity metric. We compute the alignment score $s_{ij}$ for the synthetic data-prompt as:
\begin{equation}
    s_{ij} = \mathcal{S}(x_{\text{gen}_i}, p_{\text{probe}_i}^{(j)}).
\end{equation}
Here, higher values of $s_{ij}$ indicate stronger semantic consistency between the generated image $x_{\text{gen}_i}$ and its generation prompt $p_{\text{probe}_i}^{(j)}$. The resulting alignment scores $s_{ij}$ serve as a quantitative measures of prompt fidelity and act as the primary criteria for evaluating the quality of synthetic samples in the subsequent autolabeling stage. Based on our comprehensive analysis in \cref{subsec:txt_img_score_analysis}, we adopt BLIPScore as our metric $\mathcal{S}$.

\subsection{Stage C: Similarity-aware Pseudo-Labeling}
\label{subsec:method_stage_c}

This stage aims to transform the continuous similarity indicators $s_{ij}$ into reliable supervision signals, mitigating the impact of potential generative hallucinations. To ensure that only semantically consistent samples guide the learning process, we employ a labeling policy that maps these scores into an attribute-level pseudo-label vector $\tilde{y}_i = [\tilde{y}_{i,1}, \dots, \tilde{y}_{i,K}]$ , where $K$ denotes the total number of attributes. While the framework conceptually supports soft-label probabilities in the interval $[0, 1]$, we discretize the output to explicitly distinguish between certain and ambiguous predictions. Specifically, each element $\tilde{y}_{i,k} \in \{0, 1, -1\}$ represents the confident presence ($1$), absence ($0$), or an ignore flag ($-1$) for uncertain attributes. This formulation may allow for a masked Binary Cross-Entropy (BCE) loss, effectively neutralizing the influence of mismatched attributes on gradient updates and prioritizing label precision over raw data volume.

\paragraph{Labeling Policy}
This policy converts the probabilistic output into hard pseudo-labels. The final pseudo-label vector $\tilde{y}_i$ is assigned element-wise as:
\begin{equation}
    \tilde{y}_i^{(k)} = 
    \begin{cases} 
      \hat{d}_i^{(k)} & \text{if } y_i^{(k)} = 1 \text{ (if attribute $k \in \text{probe}$)}, \\
      -1 & \text{if } y_i^{(k)} = 0 \text{ (if attribute $k \notin \text{probe}$)}.
    \end{cases}
\end{equation}

This formulation explicitly distinguishes between attributes evaluated in the probe-prompt and those that were not. Since the visual state of unprompted attributes remains unverified, they are assigned a value of $-1$ to act as an ignore flag. Consequently, only the target attributes actively confirmed by the Similarity-Indicator Classifier contribute to the downstream training loss.

Let $\hat{d}_i^{(k)} = \mathbb{I}[P(\text{aligned} \mid s_{ij}) > \tau]$ be the binary decision indicating whether the generated image successfully reflects the target attributes, where $\tau \in [0,1]$ is a decision threshold. The parameter $\tau$ dictates the restrictiveness of the acceptance policy. While thresholds approaching $1$ enforce strict attribute fidelity at the cost of discarding many viable samples, and values near $0$ act as a highly permissive filter that admits generative noise, we adopt a neutral decision boundary of $\tau = 0.5$ to optimally balance semantic accuracy with synthetic data yield.

\paragraph{Similarity-Indicator Classifier}
We propose a Bayesian classifier that estimates the posterior probability of the synthetic image $x_{\text{gen}_i}$ being aligned with the probe prompt $p_{\text{probe}_i}^{(1)}$. This posterior probability of alignment is defined using the function $\sigma(\cdot)$ applied to the Log-Likelihood Ratio (LLR):
\begin{equation}
    P(\text{aligned} \mid s_{ij}) = \sigma(\mathrm{LLR}(s_{ij})) = \frac{1}{1 + e^{-\mathrm{LLR}(s_{ij})}}.
    \label{eq:post}
\end{equation}

Since raw output probabilities lack an absolute scale, we must explicitly discern whether a given score represents true visual alignment with the image or visual misalignment. To model this LLR and establish this discriminative baseline, we instantiate Stage B to generate a \textit{contrastive pair} ($N=2$) for calibration. Specifically, for each positive prompt $p_{probe_i}^{(j)}$, we define the \textit{positive probe} $p_{\text{pos}_i}^{(1)}$ by setting the probe prompt directly to the generation prompt used in Stage A, such that $p_{\text{pos}_i}^{(1)} = p_{\text{probe}_i}^{(1)} = p_{\text{gen}_i}^{(1)}$. Similarly, we construct a \textit{negative probe} $p_{\text{neg}_i^{(1)}}$ by flipping a fraction $\rho \in (0, 1]$ of the active attributes in $p_{\text{probe}_i}^{(1)}$ to their attribute-family complementaries defined by the dataset (e.g., swapping "long hair" for "short hair"). In our primary contrastive setup, we apply full complementation ($\rho = 1$).

Using the similarity scores from both the positive prompt ($s_{ij}$) and the negative complemented prompt ($\bar{s}_{ij} = \mathcal{S}(x_{\text{gen}_i}, p_{\text{neg}_i}^{(j)})$), the LLR is formulated based on their respective distributions, $P(s \mid p_{\text{pos}}^{(j)})$ and $P(s \mid p_{\text{neg}}^{(j)})$:
\begin{equation}
    \mathrm{LLR}(s_{ij}) = \ln \frac{P(s_{ij} \mid p_{\text{pos}}^{(j)})}{P(s_{ij} \mid p_{\text{neg}}^{(j)})} + \ln \frac{\pi_{\text{pos}}}{1-\pi_{\text{pos}}},
    \label{eq:llr}
\end{equation}
where $\pi_{\text{pos}}$ represents the prior probability of alignment. Given our contrastive design which yields an equal number of positive and negative probes and the absence of empirical knowledge regarding the generative model's exact success rate, we naturally adopt an uninformative balanced prior ($\pi_{\text{pos}} = 0.5$). Consequently, the log-prior term vanishes ($\ln 1 = 0$), ensuring that the LLR is driven purely by the observational visual-semantic likelihoods.

\section{Experimental results}
\label{sec:experimental_results}

This section focuses on the empirical validation of the internal components and intermediate stages of the ReSAGE-PAR framework. First, we investigate the impact of the LoRA rank during generative fine-tuning to determine the optimal configuration for each dataset (\cref{subsec:experimental_lora}). Second, we evaluate various text-image representational similarity metrics (\cref{subsec:txt_img_score_analysis}) to identify the most reliable scoring function. Finally, we conduct an in-depth analysis of the Bayesian autolabeling pipeline (\cref{subsec:end2end_labeling}), validating its intrinsic accuracy, threshold robustness, and capacity to effectively filter semantic noise prior to any downstream application.

\subsection{Implementation Details}

\paragraph{Datasets} We evaluate ReSAGE-PAR on four widely used PAR benchmarks: PETA~\cite{peta}, PA100K~\cite{pa100k}, and RAP (v1~\cite{RAPv1} and v2~\cite{RAPv2}). Specifically, PETA consists of 19,000 images with 61 binary attributes, while PA100K—the largest in terms of volume—contains 100,000 images captured from 26 outdoor surveillance cameras. The RAP series provides high-resolution samples (31,268 for v1 and 84,928 for v2) with fine-grained annotations. Furthermore, we evaluate generalization performance using the zero-shot variants (PETAzs, RAPzs) introduced in~\cite{rethinking}, which test the models' ability to recognize unseen attribute combinations in diverse surveillance environments.

\paragraph{Implementation and Base Code} For the generative augmentation phase, we utilize the pre-trained weights of Stable Diffusion v2.1~\cite{stable_diffusion}. A key technical contribution of our framework is the custom adaptation of the LoRA framework~\cite{lora} for img2img tasks, a configuration not natively supported in standard LoRA-based PAR generation scripts. This modification, implemented via the Hugging Face diffusers library~\cite{diffusers}, allows for precise domain steering while strictly preserving the spatial and structural priors of the original pedestrian crops. To compute the alignment scores ($s$), we integrate a comprehensive suite of vision-language models, including BLIP~\cite{blip}, CLIPScore~\cite{clipscore}, ImageReward~\cite{imgreward}, and HPSv2~\cite{hpsv2score}. For the downstream PAR evaluation, we adapted the standard experimental framework to natively incorporate our synthetically generated data. Specifically, we modified the data loading pipeline to ingest hybrid batches of real and synthetic samples, and we adapted the objective function by introducing a masked loss that explicitly ignores unannotated or unprompted attributes (labeled as $-1$) during backpropagation. To demonstrate the broad compatibility of our approach, we integrated this adapted framework into several state-of-the-art architectures, including the Rethinking baseline~\cite{rethinking}, PromptPAR~\cite{promptPAR}, and SequencePAR~\cite{sequencepar}.

\paragraph{Training Hyperparameters} 

For the generative fine-tuning phase (Stage A), we optimize the diffusion model using LoRA via the \texttt{diffusers} library, applying a constant learning rate of $1\times10^{-4}$ and a batch size of 16. To prevent the degradation of spatial features caused by standard square cropping, we configure the training to strictly respect the aspect ratios of pedestrian crops by fixing the resolution to $256 \times 192$ pixels ($H \times W$), a standard resizing dimension in PAR~\cite{rethinking}. We performed a grid search over the LoRA rank $r$, selecting the optimal dimension for each benchmark based on attribute fidelity. The generation is conducted within an \textit{img2img} framework, utilizing the original real images as spatial conditioning. For the automated pseudo-labeling phase (Stage B), we establish the contrastive evaluation parameters by applying full attribute complementation ($\rho = 1.0$). The Bayesian classifier is configured with an uninformative prior ($\pi_{\text{pos}} = 0.5$) and adopts a neutral decision boundary threshold of $\tau = 0.5$, optimally balancing semantic precision with synthetic data yield. For the downstream PAR models, we strictly adhere to the default hyperparameter settings specified for each architecture (Rethinking, PromptPAR, and SequencePAR) to ensure a fair evaluation of the data augmentation impact. When experimenting with different real-to-synthetic data ratios (e.g., 1:1 or 1:2), we construct each training batch by dynamically sampling from both the real and our verified synthetic datasets according to the specified proportion. All experiments, including generative fine-tuning, automated thresholding, and downstream PAR model training, were conducted on a single NVIDIA A40 GPU.

\subsection{Effect of LoRA Rank on Synthetic Image Generation}
\label{subsec:experimental_lora}
\paragraph{Objective}
We quantify how the LoRA rank affects the realism and domain similarity of the generation in Stage A (see \cref{fig:sec_method_overview}).

\paragraph{Experimental protocol} For each benchmark, we fine-tune the Stable Diffusion model using the training split of the corresponding dataset across four LoRA rank configurations: $r \in \{4, 8, 16, 32\}$. Following adaptation, we generate synthetic samples using the img2img pipeline conditioned on the training set's ground-truth labels and the prompts defined in \cref{subsec:method_stage_a}. To identify the optimal rank for each dataset, we evaluate the distributional similarity between the synthetic outputs and the real training data using four complementary metrics: FID~\cite{fid}, FD-DINO~\cite{fddino}, CMMD~\cite{cmmd}, and CFID~\cite{cfid}. The final selection is determined by the mean ranking aggregation across these indicators, identifying the configuration that yields the best overall balance between visual fidelity and domain alignment.

\paragraph{Results}
In \cref{tab:lora_resolutions}, we observe a clear relation between dataset image resolution and the best LoRA rank. For RAPv1 and RAPv2, whose average real resolutions are around $128\times 324$, the optimal ranks are relatively small (4-8). These sizes are closer to the native resolution (512x512) at which the diffusion model was pretrained, so only a modest low-rank update is needed to capture their style. In contrast, PA100K and PETA have smaller average resolutions, roughly $85\times 227$ for PA100K and $73\times 171$ for PETA/PETAzs, and the pedestrians are often farther away, and of lower visual quality than the RAP streams. Because these datasets are both farther in resolution and appearance from the Stable Diffusion training samples, they benefit from a higher LoRA rank (32), which gives the adapter more capacity to bridge the gap to the target surveillance style, for further information check the supplementary material. Finally, the zero-shot variants (RAPzs and PETAzs) inherit the best ranks of their parent datasets (RAPv2 and PETA, respectively), since they share the same acquisition conditions and resolution characteristics. Please refer to the supplementary material for additional details regarding the metric saturation  and the extended analysis of the LoRA rank metric analysis. We use the selected the rank per dataset in the following experiments.

\begin{table}[t]
  \centering
  \caption{Best LoRA rank \(r\) per dataset with average image resolution (in pixels) and distribution distances (↓ lower is better). The synthetic datasets PETAzs and RAPzs were generated using the LoRA models trained on their respective parent datasets (PETA and RAPv2).}

  \label{tab:lora_resolutions}
  \resizebox{\columnwidth}{!}{%
  \begin{tabular}{lccccccc}
    \toprule
Dataset & Best \(r\) & \(w\) (px) & \(h\) (px) & CFID$\downarrow$ & FID$\downarrow$ & FD-DINO$\downarrow$ & CMMD$\downarrow$ \\
    \midrule
    PA100K        & 32 & 85.6  & 227.4 & 25.11 & 48.54 & 1194.83 & 1.36 \\
    PETA          & 32 & 72.3  & 170.7 & 97.51 & 52.98 & 1516.49 & 1.36 \\
    PETAzs        & 32 & 72.9  & 171.5 & 92.93 & 57.07 & 1520.99 & 1.39 \\
    RAPv1         & 4  & 130.2 & 325.7 & 46.33 & 37.43 & 911.91  & 1.36 \\
    RAPv2         & 8  & 126.4 & 322.8 & 35.89 & 43.74 & 1125.52 & 1.50 \\
    RAPzs    & 8  & 133.9 & 333.6 & 70.19 & 45.81 & 1134.79 & 1.54 \\
    \bottomrule
  \end{tabular}%
  }
\end{table}

\subsection{Text-Image Representational similarity metrics analysis}
\label{subsec:txt_img_score_analysis}

\paragraph{Objective}
This experiment explores the optimal Text-Image representational similarity metric $\mathcal{S}$ for Stage~B (see \cref{fig:sec_method_overview}) by identifying the function that best captures semantic alignment.

\paragraph{Experimental Protocol} 
To evaluate the discriminative power of each similarity indicator $\mathcal{S}$, we analyze the separability between aligned $P(s \mid p_{\text{pos}})$ and misaligned $P(s \mid p_{\text{neg}})$ score distributions across all training splits. Following the strategy detailed in \cref{subsec:method_stage_c}, we generate positive probes $p_{\text{pos}_i}$ and contrast them with negative probes $p_{\text{neg}_i}$ created through attribute-flipping. We explicitly quantify the statistical distance and discriminative capacity between these distributions using the Bhattacharyya distance and the Area Under the Receiver Operating Characteristic curve (AUROC). For graphical clarity, we primarily visualize the results using the PETAzs dataset. Since it contains the highest density of active attributes per image, it provides the most comprehensive and representative scenario for analyzing prompt-length dynamics. However, the observed trends remain strictly consistent across all other evaluated datasets, as detailed in the supplementary material.

\paragraph{Sensitivity to Image-Text Semantic Misalignment} 
We first evaluate the metrics' sensitivity to the degree of semantic contradiction by systematically varying the complementation ratio $\rho \in \{0.25, 0.50, 0.75, 1.0\}$. For clarity of presentation, \cref{tab:comp_prompt_sepa_50_100} highlights the representative thresholds of partial ($\rho = 0.50$) and full ($\rho = 1.0$) complementation. As observed, BLIPScore consistently provides the highest separability, which increases monotonically as a larger fraction of attributes is inverted. An exhaustive breakdown of the performance across all $\rho$ values is provided in the supplementary material.

\begin{figure}[t]
  \centering
  
\includegraphics[width=\linewidth]{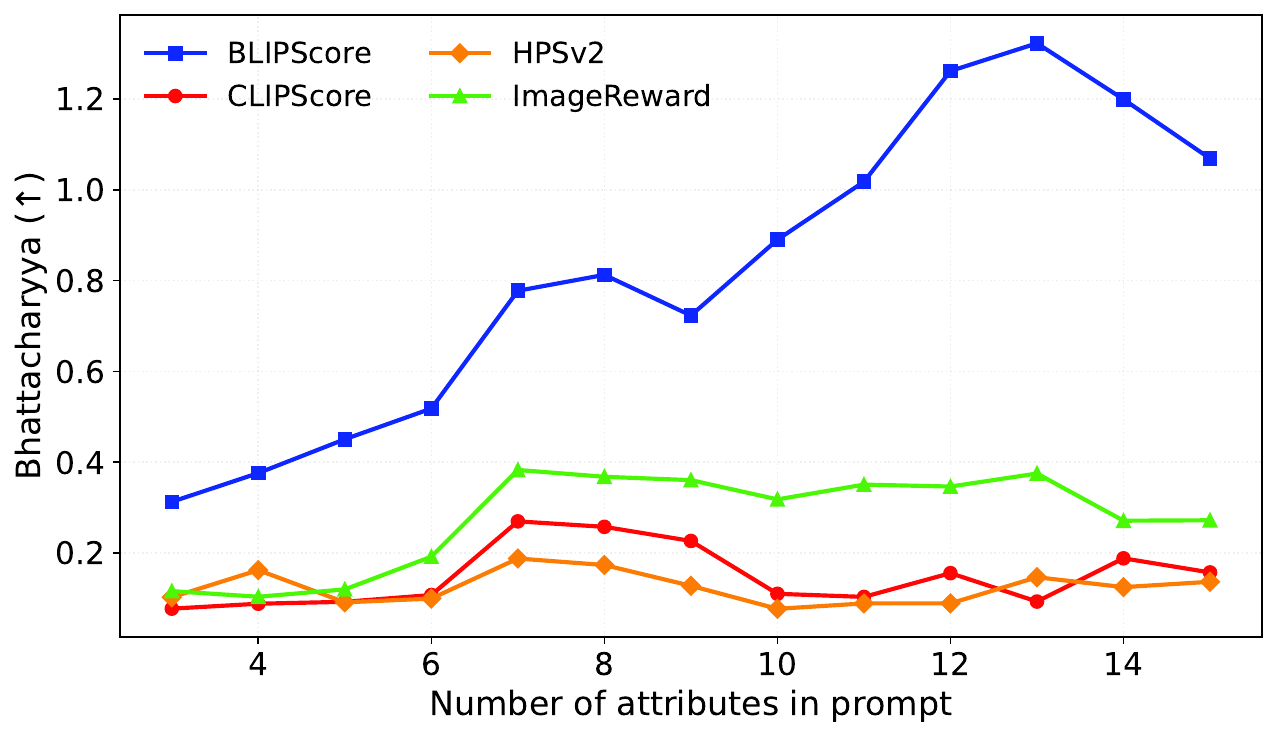}
  \caption{Impact of prompt length on metric separability. We report the Bhattacharyya distance (higher is better) between the BLIPScore distributions of the positive prompts ($\rho = 0$) and fully complemented negative prompts ($\rho = 1$). By fixing the complementation to its maximum ($\rho = 1$), this evaluation isolates how the total number of active attributes affects separability. Evaluated on the PETAzs testing split.}
\label{fig:com_num_attrib_metrics_PETAzs_bhat}
\end{figure}

\paragraph{Impact of Prompt Length} 
We further analyze the impact of the number of active attributes $L$ by fixing $\rho=1.0$ (fully complemented prompts) and grouping contrastive pairs by length. To ensure statistical robustness and prevent skew, each length group is balanced to 1,000 samples. As shown in \cref{fig:com_num_attrib_metrics_PETAzs_bhat}, the discriminative capacity of the BLIPScore is highly dependent on $L$. The BLIPScore curve demonstrate that the separability between positive and negative samples as more  attributes are added to the prompt. Extremely short prompts yield ambiguous scores due to insufficient semantic context, while an optimal range between 6 and 13 attributes is identified where BLIPScore achieves its most stable and discriminative behavior. For a detailed analysis of how prompt length impacts metric separability across all evaluated datasets, we direct the reader to the supplementary document.

\begin{table*}[!t]
\centering
\caption{Separability of representational similarity metrics under prompt complementation. For each dataset, we compare the scores of the original positive prompt ($\rho = 0.00$, no altered attributes) against negative prompts with partial ($\rho = 0.50$, half of the attributes flipped) and full ($\rho = 1.0$, all active attributes flipped) complementation. We report AUROC$\uparrow$ and Bhattacharyya distance (BHAT$\uparrow$) for four scorers. Higher is better.}
\label{tab:comp_prompt_sepa_50_100}
\resizebox{\textwidth}{!}{%
\begin{tabular}{l
cc cc  
cc cc  
cc cc  
cc cc  
}
\toprule
& \multicolumn{4}{c}{\textbf{BLIPScore} \cite{blip}} & \multicolumn{4}{c}{\textbf{CLIPScore} \cite{clipscore}} & \multicolumn{4}{c}{\textbf{ImgReward} \cite{imgreward}} & \multicolumn{4}{c}{\textbf{HPSv2Score} \cite{hpsv2score}} \\
\cmidrule(lr){2-5} \cmidrule(lr){6-9} \cmidrule(lr){10-13} \cmidrule(lr){14-17}
Dataset
& \multicolumn{2}{c}{p-0.50} & \multicolumn{2}{c}{p-1.00}
& \multicolumn{2}{c}{p-0.50} & \multicolumn{2}{c}{p-1.00}
& \multicolumn{2}{c}{p-0.50} & \multicolumn{2}{c}{p-1.00}
& \multicolumn{2}{c}{p-0.50} & \multicolumn{2}{c}{p-1.00} \\
& AUROC & BHAT & AUROC & BHAT
& AUROC & BHAT & AUROC & BHAT
& AUROC & BHAT & AUROC & BHAT
& AUROC & BHAT & AUROC & BHAT \\
\midrule
PA100K
& \textbf{0.72} & \textbf{0.08} & \textbf{0.84} & \textbf{0.26}
& 0.56 & 0.01 & 0.74 & 0.11
& 0.64 & 0.03 & 0.79 & 0.19
& 0.69 & 0.06 & \textbf{0.84} & \textbf{0.26} \\
PETA
& \textbf{0.85} & \textbf{0.28} & \textbf{0.98} & \textbf{1.09}
& 0.63 & 0.04 & 0.73 & 0.11
& 0.69 & 0.10 & 0.81 & 0.29
& 0.69 & 0.07 & 0.72 & 0.11 \\
PETAzs
& \textbf{0.89} & \textbf{0.39} & \textbf{0.98} & \textbf{0.99}
& 0.64 & 0.04 & 0.70 & 0.08
& 0.71 & 0.13 & 0.78 & 0.24
& 0.69 & 0.07 & 0.66 & 0.06 \\
RAPv1
& \textbf{0.86} & \textbf{0.29} & \textbf{0.96} & \textbf{0.75}
& 0.66 & 0.06 & 0.67 & 0.05
& 0.68 & 0.06 & 0.82 & 0.22
& 0.85 & 0.27 & 0.83 & 0.23 \\
RAPv2
& \textbf{0.89} & \textbf{0.38} & \textbf{0.98} & \textbf{0.97}
& 0.67 & 0.06 & 0.72 & 0.08
& 0.71 & 0.08 & 0.85 & 0.27
& 0.86 & 0.31 & 0.87 & 0.33 \\
RAPzs
& \textbf{0.89} & \textbf{0.39} & \textbf{0.97} & \textbf{0.92}
& 0.68 & 0.07 & 0.71 & 0.09
& 0.71 & 0.08 & 0.84 & 0.26
& 0.86 & 0.32 & 0.87 & 0.33 \\
\bottomrule
\end{tabular}%
}
\end{table*}

\subsection{Autolabeling Performance}
\label{subsec:end2end_labeling}

\paragraph{Objective}
This section evaluates the accuracy of our Bayesian framework as a classifier. Specifically, we assess the precision of the posterior probability $P(\text{aligned} \mid s)$ to serve as a reliable, high-fidelity autolabeling mechanism prior to downstream model training.

\begin{figure}[t]
  \centering
  
  \includegraphics[width=\linewidth]{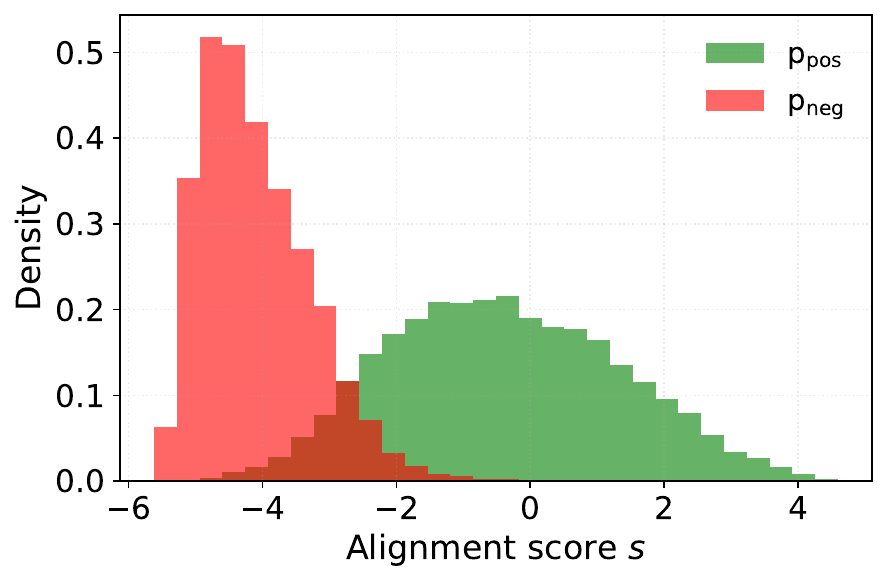}
  \caption{BLIPScore ($s$) distributions on the PETAzs training split. The histogram compares the scores obtained for the same generated images when evaluated against their label-consistent prompt $p_{\mathrm{pos}}$ ($\rho = 0$) (green) versus the fully complemented ($\rho = 1$) prompt $p_{\mathrm{neg}}$ (red).}
  \label{fig:blipscore_distributions_PETAzs}
\end{figure}

\paragraph{Experimental Protocol}
To establish a consistent baseline for the following experiments, we train the Bayesian classifier independently for each dataset using its respective real training split. We define two distinct sets of alignment scores: positive scores ($s^+$) extracted from label-consistent probes ($p_{\mathrm{probe}}$, $\rho=0$), and negative scores ($s^-$) extracted from fully complemented inconsistent probes ($p_{\mathrm{probe}}$, $\rho=1$). As illustrated in \cref{fig:blipscore_distributions_PETAzs}, these empirical score distributions exhibit unimodal and smooth characteristics with clear separability. This behavior directly justifies our two-component Gaussian formulation (\cref{subsec:method_stage_c}), providing a stable, closed-form approximation to effectively isolate valid generative attributes from semantic noise. Once the likelihood distributions are estimated and the model is calibrated, we evaluate its generalization and practical efficacy through three progressive analyses: (i) \emph{Alignment Classification and Attribute Labeling Performance}, (ii) \emph{Attribute Verification}, and (iii) \emph{Threshold Sensitivity Analysis}. While we illustrate the score distribution exclusively for PETAzs, the comprehensive empirical distributions demonstrating consistent separability for all other benchmarks can be found in the supplementary material.

\paragraph{Alignment Classification and Attribute Labeling Performance}
We first assess the classifier's core ability to accurately distinguish between $s^+$ and $s^-$ samples derived from the real test splits. As reported in \cref{tab:autolabeling_performance_combined}, the Bayesian classifier achieves a high global alignment accuracy exceeding 90\% across most datasets, with PA100K presenting a moderately lower but solid performance ($\sim$75\%). This strong macro-level classification successfully translates to an exceptionally high attribute-level pseudo-label mean Accuracy (mA), stabilizing around 0.90 for the PETA and RAP benchmarks, and reaching 0.75 for PA100K. Crucially, this reported mA is calculated exclusively on the target attributes $a_i$. This targeted evaluation specifically reflects the framework's capacity to detect only the subset of target attributes it was instructed to synthesize. Consequently, this metric is not directly comparable to standard PAR evaluations that predict the full set of attributes, since our approach intentionally filters out the non targeted attributes $a_i$. These results demonstrate that the Bayesian filter detects the presence of target attributes $a_i$ within the generated images with high precision, producing highly reliable pseudo-labels for dataset expansion. A comprehensive breakdown of the autolabeling accuracy for each individual attribute is provided in the supplementary material. Furthermore, we provide qualitative examples to offer visual insight into the decision boundaries of our method, with a detailed visual analysis available in the supplementary material.

\begin{figure}[h]
  \centering

  \includegraphics[width=\linewidth]{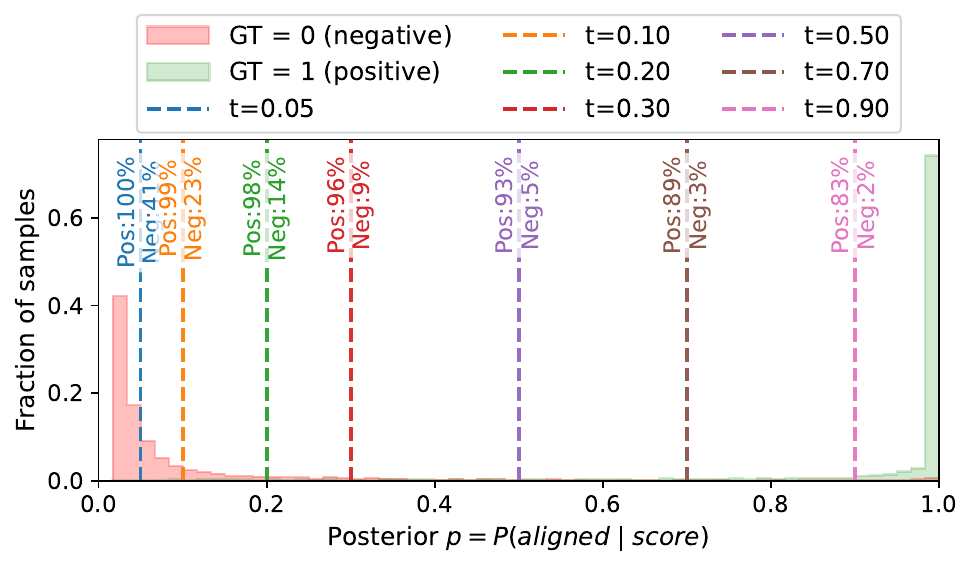}
  
   \caption{Posterior $P(\text{aligned}\mid s)$ from the Bayesian filter for ground-truth negatives and positives under varying decision thresholds $\tau$. The annotated percentages explicitly illustrate the filtering trade-off at each threshold: retaining valid generative attributes (Pos) versus blocking semantic noise (Neg). Scores are extracted from the PETAzs testing split.}
  \label{fig:posterior_hist}
\end{figure}

\paragraph{Threshold Sensitivity Analysis}
To determine the optimal decision threshold $\tau$, we analyze its direct impact on the classification error rates during the autolabeling process. As quantified by the annotated percentages in \cref{fig:posterior_hist}, varying $\tau$ controls the fundamental trade-off between false positives and false negatives. Our evaluation identifies $\tau = 0.5$ as the optimal operating point, preserving approximately 93\% of the true positive samples while keeping the false positive rate at merely 5\%. Furthermore, empirical tests varying $\tau \in \{0.3, 0.5, 0.7, 0.9\}$ revealed only marginal differences in the final pseudo-label assignments, demonstrating the robustness of the posterior probabilities. To validate that this threshold stability holds beyond the labeling phase, \cref{subsubsec:importance_score_threshold} provides a dedicated analysis on how these variations in $\tau$ specifically impact the final mA of the downstream PAR models. Please refer to the supplementary material for the complete empirical posterior distributions and the extended threshold sensitivity analysis across the remaining datasets.

\begin{table}[t]
\centering
\caption{Global pseudo-labeling performance across datasets. The table reports the validation/test accuracy of the Bayesian classifier, the overall mean accuracy (mA) of the generated pseudo-labels for real data, and the median active attributes evaluated per prompt.}
\label{tab:autolabeling_performance_combined}
\resizebox{\columnwidth}{!}{%
\renewcommand{\arraystretch}{1.15}
\begin{tabular}{@{}lcccccc@{}}
\toprule
Metric & PETA & PETAzs & RAPv1 & RAPv2 & RAPzs & PA100K \\
\midrule
Bayes acc. (val/test) & 0.91/0.92 & 0.93/0.94 & 0.90/0.90 & 0.92/0.92 & 0.91/0.92 & 0.75/0.73 \\
Pseudo-label mA       & 0.90      & 0.93      & 0.89      & 0.91      & 0.90      & 0.75      \\
Median Active Attr.   & 14        & 15        & 6         & 5         & 5         & 5         \\
\bottomrule
\end{tabular}%
}
\end{table}

\section{Application to Downstream PAR}
\label{sec:application_par}

The primary objective of this section is to determine whether synthetic data, generated and verified through ReSAGE-PAR, effectively enhances downstream PAR performance. To comprehensively assess this, our analysis is structured to validate four key hypotheses: (i) that generative augmentation outperforms standard pixel-level transformations; (ii) that our score-driven verification is the primary driver of performance gains; (iii) that these benefits are architecture-agnostic and scale efficiently with data volume; and (iv) that ReSAGE-PAR can push state-of-the-art frameworks. We adopt a standardized experimental protocol. Specifically, we construct augmented training sets by mixing real samples with our Bayesian-verified synthetic data, while strictly preserving the original hyperparameters and optimization settings (e.g., optimizer type, learning rate, and scheduler) of each evaluated PAR architecture.

\subsection{Evaluating ReSAGE-PAR as a Data Augmentation Strategy}
\label{subsubsec:comparison_dataaug}
\paragraph{Objective} 
We assess whether ReSAGE-PAR outperforms existing data augmentation techniques ranging from standard pixel-level transformations to recent generative approaches when applied to PAR tasks.

\paragraph{Experimental Protocol} 
We constructed an augmented training set for each dataset by generating synthetic samples at a 1:1 ratio relative to the original training size. Following our proposed pseudo-labeling, we retrain the downstream PAR models using the Rethinking \cite{rethinking} framework as our standardized baseline.

\paragraph{Results} 
As summarized in \cref{tab:labeling_syn_results}, ReSAGE-PAR achieves consistent mean Accuracy (mA) improvements across all datasets when training the baseline PAR model with the combined data. Notably, it yields the largest gains on PETAzs, RAPv1, and RAPzs. Our approach demonstrates stronger and more consistent performance across the board than prior references, confirming that ReSAGE-PAR is significantly more effective than standard pixel-level perturbations or naive generative augmentations. 

To provide deeper insight into these performance gains, \cref{tab:autolabeling_diagnostics} reports metrics on the autolabeling pipeline. The results demonstrate that ReSAGE-PAR significantly enhances attribute recognition. For instance, on RAPv2, 95.5\% of attributes show improved performance over the baseline. This success stems from our conservative refinement strategy, which systematically mitigates generative noise by neutralizing unverified attributes (setting pseudo-labels to 0) rather than discarding entire images. As shown in the "Data Refinement Actions" block, this process results in the systematic neutralization of an average of 4 to 7 unverified attributes per generated sample. Notably, ReSAGE-PAR consistently improves the vast majority of attributes while safely filtering only a small fraction of noisy samples.

\begin{table}[t]
\centering
\caption{Comparison of augmentation methods for PAR (mA$\uparrow$, \%). All are our runs of each method with Rethinking as PAR method \cite{rethinking}. ReSAGE-PAR (naive labelling) mixes generated images with real ones but copies the full label vector from the conditioning real image \cite{avss_zs}; ReSAGE-PAR uses our score-based autolabels. ``--'' indicates not evaluated.}
\label{tab:labeling_syn_results}
\resizebox{\columnwidth}{!}{%
\begin{tabular}{@{}lcccccc@{}}
\toprule
Method & PETA & PETAzs & RAPv1 & RAPv2 & RAPzs & PA100K \\
\midrule
Base \cite{rethinking}       & 84.10 & 71.47 & 79.60 & 78.50 & 71.98 & 80.44 \\
AutoAug \cite{autoaug}       & 84.48 & 71.19 & 79.93 & 78.56 & 72.56 & 80.98 \\
CutMix \cite{cutmix}         & 83.47 & 69.66 & 78.25 & 77.99 & 69.78 & 78.44 \\
Mixup \cite{mixup}           & 81.70 & 69.13 & 76.79 & 77.19 & 69.41 & 77.83 \\
RandAug \cite{randaug}       & 84.50 & 71.61 & 79.92 & 78.59 & 72.98 & 81.14 \\
Trivial \cite{trivaug}       & 84.25 & 71.48 & 80.05 & 78.75 & 73.10 & 80.93 \\
AugMix \cite{augmix}         & 84.31 & 71.71 & 79.79 & 78.48 & 73.02 & 80.73 \\
EnhancingZeroShot \cite{avss_zs} & --  & 73.07 & --    & --    & 75.23 & 81.58 \\
DataCentric \cite{method_avss_alonso} & -- & --  & 81.14 & 79.75 & 74.57 & -- \\
\midrule
\textbf{ReSAGE-PAR (naive labelling)}    & 84.40 & 72.90 & 79.80 & 78.48 & 72.70 & 81.30 \\
\textbf{ReSAGE-PAR}                & \textbf{85.30} & \textbf{75.30} & \textbf{82.40} & \textbf{80.94} & \textbf{76.20} & \textbf{83.00} \\
\bottomrule
\end{tabular}%
}
\end{table}

\begin{table}[t]
\centering
\caption{Impact of the autolabeling pipeline across datasets. The Accuracy block details the number of attributes where our generative augmentation improves the Rethinking baseline \cite{rethinking} performance (\textit{Imp.}), alongside the overall success rate (\textit{\%}). The Data Refinement block reports the total generated samples, the percentage of samples neutralized by our Bayesian decision threshold ($\tau = 0.5$) (\textit{Neut.}), and the average number of label corrections applied per retained image (\textit{Corr.}).}
\label{tab:autolabeling_diagnostics}
\resizebox{\columnwidth}{!}{%
\renewcommand{\arraystretch}{1.15} 
\begin{tabular}{@{}lcccccc@{}}
\toprule
Metric & PETA & PETAzs & RAPv1 & RAPv2 & RAPzs & PA100K \\
\midrule
\multicolumn{7}{@{}l}{\textbf{Accuracy}} \\
\quad Imp.    & 20   & 27   & 35   & 42   & 35   & 14   \\
\quad No Imp. & 10   & 4    & 8    & 2    & 7    & 4    \\
\quad \%      & 66.7 & 87.1 & 81.4 & 95.5 & 83.3 & 77.8 \\
\midrule
\multicolumn{7}{@{}l}{\textbf{Data Refinement Actions}} \\
\quad Total Syn. & 9.5K & 11K  & 33K  & 51K  & 17K  & 80K  \\
\quad Neut.(\%)  & 10.8 & 9.9  & 13.5 & 9.1  & 9.9  & 27.6 \\
\quad Corr.      & 7    & 7    & 5    & 5    & 4    & 4    \\
\bottomrule
\end{tabular}%
}
\end{table}

\subsection{The Role of Score-Driven Verification in ReSAGE-PAR}
\label{subsubsec:importance_score_threshold}

\paragraph{Objective} 
Building upon the overall gains reported in \cref{subsubsec:comparison_dataaug}, we isolate the exact source of these improvements by evaluating the critical role of the score-driven verification Stage~C. Specifically, we aim to demonstrate two key points: (i) that the selective neutralization of misaligned attributes (setting pseudo-labels to 0) and active label correction are the primary drivers of performance, rather than mere data volume; and (ii) that this autolabeling process is highly robust to the specific choice of the decision threshold $\tau$.

\paragraph{Experimental Protocol} 

To establish a direct comparison, both the full pipeline and the ReSAGE-PAR (naive labelling) variant utilize a fixed 1:1 real-to-synthetic initial generation ratio. Using the Rethinking \cite{rethinking} framework as our baseline, we conduct an ablation study to isolate the impact of ours Stage~B and Stage~C (see \cref{subsec:method_stage_b,subsec:method_stage_c}). The ReSAGE-PAR (naive labelling) variant follows the strategy of prior generative methods \cite{avss_zs}, where each synthetic sample is trained using the full attribute vector copied directly from its conditioning real image $x^{(k)}$ being $\tilde{\mathbf{y}}^{(k)} = \mathbf{y}^{(k)}$. This approach completely bypasses the verification and refinement logic of Stage~B and Stage~C (see \cref{subsec:method_stage_b,subsec:method_stage_c}). To evaluate the sensitivity of this Stage~C, we test our framework across a range of thresholds $\tau \in \{0.3, 0.5, 0.7, 0.9\}$, using the equilibrium point $\tau=0.5$ (established in \cref{subsec:end2end_labeling}) as our reference.

\begin{table}[t]
\centering
\caption{Ablation study on the decision threshold $\tau$ during Stage~C. Performance is reported in mean Accuracy (mA, \%). The results demonstrate that the autolabeling process is highly robust, maintaining stable downstream PAR performance across a wide range of threshold values. Best results in bold.}
\label{tab:threshold_ablation}
\resizebox{\columnwidth}{!}{%
\begin{tabular}{ccccccc}
\toprule
Threshold ($\tau$) & PETA & PETAzs & RAPv1 & RAPv2 & RAPzs & PA100K \\
\midrule
0.3 & 85.43 & 75.05 & \textbf{82.79} & 81.03 & 75.12 & 82.82 \\
0.5 & 85.30 & \textbf{75.30} & 82.40 & 80.94 & \textbf{76.20} & 83.00 \\
0.7 & \textbf{85.60} & 75.21 & 82.03 & 80.92 & 75.41 & \textbf{83.06} \\
0.9 & 85.41 & 74.79 & 82.31 & \textbf{81.05} & 75.46 & 82.40 \\
\bottomrule
\end{tabular}%
}
\end{table}

\paragraph{Results} 
As shown in \cref{subsubsec:comparison_dataaug}, the ReSAGE-PAR (naive labelling) variant provides only marginal benefits over the baseline, clearly indicating that image generation alone is insufficient due to partial misalignment in the fine-grained attribute rendering of the synthesized output. Incorporating our score-driven autolabeling explicitly solves this bottleneck. As detailed in \cref{tab:threshold_ablation}, the downstream performance remains remarkably stable across a wide range of values for $\tau$. Operating at the reference $\tau=0.5$ successfully neutralizes misaligned attributes by setting their pseudo-labels to 0, ensuring that the supervision signal remains semantically grounded. This systematic validation effectively transforms raw generative outputs into reliable supervision, significantly improving per-attribute accuracy without requiring sensitive hyperparameter tuning.

\subsection{Synthetic Data Expansion and Backbone Generalization}
\label{subsubsec:syn_expansion}
\paragraph{Objective} 
A robust data augmentation module must not only provide gains for a specific setup, but also generalize across different network architectures and scale effectively with increasing volumes of synthetic data. We evaluate both the architectural versatility of our pipeline and the effect of scaling for synthetic data expansion.

\paragraph{Experimental Protocol} 
Using the Rethinking pipeline, we evaluate our autolabeled synthetic data across three distinct backbones architectures: ResNet50, BN-Inception, and Swin Transformer. To analyze synthetic data volume, testing real-to-synthetic ratios of 1:0.5, 1:1, and 1:2.

\paragraph{Results} 
As detailed in \cref{tab:rebuttal_rethinking}, our augmentation consistently improves mA across all three evaluated backbones, confirming that the benefits of our ReSAGE-PAR are architecture-agnostic. Scaling the verified synthetic data further to a 1:2 ratio yields additional gains. For instance, using ResNet50, even a conservative ratio of 1:0.5 with ReSAGE-PAR outperforms a 1:1 ratio with naive labeling from \cref{subsubsec:comparison_dataaug}. Notably, the results demonstrate that progressively scaling the amount of our verified synthetic data consistently improves downstream performance across all evaluated architectures. Those results demonstrate that quality control matters significantly more than raw quantity. 

\begin{table}[h]
\centering
\caption{mA results per dataset for \textit{Rethinking} \cite{rethinking} with different backbones and real:synthetic ratios. Bold indicates best results per backbone.}
\label{tab:rebuttal_rethinking}

\resizebox{\columnwidth}{!}{%
\renewcommand{\arraystretch}{1.10}
\begin{tabular}{@{}lccccccc@{}}
\toprule
\textbf{Backbone} & \textbf{Ratio (Real:Syn)} & PETA & PETAzs & RAPv1 & RAPv2 & RAPzs & PA100K \\
\midrule

\multirow{4}{*}{\textbf{ResNet50}} 
 & 1:0    & 84.10 & 71.47 & 79.60 & 78.50 & 71.98 & 80.44 \\
 & 1:0.5  & 84.87 & 73.85 & 81.41 & 80.00 & 74.48 & 81.56 \\
 & 1:1    & 85.30 & 75.30 & 82.40 & 80.94 & 76.20 & 83.00 \\
 & 1:2    & \textbf{85.75} & \textbf{75.58} & \textbf{83.10} & \textbf{81.82} & \textbf{76.91} & \textbf{83.07} \\
\midrule

\multirow{4}{*}{\textbf{BN-Inception}} 
 & 1:0    & 83.47 & 70.95 & 78.49 & 77.85 & 70.68 & 79.60 \\
 & 1:0.5  & 84.17 & 73.34 & 80.55 & 79.45 & 73.42 & 81.65 \\
 & 1:1    & 85.15 & 75.29 & 81.92 & 80.41 & 74.65 & 82.20 \\
 & 1:2    & \textbf{85.49} & \textbf{75.76} & \textbf{82.33} & \textbf{81.38} & \textbf{76.44} & \textbf{82.42} \\
\midrule

\multirow{4}{*}{\textbf{Swin}} 
 & 1:0    & 73.01 & 59.69 & 68.24 & 70.00 & 62.58 & 72.06 \\
 & 1:0.5  & 77.47 & 65.00 & 73.35 & 73.15 & 67.72 & 74.85 \\
 & 1:1    & 78.67 & 66.25 & 74.88 & 74.60 & 69.92 & 75.76 \\
 & 1:2    & \textbf{79.19} & \textbf{68.21} & \textbf{75.98} & \textbf{76.04} & \textbf{71.31} & \textbf{76.29} \\
\bottomrule
\end{tabular}%
}
\end{table}

\begin{table}[h]
\centering
\caption{Comparison with SOTA PAR methods. We integrate our ReSAGE-PAR synthetic data (\checkmark) at a 1:2 (real:synthetic) ratio into representative architectures.  Bold indicates the best results per dataset/SOTA method.}
\label{tab:rebuttal_best_promptpar_sota}
\resizebox{\columnwidth}{!}{%
\renewcommand{\arraystretch}{1.15} 
\begin{tabular}{@{}lccccccc@{}}
\toprule
\textbf{Method} & \textbf{ReSAGE-PAR} & \textbf{PETA} & \textbf{PETAzs} & \textbf{RAPv1} & \textbf{RAPv2} & \textbf{RAPzs} & \textbf{PA100K} \\
\midrule
Rethinking \cite{rethinking} & & 84.10 & 71.47 & 79.60 & 78.50 & 71.98 & 80.44 \\
Rethinking \cite{rethinking} & \checkmark & \textbf{85.75} & \textbf{75.58} & \textbf{83.10} & \textbf{81.82} & \textbf{76.91} & \textbf{83.08} \\
\midrule
SequencePAR \cite{sequencepar} & & 82.52 & \textbf{73.83} & 79.05 & 76.95 & 75.49 & \textbf{83.91} \\
SequencePAR \cite{sequencepar} & \checkmark & \textbf{84.01} & 73.04 & \textbf{79.46} & \textbf{78.40} & \textbf{75.54} & 83.77 \\
\midrule
PromptPAR \cite{promptPAR} & & 88.32 & 79.11 & 86.33 & 83.83 & 80.91 & 87.67 \\
PromptPAR \cite{promptPAR} & \checkmark & \textbf{89.02} & \textbf{80.94} & \textbf{87.71} & \textbf{85.94} & \textbf{83.47} & \textbf{89.37} \\

\bottomrule
\end{tabular}%
}
\end{table}

\subsection{Integration with SOTA PAR Frameworks}

\paragraph{Objective} 
Since our ReSAGE-PAR is inherently architecture-agnostic, the objective of this final evaluation is to benchmark its ultimate potential by integrating it into modern SOTA PAR pipelines.

\paragraph{Experimental Protocol} 
To benchmark the scalability and architectural compatibility of ReSAGE-PAR, we integrate our verified synthetic data into three representative PAR frameworks: the optimized \textit{Rethinking} baseline \cite{rethinking}, the sequence-based \textit{SequencePAR} \cite{sequencepar}, and the transformer-based \textit{PromptPAR} \cite{promptPAR}. For each architecture, we strictly adhere to its original training configurations and hyperparameters, adopting a 1:2 real-to-synthetic ratio as suggested by the scaling trends in \cref{subsubsec:syn_expansion}.

\paragraph{Results}
the results summarized in \cref{tab:rebuttal_best_promptpar_sota} demonstrate the consistent efficacy of ReSAGE-PAR across diverse PAR architectures. When integrated at a 1:2 ratio, our verified synthetic data yields significant performance gains for both the Rethinking baseline and the transformer-based PromptPAR across all evaluated datasets. Notably, the combination with PromptPAR achieves the highest mean Accuracy (mA) in every benchmark, reaching up to 89.37\% on PA100K and 83.47\% on the challenging RAPzs split.

While the augmentation proves highly beneficial for the majority of configurations, we observe a slight performance when using SequencePAR. In this specific architecture, although ReSAGE-PAR improves performance in four out of six datasets (PETA, RAPv1, RAPv2, and RAPzs), it shows a marginal decrease on the PETAzs and PA100K splits. Nevertheless, the consistent gains achieved in Rethinking and PromptPAR confirm that our framework successfully improves both standard and SOTA models by providing more reliable supervision for the synthetic samples.

\section{Conclusions}
\label{sec:conclusions}

In this work, we addressed the dual challenge of domain adaptation and label reliability in synthetic data generation for PAR.
We presented ReSAGE-PAR, which efficiently adapts Stable Diffusion to low-resolution surveillance domains using LoRA, coupled with a Bayesian verification mechanism to filter generative hallucinations. 

Our extensive experiments yield three key insights.
First, alignment matters more than scale: we demonstrated that a smaller, verified synthetic subset (ratio 1:0.5) consistently outperforms larger, naively labeled sets (1:1), proving that rigorous quality control is the bottleneck for effective data expansion.
Second, our proposed Bayesian autolabeler is robust and effective, showing insensitivity to threshold variations while effectively separating aligned from misaligned attributes.
Third, the method is architecture-agnostic and SOTA-compatible; it acts as a plug-and-play augmentation module that yields consistent gains across diverse backbones (ResNet, Inception, Swin) and pushes SOTA frameworks like PromptPAR to new performance highs on challenging zero-shot benchmarks. Unlike resource-intensive MLLM-based annotation, ReSAGE-PAR offers a scalable, computationally efficient path for massive dataset expansion.

\section{Limitations and Future Work}
\label{sec:limitations_future}

While our pipeline effectively generates synthetic PAR data with reliable labels, certain limitations remain. First, our generation relies on the semantic knowledge of the pretrained Stable Diffusion model. Although LoRA adapts the visual style, extremely rare or fine-grained attributes not well-represented in the base model's pretraining data (e.g., specific logo types or obscure clothing patterns) may suffer from low generation fidelity, limiting the effectiveness of autolabeling in long-tail scenarios. Second, our autolabeling mechanism operates at the prompt-level granularity. Because the LLR is computed based on the similarity between the full set of active attributes and the image, the Bayesian filter effectively assesses the global alignment of the generated sample. Consequently, if a generated image fails to render a specific attribute, our conservative neutralization strategy may suppress the labels for other, correctly rendered attributes within the same prompt, thereby limiting our ability to reclaim valid supervision signals from partially successful generations. Third, our Bayesian classifier binarizes the continuous alignment scores into hard labels (0 or 1). This simplification ignores the inherent uncertainty of ambiguous samples, which, while beneficial for training stability, discards potential nuance that could be valuable for more complex learning objectives.

Future work will focus on addressing the long-tailed distribution of pedestrian attributes by leveraging our pipeline to synthesize rare classes. Additionally, we plan to extend our Bayesian framework to support fine-grained, per-attribute verification, allowing for more granular control over label neutralization. This will involve investigating attribute inter-dependencies to explore how our scoring mechanism can selectively validate co-occurring features, preserving supervision signals even when specific attributes are missing while others are visually present.

\bibliographystyle{IEEEtran}
\bibliography{main}

\vspace{-2cm}

\begin{IEEEbiography}[{\includegraphics[width=1in,height=1.25in,clip,keepaspectratio]{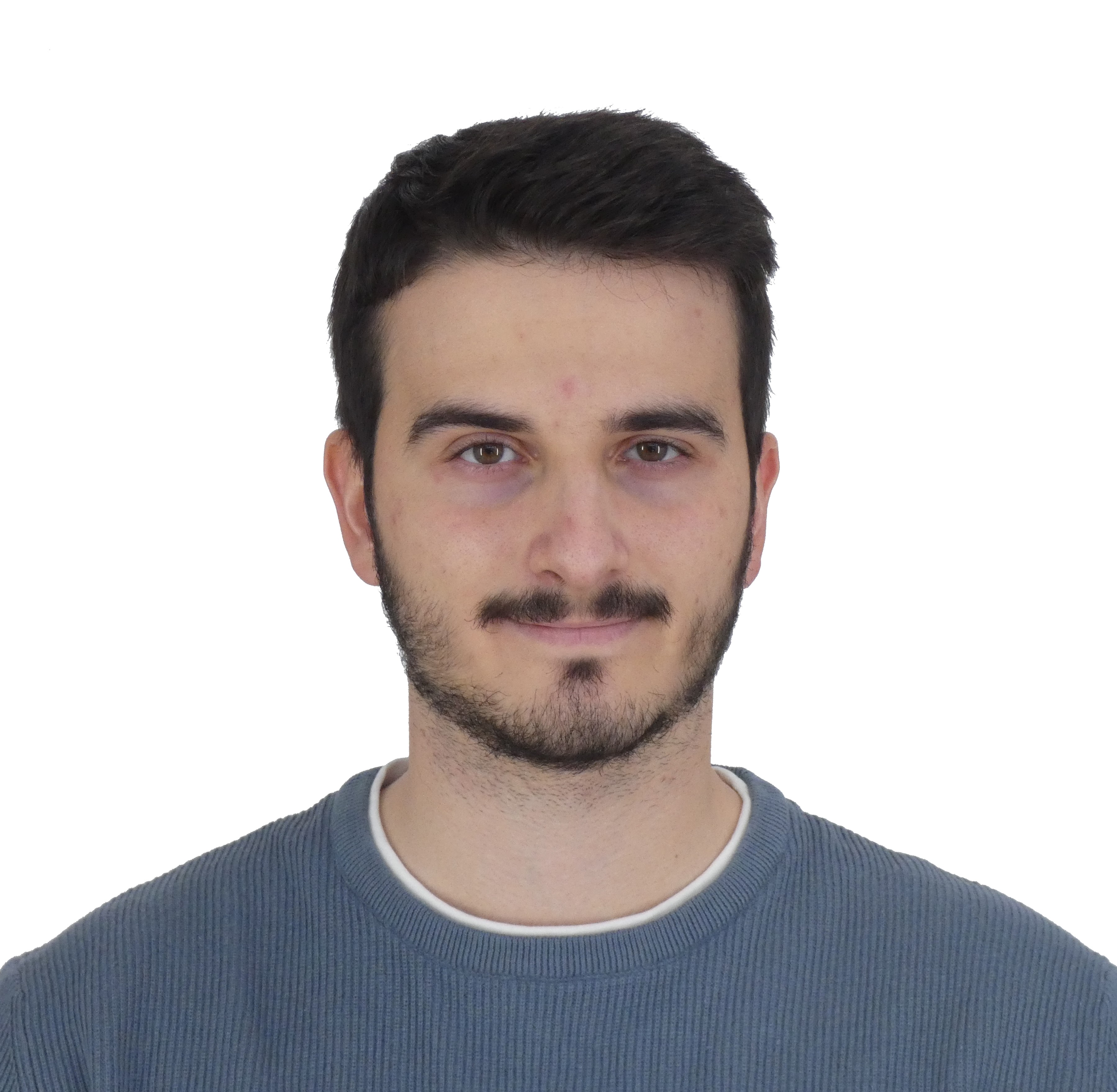}}]{Pablo Ayuso-Albizu} received the B.S. degree in Computer Engineering in 2021, and the M.S. degree in Deep Learning for Audio and Video Signal Processing in 2022, both from the Universidad Autónoma de Madrid (UAM), Madrid, Spain. In 2024, he began pursuing his Ph.D. degree with the Video Processing and Understanding (VPU) Lab at UAM. His current research interests include deep learning, pedestrian attribute recognition, vision-language models, and generative data augmentation.

\end{IEEEbiography}

\vspace{-2cm}

\begin{IEEEbiography}[{\includegraphics[width=1in,height=1.25in,clip,keepaspectratio]{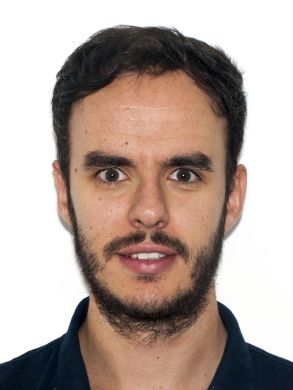}}]{Pablo Carballeira} received the Telecommunication Engineering degree (five years engineering program), ommunications Technologies and Systems Master degree (two year MS program) and the Ph.D. degree in Telecommunication from the Universidad Polit\'ecnica de Madrid (UPM) in 2007, 2010 and 2014 respectively. From 2008 to 2017 he has been a member of the Grupo de Tratamiento de Im\'agenes (Image Processing Group) at the UPM. Since 2017 he is a member of the Video Processing and Understanding Lab, at the Universidad Aut\'onoma de Madrid (UAM), and Associate Professor at UAM since 2022. His research interests include computer vision, video coding, and quality of experience evaluation for immersive visual media. He has been actively involved in European projects, national projects, and standardization activities from ISO\textquotesingle s Moving Picture Experts Group (MPEG) related to lightfield and free-navigation video, technologies.
\end{IEEEbiography}

\begin{IEEEbiography}
[{\includegraphics[width=1in,height=1.25in,clip,keepaspectratio]{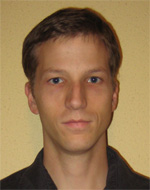}}]{Juan C. SanMiguel} received the Ph.D. degree in computer science and telecommunication from the Autonomous University of Madrid (UAM), Madrid, Spain, in 2011. From 2013 to 2014, he was a Postdoctoral Researcher with Queen Mary University of London, London, U.K., under a Marie Curie IAPP Fellowship. Since 2025, he has served as Institutional Delegate for Artificial Intelligence at UAM. He is currently an Associate Professor with the Autonomous University of Madrid and a Researcher with the Video Processing and Understanding Laboratory. He is also an Associate Editor of The Visual Computer. He has authored more than 70 journal and conference papers. His research interests include computer vision, with a focus on domain adaptation, synthetic data reliability estimation, and multi-camera activity understanding for video segmentation and tracking.

\end{IEEEbiography}

\vspace{-15cm}

\begin{IEEEbiography}[{\includegraphics[width=1in,height=1.25in,clip,keepaspectratio]{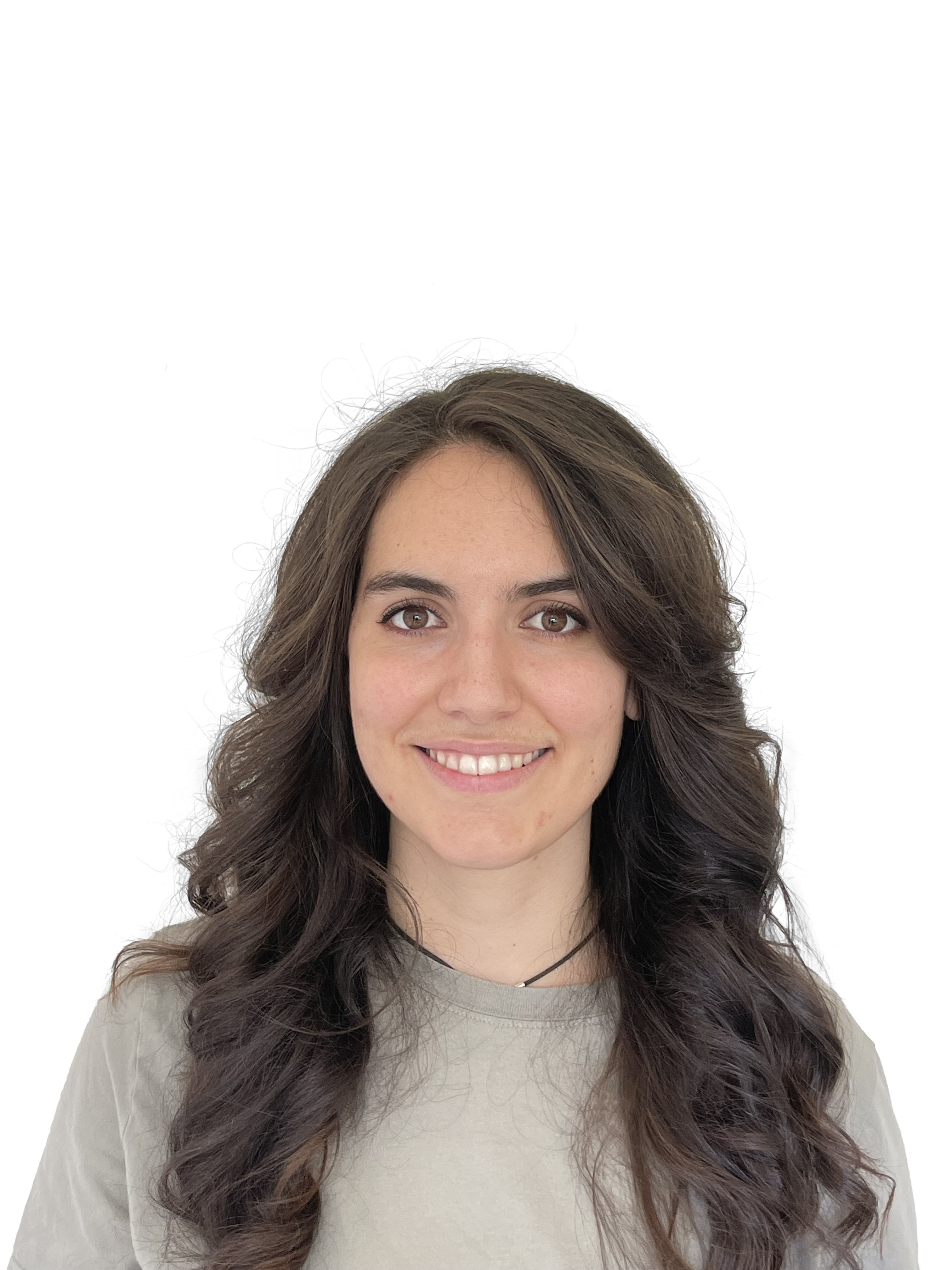}}]{Paula Moral de Eusebio} received the degree in Telecommunications Engineering in 2017 at the Universidad Autónoma de Madrid. In 2019 she obtained the titles belonging to the International Joint Master Program in Image Processing and Computer Vision (IPCV) at the Universidad Autónoma de Madrid (Spain), the Pázmány Péter Catholic University (Hungary) and the Université de Bordeaux (France). In 2024, she obtained her PhD in Computer Engineering and Telecommunications. From 2017 to the present, she has been with the Video Processing and Understanding Lab (VPU-Lab) at the Universidad Autónoma de Madrid as a researcher. Throughout her career, she has been the recipient of a doctoral research initiation grant (2019–2022) and an FPI-UAM doctoral fellowship (2022–2025). Following her role as a Substitute Professor in 2025, she now serves as an Assistant Professor. Her lines of research are focused on object detection and association in multiview scenarios.
\end{IEEEbiography}


\clearpage 

\begin{center}
    \LARGE \textbf{Supplementary Material for ReSAGE-PAR: Representational Similarity Assessment for Generative Expansion in PAR} 
\end{center}
\vspace{1cm}

\setcounter{section}{0}
\setcounter{figure}{0}
\setcounter{table}{0}
\setcounter{equation}{0}

\renewcommand{\thesection}{S-\Roman{section}}
\renewcommand{\thefigure}{S\arabic{figure}}
\renewcommand{\thetable}{S-\Roman{table}}
\renewcommand{\theequation}{S\arabic{equation}}

\renewcommand{\thesection}{S-\Roman{section}}
\renewcommand{\thefigure}{S\arabic{figure}}
\renewcommand{\thetable}{S-\Roman{table}}

\IEEEpubidadjcol


\maketitle

This supplementary material provides additional technical details, extended empirical evaluations, and comprehensive quantitative results to support the findings presented in the main manuscript of ReSAGE-PAR. The document is structured as follows. First, \textbf{\cref{sec:anexa_promptconstruction}} details the generation-prompt construction process and outlines the dataset-specific attribute complementation strategy used to generate negative probes. \textbf{\cref{sec:anex_lora}} expands on the Dataset-Aware Generative Fine-Tuning phase, providing further insights into the LoRA adaptation and its impact on image resolution and style. \textbf{\cref{sec:anexa_promptfidelity}} presents the extended Text-Image Representational Similarity Metrics Analysis, including the exhaustive evaluation of separability across different complementation ratios. \textbf{\cref{sec:anexa_distributionsBlipScore}} visualizes the complete Analysis of BLIPScore Distributions for all evaluated datasets, confirming the unimodal and separable nature of the scores. \textbf{\cref{sec:anexa_attributesimproved}} provides the comprehensive Autolabeling Accuracy per Attribute, detailing the precise pseudo-labeling performance for every individual attribute and visual results. Finally, \textbf{\cref{sec:anexa_thresholdAnalysis}} contains the extended Threshold Sensitivity Analysis, demonstrating the robustness of our Bayesian decision boundary across the remaining benchmarks.

{

\section{Prompt construction}
\label{sec:anexa_promptconstruction}

This section details how we build the generation prompts $p_{\text{gen}}$ deterministically from binary attributes (emitting a clause if and only if the attribute is active), using dataset-aware templates that control attribute ordering and a single \texttt{wearing} gate for clothing tokens. We also describe how the negative probes $p_{\text{neg}}$ are constructed via intra-group attribute substitution specifically, by replacing a target attribute with another randomly sampled from the same mutually exclusive family (e.g., hs, ub, lb) and how partial complementation ($\rho \in \{0.25, 0.50, 0.75, 1.0\}$) is applied. For concreteness, \cref{tab:supp_prompt_examples} lists the positive probes $p_{\text{pos}}$ (where $p_{\text{pos}} = p_{\text{gen}}$) and their corresponding fully complemented negative probes $p_{\text{neg}}$ $\rho = 100\%$ for each dataset.

\paragraph{Prompt construction}
We deterministically synthesize the generation prompt $p_{\text{gen}}$ from binary attributes: a clause is emitted if and only if the attribute is $1$, and omitted otherwise; the prompt begins with a random stub from \{\texttt{a}, \texttt{there is a}\}. For \textbf{RAPzs/RAPv2/RAPv1}, if \emph{gender} is available we emit \texttt{woman}/\texttt{man}, then append attribute groups in a fixed order: hair style (\texttt{hs-*}) as ``\texttt{with <hair>}'', actions (\texttt{action-*}) as ``\texttt{is <action>}'', and attachments (\texttt{attachment-*}) as bare tokens. Clothing uses a single \texttt{wearing} word, after which we concatenate upper-body (\texttt{ub-*}), lower-body (\texttt{lb-*}), and footwear (\texttt{shoes-*}) items that are active. For \textbf{PETA/PETAzs}, we emit gender (if present), then hair \emph{color} and \emph{type} (both under ``\texttt{with}''), carrying items (``\texttt{carrying <item>}'') and accessories (``\texttt{with <accessory>}''), finishing with a single \texttt{wearing} gate for colors/types of upper/lower garments and footwear. \textbf{PA100K} adds a \emph{view} clause (``\texttt{from <front/side/back>}'') before the generic ``\texttt{with}'' attributes and the gated \texttt{wearing} for upper/lower items and boots; some type slots take the article ``\texttt{a}'' (e.g., ``\texttt{wearing a <upper-type>}'').

\paragraph{Negative probes}
To form the complemented negative probes $p_{\text{neg}}$, we modify a fraction $\rho$ of the emitted clauses using an intra-group attribute substitution strategy. Attributes are partitioned into mutually exclusive families (e.g., \emph{gender}, \emph{hair style/color}, \emph{upper/lower garment types}, \emph{footwear}, \emph{view}, \emph{roles}, \emph{actions}), and for each attribute $a$ we define $C(a)\subseteq\mathcal{A}$ as the set of admissible alternatives within the same family, excluding $a$ itself (e.g., \texttt{woman}$\to$\texttt{man}, \texttt{long hair}$\to$\texttt{short}/\texttt{bald}, \texttt{jeans}$\to$\texttt{skirt}/\texttt{trousers}, \texttt{front}$\to$\texttt{back}/\texttt{side}). For each selected active clause, we replace $a$ with an element drawn from $C(a)$, preserving the original template token (\texttt{with/is/wearing/from}), the attribute order, and the single \texttt{wearing} gate. Families are processed independently; when $|C(a)|>1$, we select one alternative using a uniform random distribution.

\begin{table*}[t]
\centering
\caption{Examples of positive probes ($p_{\text{pos}}$) and their fully complemented negative counterparts ($p_{\text{neg}}$ $\rho = 100\%$ ) per dataset. Text reflects the dataset vocabulary.}
\label{tab:supp_prompt_examples}
\small
\setlength{\tabcolsep}{4pt}
\renewcommand{\arraystretch}{1.12}
\begin{tabularx}{\textwidth}{@{}p{1.6cm}>{\ttfamily\raggedright\arraybackslash}X>{\ttfamily\raggedright\arraybackslash}X@{}}
\toprule
\textbf{Dataset} & \textbf{Aligned prompt ($p_{\text{pos}}$)} & \textbf{Complemented ($p_{\text{neg}}$)} \\
\midrule
PA100K &
there is a woman from front wearing short sleeve a plaid skirt or dress &
there is a man from side wearing long sleeve a logo trousers \\

PETA &
there is a woman with black short hair carrying messenger bag with nothing wearing black casual other black casual long skirt &
there is a man with white long hair carrying luggage case with muffler wearing yellow formal logo white jeans shorts \\

PETAzs &
there is a woman with brown long hair carrying nothing with nothing wearing blue long sleeve casual jacket grey casual trousers white shoes &
there is a man with black bald carrying luggage case with muffler wearing pink sweater no sleeve logo white shorts shortskirt pink boots \\

RAPv1 &
a man with black hair is carrying by hand plastic bag wearing jacket jeans sports shoes &
a woman with bald head is calling hand trunk wearing sweater long trousers casual shoes \\

RAPv2 &
there is a woman with long hair with black hair is carrying by hand shoulder bag paper bag wearing tight long trousers sports shoes &
a man with bald head with glasses is holding backpack plastic bag wearing sweater tight trousers boots \\

RAPzs &
there is a woman with long hair with black hair is carrying by hand hand bag wearing cotton tight trousers boots &
a man with bald head with glasses is talking backpack wearing shirt skirt sports shoes \\
\bottomrule
\end{tabularx}
\end{table*}

\section{Extended Analysis of Dataset-Aware Generative Fine-Tuning}
\label{sec:anex_lora}

Following the experimental protocol detailed in the main manuscript, this section provides a comprehensive quantitative evaluation of the Stable Diffusion adaptation process across varying LoRA ranks ($r \in \{4, 8, 16, 32\}$). While the main text summarizes the optimal rank selection driven by dataset resolution and target surveillance style, here we present the full numerical breakdown. Specifically, \cref{tab:supp_lora_metrics} reports the complete results for CFID, FID, FD-DINO, and CMMD across all datasets, detailing the mean values over five independent seeds and the ranking aggregation used to determine the optimal configuration per benchmark.
 
\paragraph{Metric Saturation} 
To further understand how the synthetic sample size affects the stability of these distributional metrics, we analyze their convergence as the number of generated images ($N$) increases. Since this evaluation benefits significantly from large sample volumes to reliably measure distances, we provide these saturation plots exclusively for PA100K, which, as the largest benchmark, offers the most robust scenario. We plot CFID, FID, FD-DINO, and CMMD ($\downarrow$) as a function of the number of samples employed (N) from the training split available. For every $N$, we draw synthetic populations from five independent random seeds and report the aggregated distance. As illustrated in \cref{fig:supp_lora_FID_pa100k} and \cref{fig:supp_lora_FDDINO_pa100k}, both FID$\downarrow$ and FD-DINO$\downarrow$ improve rapidly and largely saturate by $N \approx 5000$. In \cref{fig:supp_lora_CMMD_pa100k}, CMMD$\downarrow$ continues to decrease until about $N \approx 10000$ before flattening. In contrast, \cref{fig:supp_lora_CFID_pa100k} shows that CFID$\downarrow$ decays much more slowly; because it is heavily tied to text-image conditioning, it requires comparison against the full synthetic population (ensuring comprehensive semantic and attribute coverage) to fully stabilize and reach its plateau.

\clearpage

\begin{figure}[t]
  \centering

\includegraphics[trim=0.25cm 0.4cm 0.25cm 1cm, clip, width=\linewidth]{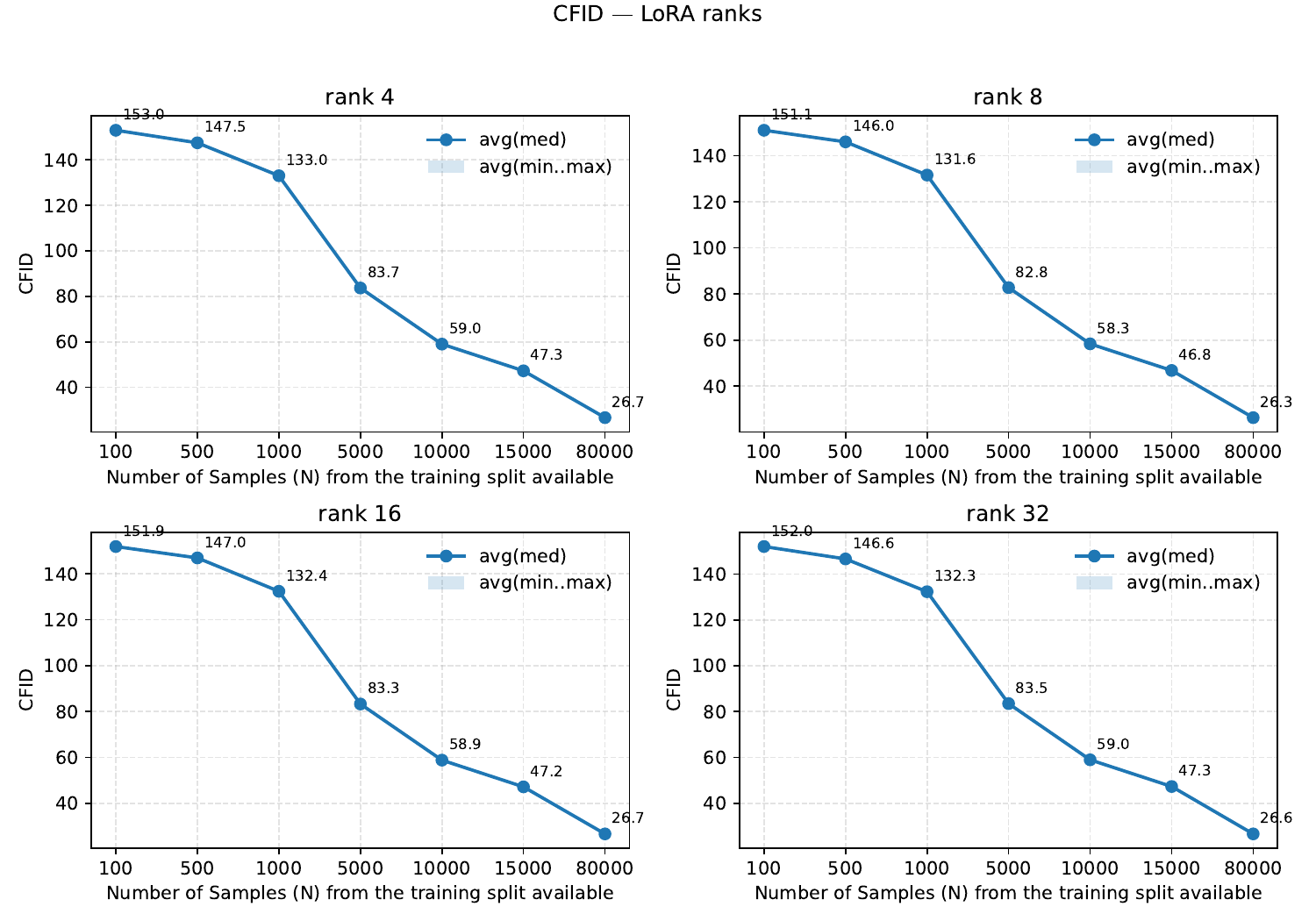}
\caption{\textbf{CFID (↓) vs.\ synthetic sample size \(N\)} for the PA100K dataset across LoRA ranks \(r\in\{4,8,16,32\}\). We report avg.\ median and the avg.\ min–max envelope over five seeds per \(N\). All ranks converge to similar CFID at large \(N\), with small rank effects; increasing \(N\) drives most of the gain.}
\label{fig:supp_lora_CFID_pa100k}
\end{figure}

\begin{figure}[t]
  \centering
  
\includegraphics[trim=0.25cm 0.4cm 0.25cm 1cm, clip, width=\linewidth]{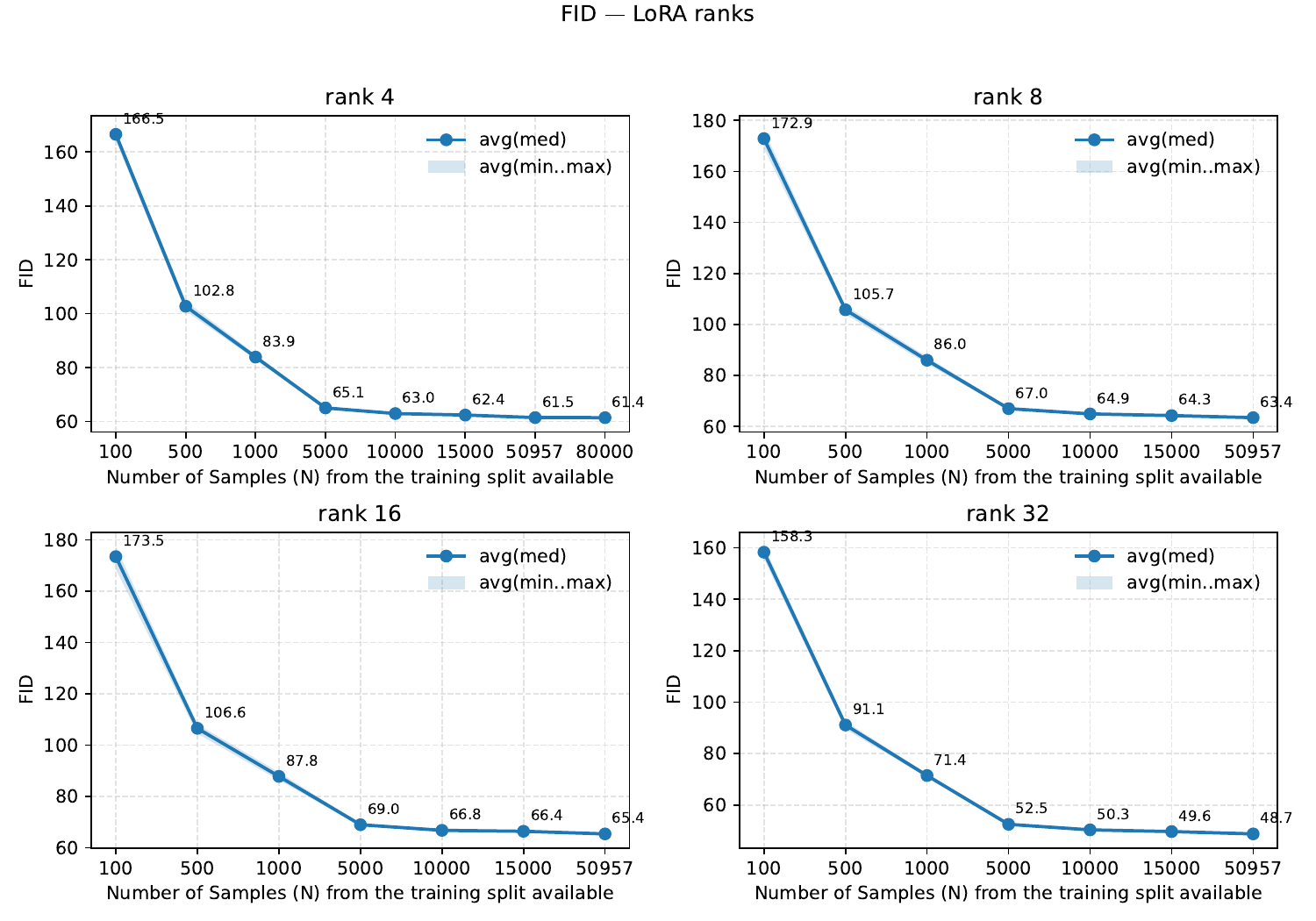}
\caption{\textbf{FID (↓) vs.\ synthetic sample size \(N\)} for the PA100K dataset across LoRA ranks \(r\in\{4,8,16,32\}\). Each point averages across five seeded populations per \(N\); we plot the avg.\ median and the avg.\ min–max envelope. Distance improves (decreases) as \(N\) grows, saturating at large \(N\); the highest rank (\(r{=}32\)) consistently achieves the lowest FID.}
\label{fig:supp_lora_FID_pa100k}
\end{figure}

\begin{figure}[t]
  \centering

\includegraphics[trim=0.25cm 0.4cm 0.25cm 1cm, clip, width=\linewidth]{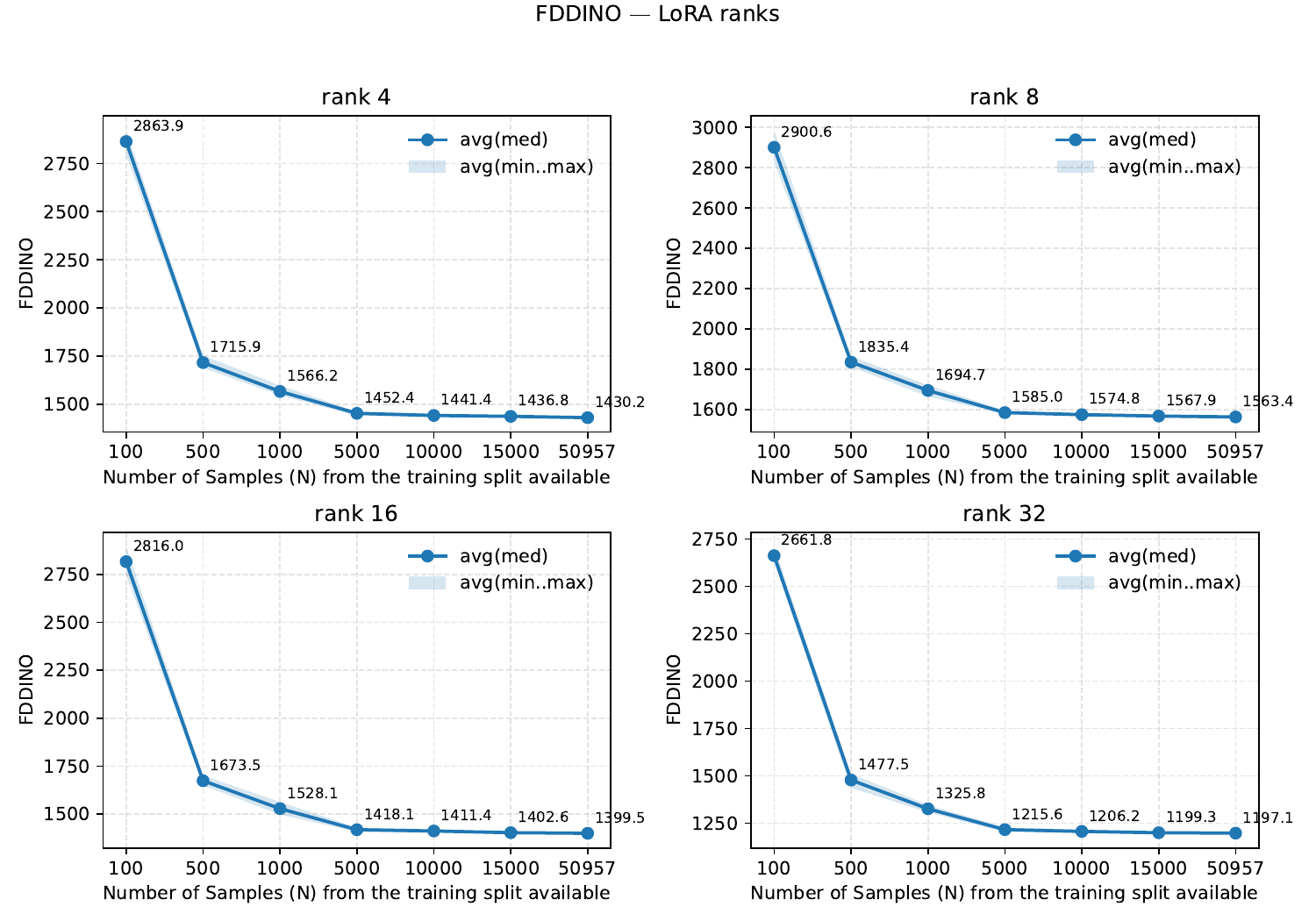}
  \caption{\textbf{FD-DINO distance (↓) vs.\ synthetic sample size \(N\)} for the PA100K dataset across LoRA ranks \(r\in\{4,8,16,32\}\).
Each curve aggregates five seeds per \(N\) (solid = avg.\ median across seeds; faint envelope = avg.\ min–max across seeds).
Reference distribution is the full training split of the dataset. Larger \(N\) reduces distance with diminishing returns; higher ranks tend to yield lower FD-DINO overall.}
\label{fig:supp_lora_FDDINO_pa100k}

\end{figure}

\begin{figure}[t]
  \centering
  
\includegraphics[trim=0.25cm 0.4cm 0.25cm 1cm, clip, width=\linewidth]{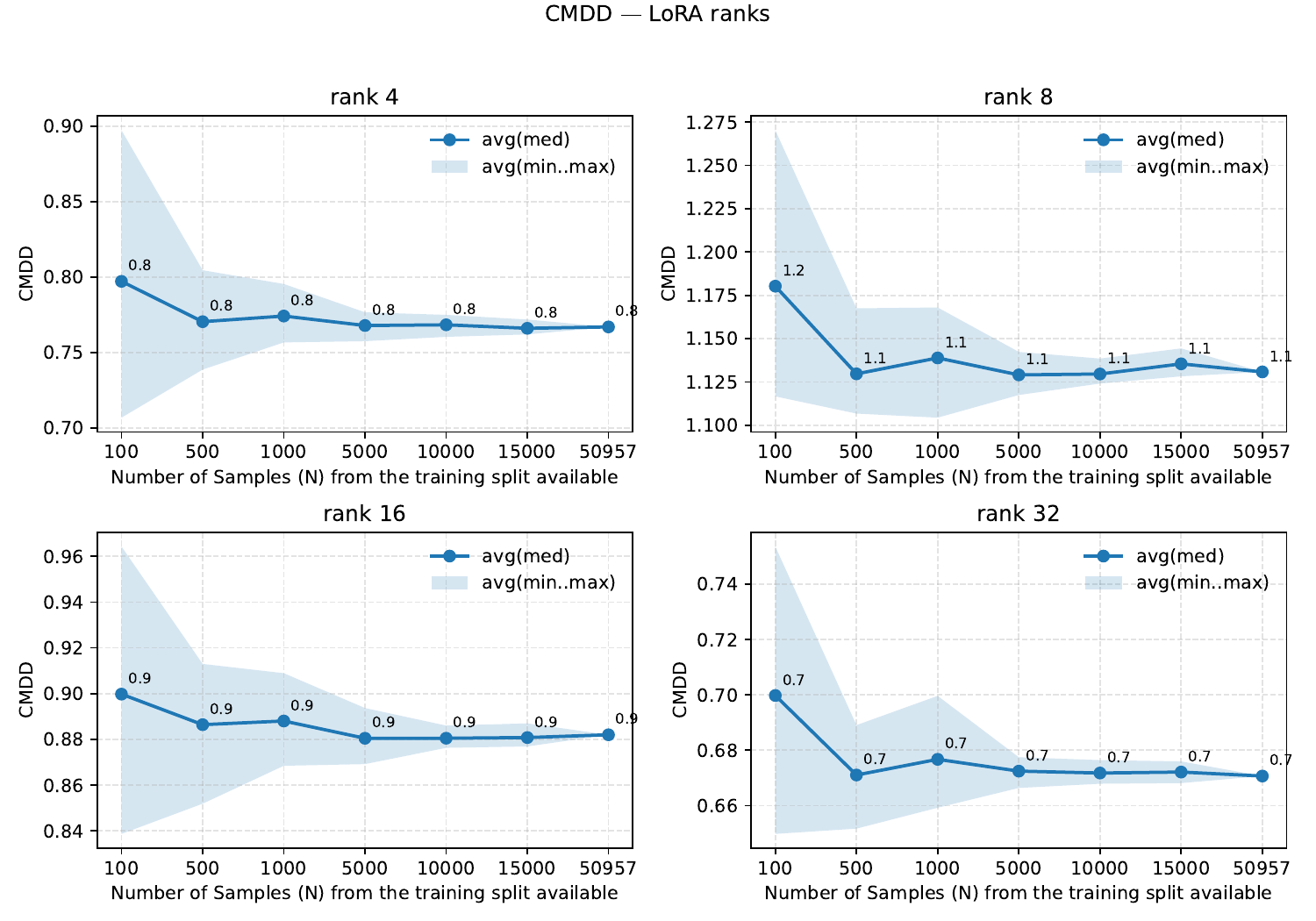}
  \caption{\textbf{CMMD (↓) vs.\ synthetic sample size \(N\)} for the PA100K dataset across LoRA ranks \(r\in\{4,8,16,32\}\).
Curves show avg.\ median and avg.\ min–max across five seeds per \(N\).
Distances decrease as \(N\) grows; higher ranks, especially \(r{=}32\), reach lower CMMD values across the range.}
\label{fig:supp_lora_CMMD_pa100k}

\end{figure}

\clearpage

\begin{table*}[t]
\centering
\caption{Filtered metrics with per-metric \emph{Value} and \emph{Pos} (rank position; 1 = best). 
\textbf{AVERAGE POS} is the average rank across the four metrics, i.e.,
\(\displaystyle \text{SUM}=\tfrac{1}{4}\big(\text{Pos}_{\text{CFID}}+\text{Pos}_{\text{FID}}+\text{Pos}_{\text{FD-DINO}}+\text{Pos}_{\text{CMMD}}\big)\) (lower is better).
\textbf{Rank} indicates the LoRA rank selected by the SUM criterion (bold = selected).}
\label{tab:supp_lora_metrics}
\setlength{\tabcolsep}{6pt}
\renewcommand{\arraystretch}{1.15}
\begin{tabular}{l *{4}{cc} cc}
\toprule
\multirow{2}{*}{\textbf{Dataset}} &
\multicolumn{2}{c}{\textbf{CFID \cite{cfid}}} &
\multicolumn{2}{c}{\textbf{FID \cite{fid}}} &
\multicolumn{2}{c}{\textbf{FD-DINO \cite{fddino}}} &
\multicolumn{2}{c}{\textbf{CMMD \cite{cmmd}}} &
\multirow{2}{*}{\textbf{AVERAGE POS \(\downarrow\)}} &
\multirow{2}{*}{\textbf{Rank}} \\
\cmidrule(lr){2-3} \cmidrule(lr){4-5} \cmidrule(lr){6-7} \cmidrule(lr){8-9}
& \textbf{Value \(\downarrow\)} & \textbf{Pos\(\downarrow\)} & \textbf{Value \(\downarrow\)} & \textbf{Pos\(\downarrow\)} &\textbf{Value \(\downarrow\)} & \textbf{Pos\(\downarrow\)} & \textbf{Value \(\downarrow\)} & \textbf{Pos\(\downarrow\)} & & \\
\midrule
\multirow{4}{*}{RAPzs}
& 71.11 & 4 & 62.61 & 4 & 1445.10 & 3 & 1.94 & 4 & 3.75 & 4 \\
& 70.19 & 2 & 45.81 & 1 & 1134.79 & 1 & 1.54 & 3 & 1.75 & \textbf{8} \\
& 70.28 & 3 & 57.63 & 2 & 1440.72 & 2 & 1.44 & 2 & 2.25 & 16 \\
& 70.17 & 1 & 61.06 & 3 & 1537.66 & 4 & 1.41 & 1 & 2.25 & 32 \\
\addlinespace
\multirow{4}{*}{RAPv1}
& 46.33 & 1 & 37.43 & 1 & 911.91  & 1 & 1.36 & 1 & 1.00 & \textbf{4} \\
& 46.73 & 2 & 39.41 & 2 & 1218.43 & 2 & 1.89 & 3 & 2.25 & 8 \\
& 47.02 & 3 & 55.36 & 4 & 1458.15 & 4 & 1.95 & 4 & 3.75 & 16 \\
& 47.05 & 4 & 50.11 & 3 & 1238.25 & 3 & 1.46 & 2 & 3.00 & 32 \\
\addlinespace
\multirow{4}{*}{RAPv2}
& 36.19 & 4 & 59.09 & 4 & 1371.07 & 2 & 1.89 & 4 & 3.50 & 4 \\
& 35.89 & 1 & 43.74 & 1 & 1125.52 & 1 & 1.50 & 3 & 1.50 & \textbf{8} \\
& 36.12 & 3 & 53.55 & 2 & 1382.29 & 3 & 1.44 & 2 & 2.50 & 16 \\
& 35.93 & 2 & 58.14 & 3 & 1484.03 & 4 & 1.36 & 1 & 2.50 & 32 \\
\addlinespace
\multirow{4}{*}{PA100K}
& 25.03 & 2 & 61.35 & 2 & 1428.91 & 3 & 1.89 & 4 & 2.75 & 4 \\
& 24.81 & 1 & 63.30 & 3 & 1563.60 & 4 & 1.50 & 3 & 2.75 & 8 \\
& 25.06 & 3 & 65.27 & 4 & 1396.94 & 2 & 1.44 & 2 & 2.75 & 16 \\
& 25.11 & 4 & 48.54 & 1 & 1194.83 & 1 & 1.36 & 1 & 1.75 & \textbf{32} \\
\addlinespace
\multirow{4}{*}{PETAzs}
& 94.93 & 4 & 67.00 & 4 & 1727.25 & 2 & 1.72 & 2 & 3.00 & 4 \\
& 93.80 & 2 & 57.16 & 2 & 1750.16 & 3 & 1.83 & 3 & 2.50 & 8 \\
& 94.43 & 3 & 61.50 & 3 & 1902.65 & 4 & 1.89 & 4 & 3.50 & 16 \\
& 92.86 & 1 & 53.46 & 1 & 1530.62 & 1 & 1.37 & 1 & 1.00 & \textbf{32} \\
\addlinespace
\multirow{4}{*}{PETA}
& 99.75 & 4 & 66.28 & 4 & 1713.72 & 2 & 1.72 & 2 & 3.00 & 4 \\
& 98.27 & 2 & 57.09 & 2 & 1753.26 & 3 & 1.85 & 3 & 2.50 & 8 \\
& 99.10 & 3 & 61.47 & 3 & 1885.71 & 4 & 1.89 & 4 & 3.50 & 16 \\
& 97.51 & 1 & 52.98 & 1 & 1516.49 & 1 & 1.36 & 1 & 1.00 & \textbf{32} \\
\bottomrule
\end{tabular}
\end{table*}

\section{Extended Text-Image Representational Similarity Metrics Analysis}
\label{sec:anexa_promptfidelity}

In this annex, we expand the analysis of text-image representational similarity metrics (BLIPScore, CLIPScore, ImageReward, HPSv2Score). We organize the evaluation into two main aspects. (i) \emph{Degree of complementation}: for the same generated images, we compare the positive probe $p_{\text{pos}}$ against the negative probe $p_{\text{neg}}$ at varying complementation ratios $\rho \in \{25\%, 50\%, 75\%, 100\%\}$. (ii) \emph{Prompt length}: focusing on the full complementation scenario ($\rho=100\%$), we extend the experiments for the other datasets. Together, these setups aim to clarify how each metric behaves when faced with stronger counterfactual probes and varying attribute list lengths. 
\paragraph{Separability analysis by complementation ratio} As detailed in \cref{tab:supp_complementation_auroc} and \cref{tab:supp_complementation_bhat}, both AUROC and Bhattacharyya distance increase monotonically as the negated fraction $\rho$ in $p_{\text{neg}}$ grows, indicating stronger separability when the negative probe diverges more significantly from the original generation prompt ($p_{\text{gen}}$). Across \emph{all} values of $\rho$, BLIPScore exhibits the strongest discriminative capacity---consistently outperforming the other scorers. Most gains occur between $\rho=50\%$ and $\rho=100\%$, where the score distributions for the positive and negative probes pull apart most clearly. While all metrics benefit from a larger $\rho$, the dominance of BLIPScore across the entire range supports its selection as the primary calibrated signal for autolabeling.
\paragraph{Prompt length} Regarding the impact of prompt length, \cref{fig:supp_promptnum_bhatt_all} illustrate the Bhattacharyya distance as a function of the number of active attributes $L$. Focusing on the fully complemented scenario ($\rho=100\%$), we observe that the discriminative capacity of \textsc{BLIPScore} consistently dominates and grows as the text complexity increases. Specifically, separability peaks and saturates around $L \approx 11{-}12$ for dense datasets like PETA, and between $L \approx 4{-}7$ for the RAP benchmarks and PA100K. The variance or drops observed at the extreme right of some curves (e.g., in PA100K for $L \ge 5$) correspond directly to the annotation limits of the datasets; notably, PA100K is among the benchmarks with the fewest annotated attributes per image, resulting in a natural scarcity of samples at those extreme lengths. Overall, the consistent upward trend prior to these dataset-specific limits confirms that the metric maintains its discriminative capacity as prompt complexity increases. To ensure statistical significance when calculating the Bhattacharyya distance, we cap the evaluation at the maximum prompt length that naturally contains at least 1,000 samples (indicated in bold in \cref{tab:supp_numberattributes_promptcons}). For specific lengths within this range that fail to meet the 1,000-sample threshold, we guarantee robust populations by randomly sampling from the immediate higher length and systematically pruning a single attribute.

\begin{table*}[htbp]
\centering
\caption{Separability of representational similarity metrics (AUROC$\uparrow$) under prompt complementation. For each dataset, we score the same images but with the positive probe $p_{\text{pos}}$ compared against $p_{\text{neg}}$ with $\rho \in \{0.25, 0.50, 0.75, 1.00\}$. Higher is better.}
\label{tab:supp_complementation_auroc}
\resizebox{\textwidth}{!}{
\renewcommand{\arraystretch}{1.10}
\begin{tabular}{l *{4}{c} *{4}{c} *{4}{c} *{4}{c}}
\toprule
& \multicolumn{4}{c}{\textbf{BLIPScore}} 
& \multicolumn{4}{c}{\textbf{CLIPScore}} 
& \multicolumn{4}{c}{\textbf{ImgReward}} 
& \multicolumn{4}{c}{\textbf{HPSv2Score}} \\
\cmidrule(lr){2-5} \cmidrule(lr){6-9} \cmidrule(lr){10-13} \cmidrule(lr){14-17}

Dataset
& $\rho=0.25$ & $\rho=0.50$ & $\rho=0.75$ & $\rho=1.00$
& $\rho=0.25$ & $\rho=0.50$ & $\rho=0.75$ & $\rho=1.00$
& $\rho=0.25$ & $\rho=0.50$ & $\rho=0.75$ & $\rho=1.00$
& $\rho=0.25$ & $\rho=0.50$ & $\rho=0.75$ & $\rho=1.00$ \\
\midrule

PA100K & \textbf{0.56} & \textbf{0.66} & \textbf{0.74} & \textbf{0.82} & 0.53 & 0.58 & 0.64 & 0.71 & 0.54 & 0.60 & 0.66 & 0.73 & 0.54 & 0.61 & 0.67 & 0.72 \\
PETA & \textbf{0.79} & \textbf{0.89} & \textbf{0.92} & \textbf{0.94} & 0.66 & 0.74 & 0.78 & 0.81 & 0.64 & 0.70 & 0.73 & 0.76 & 0.57 & 0.62 & 0.66 & 0.69 \\
PETAzs & \textbf{0.80} & \textbf{0.90} & \textbf{0.93} & \textbf{0.95} & 0.64 & 0.71 & 0.75 & 0.77 & 0.64 & 0.69 & 0.72 & 0.75 & 0.54 & 0.57 & 0.59 & 0.61 \\
RAPv1 & \textbf{0.60} & \textbf{0.72} & 0.79 & 0.86 & 0.55 & 0.61 & 0.65 & 0.71 & 0.55 & 0.61 & 0.66 & 0.71 & 0.59 & 0.71 & \textbf{0.81} & \textbf{0.90} \\
RAPv2 & \textbf{0.67} & \textbf{0.84} & \textbf{0.91} & \textbf{0.97} & 0.55 & 0.61 & 0.66 & 0.72 & 0.60 & 0.72 & 0.79 & 0.86 & 0.54 & 0.61 & 0.67 & 0.74 \\
RAPzs & \textbf{0.67} & \textbf{0.83} & \textbf{0.91} & \textbf{0.96} & 0.54 & 0.61 & 0.66 & 0.72 & 0.60 & 0.71 & 0.78 & 0.85 & 0.54 & 0.60 & 0.66 & 0.74 \\
\bottomrule
\end{tabular}%
}
\end{table*}

\begin{table*}[htbp]
\centering
\caption{Separability of representational similarity metrics (Bhattacharyya distance, BHAT$\uparrow$) under prompt complementation. For each dataset, we score the same images but with the positive probe $p_{\text{pos}}$ compared against $p_{\text{neg}}$ with $\rho \in \{0.25, 0.50, 0.75, 1.00\}$. Higher is better.}
\label{tab:supp_complementation_bhat}
\resizebox{\textwidth}{!}{
\renewcommand{\arraystretch}{1.10}
\begin{tabular}{l *{4}{c} *{4}{c} *{4}{c} *{4}{c}}
\toprule
& \multicolumn{4}{c}{\textbf{BLIPScore}} 
& \multicolumn{4}{c}{\textbf{CLIPScore}} 
& \multicolumn{4}{c}{\textbf{ImgReward}} 
& \multicolumn{4}{c}{\textbf{HPSv2Score}} \\
\cmidrule(lr){2-5} \cmidrule(lr){6-9} \cmidrule(lr){10-13} \cmidrule(lr){14-17}

Dataset
& $\rho=0.25$ & $\rho=0.50$ & $\rho=0.75$ & $\rho=1.00$
& $\rho=0.25$ & $\rho=0.50$ & $\rho=0.75$ & $\rho=1.00$
& $\rho=0.25$ & $\rho=0.50$ & $\rho=0.75$ & $\rho=1.00$
& $\rho=0.25$ & $\rho=0.50$ & $\rho=0.75$ & $\rho=1.00$ \\
\midrule

PA100K & \textbf{0.01} & \textbf{0.05} & \textbf{0.10} & \textbf{0.21} & 0.00 & 0.02 & 0.03 & 0.08 & 0.00 & 0.02 & 0.05 & 0.11 & 0.00 & 0.02 & 0.05 & 0.08 \\
PETA & \textbf{0.17} & \textbf{0.38} & \textbf{0.51} & \textbf{0.60} & 0.04 & 0.10 & 0.15 & 0.18 & 0.04 & 0.11 & 0.14 & 0.19 & 0.01 & 0.02 & 0.04 & 0.07 \\
PETAzs & \textbf{0.18} & \textbf{0.40} & \textbf{0.55} & \textbf{0.66} & 0.03 & 0.08 & 0.11 & 0.14 & 0.06 & 0.11 & 0.14 & 0.19 & 0.00 & 0.01 & 0.01 & 0.02 \\
RAPv1 & \textbf{0.02} & 0.08 & 0.16 & 0.30 & 0.01 & 0.02 & 0.04 & 0.08 & 0.00 & 0.02 & 0.04 & 0.08 & \textbf{0.02} & \textbf{0.09} & \textbf{0.19} & \textbf{0.42} \\
RAPv2 & \textbf{0.05} & \textbf{0.23} & \textbf{0.46} & \textbf{0.81} & 0.01 & 0.02 & 0.04 & 0.09 & 0.02 & 0.08 & 0.16 & 0.28 & 0.01 & 0.03 & 0.05 & 0.11 \\
RAPzs & \textbf{0.05} & \textbf{0.22} & \textbf{0.43} & \textbf{0.75} & 0.01 & 0.02 & 0.04 & 0.08 & 0.02 & 0.07 & 0.15 & 0.27 & 0.01 & 0.03 & 0.05 & 0.11 \\
\bottomrule
\end{tabular}%
}
\end{table*}

\begin{table*}[htb]
    \centering
    \caption{Distribution of images based on the number of concurrently active attributes present in their corresponding generation prompts. Bold values indicate the maximum number of simultaneous attributes that still maintain a minimum support of 1,000 image samples, ensuring sufficient data to robustly perform the Representational Similarity Analysis.}
    \label{tab:supp_numberattributes_promptcons}
    \begin{tabular}{rrrrrrr}
        \toprule
        Active Attributes & RAPzs & PA100K & RAPv1 & RAPv2 & PETA & PETAzs \\
        \midrule
        0  & 0    & 0     & 3    & 2     & 0    & 0    \\
        1  & 0    & 0     & 0    & 0     & 1    & 0    \\
        2  & 2    & 0     & 29   & 34    & 0    & 0    \\
        3  & 234  & 0     & 456  & 659   & 0    & 0    \\
        4  & 820  & 22098 & 1696 & 2619  & 0    & 0    \\
        5  & 5033 & 45318 & 8982 & 15161 & 0    & 0    \\
        6  & 4674 & 11460 & 8765 & 14114 & 0    & 0    \\
        7  & 3514 & \textbf{1077}  & 7089 & 10581 & 0    & 0    \\
        8  & \textbf{1795} & 47    & 4037 & 5281  & 0    & 0    \\
        9  & 689  & 0     & \textbf{1598} & \textbf{1878}  & 3    & 0    \\
        10 & 242  & 0     & 480  & 520   & 297  & 2    \\
        11 & 50   & 0     & 108  & 88    & 318  & 333  \\
        12 & 8    & 0     & 23   & 16    & 2376 & 313  \\
        13 & 1    & 0     & 2    & 4     & 4377 & 2342 \\
        14 & 0    & 0     & 0    & 0     & \textbf{1671} & 5412 \\
        15 & 0    & 0     & 0    & 0     & 387  & \textbf{2077} \\
        16 & 0    & 0     & 0    & 0     & 65   & 632  \\
        17 & 0    & 0     & 0    & 0     & 4    & 119  \\
        18 & 0    & 0     & 0    & 0     & 1    & 10   \\
        19 & 0    & 0     & 0    & 0     & 0    & 1    \\
        \bottomrule
    \end{tabular}
\end{table*}

\begin{figure*}[t]
  \centering
  
  \subfloat[PA100K\label{fig:supp_promptnum_bhatt_pa100k}]{%
      \includegraphics[trim=0.25cm 0.3cm 0.25cm 0.0cm, clip, width=0.48\linewidth]{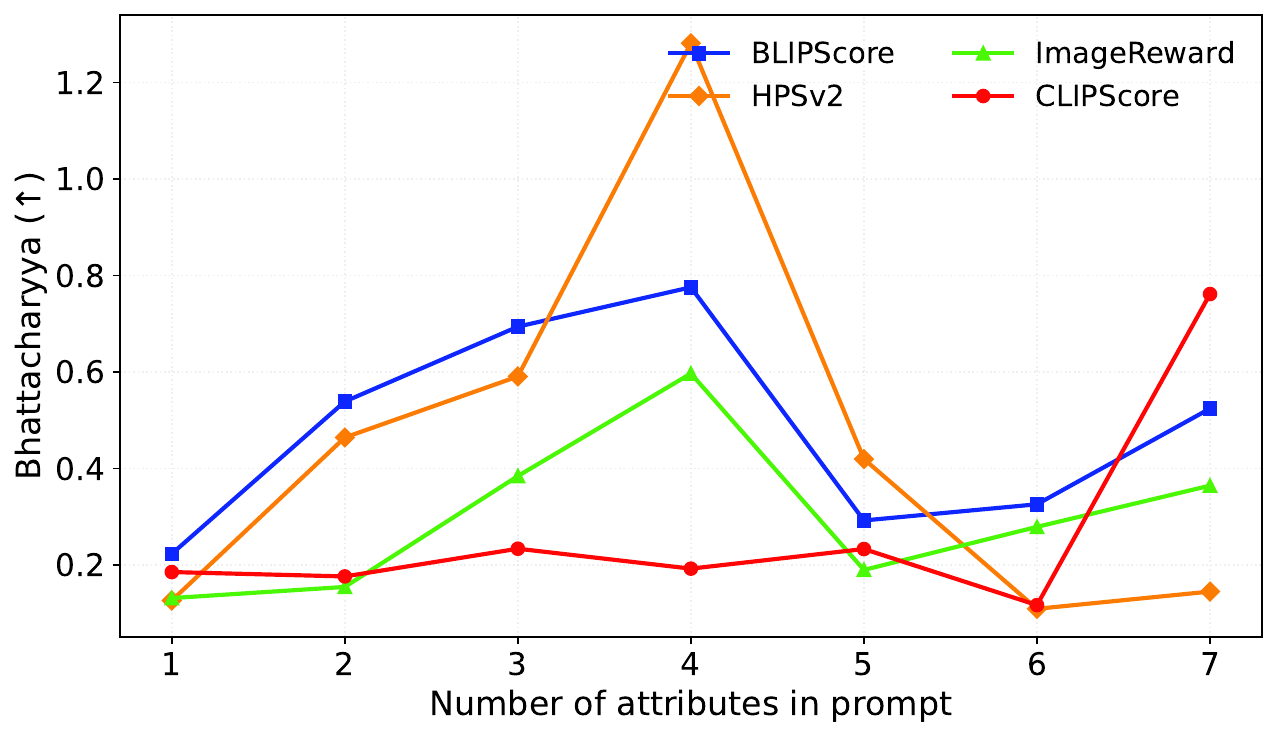}%
  }\hfill
  \subfloat[RAPv1\label{fig:supp_promptnum_bhatt_rapv1}]{%
      \includegraphics[trim=0.25cm 0.3cm 0.25cm 0.0cm, clip, width=0.48\linewidth]{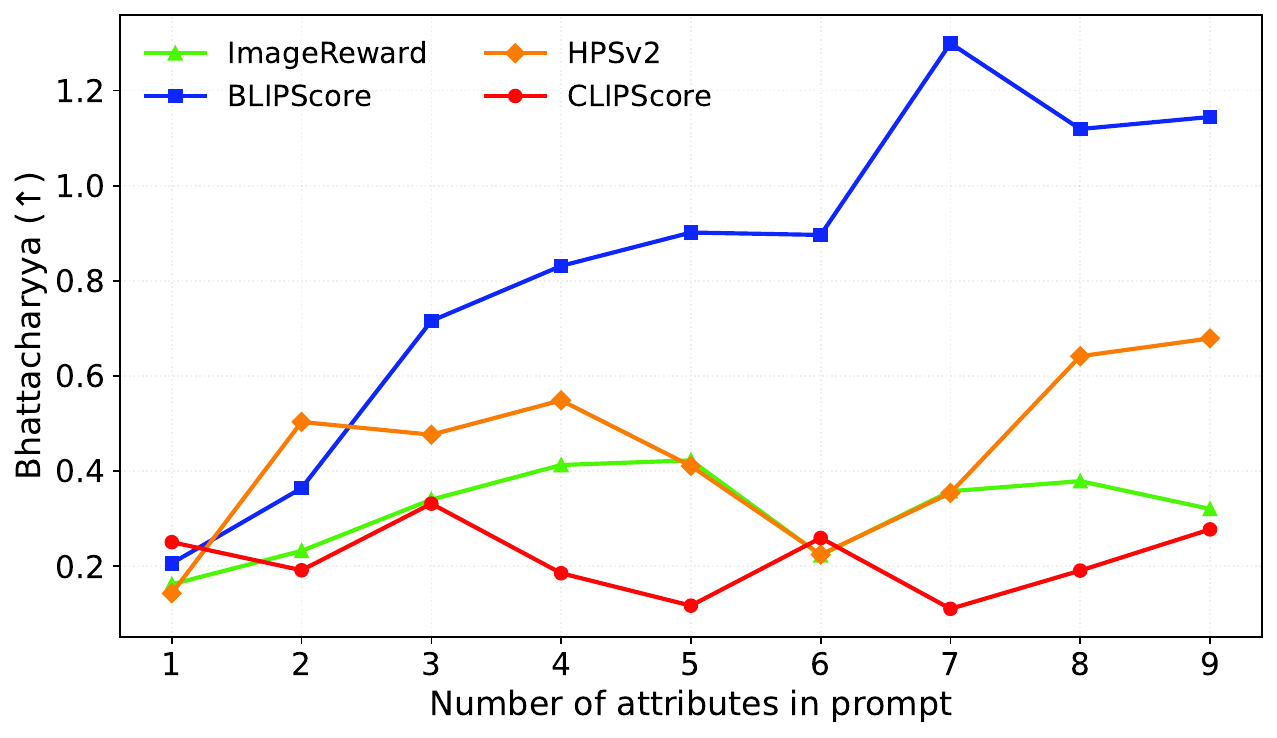}%
  }

  \vspace{0.3cm} 
  
  \subfloat[RAPv2\label{fig:supp_promptnum_bhatt_rapv2}]{%
      \includegraphics[trim=0.25cm 0.3cm 0.25cm 0.0cm, clip, width=0.48\linewidth]{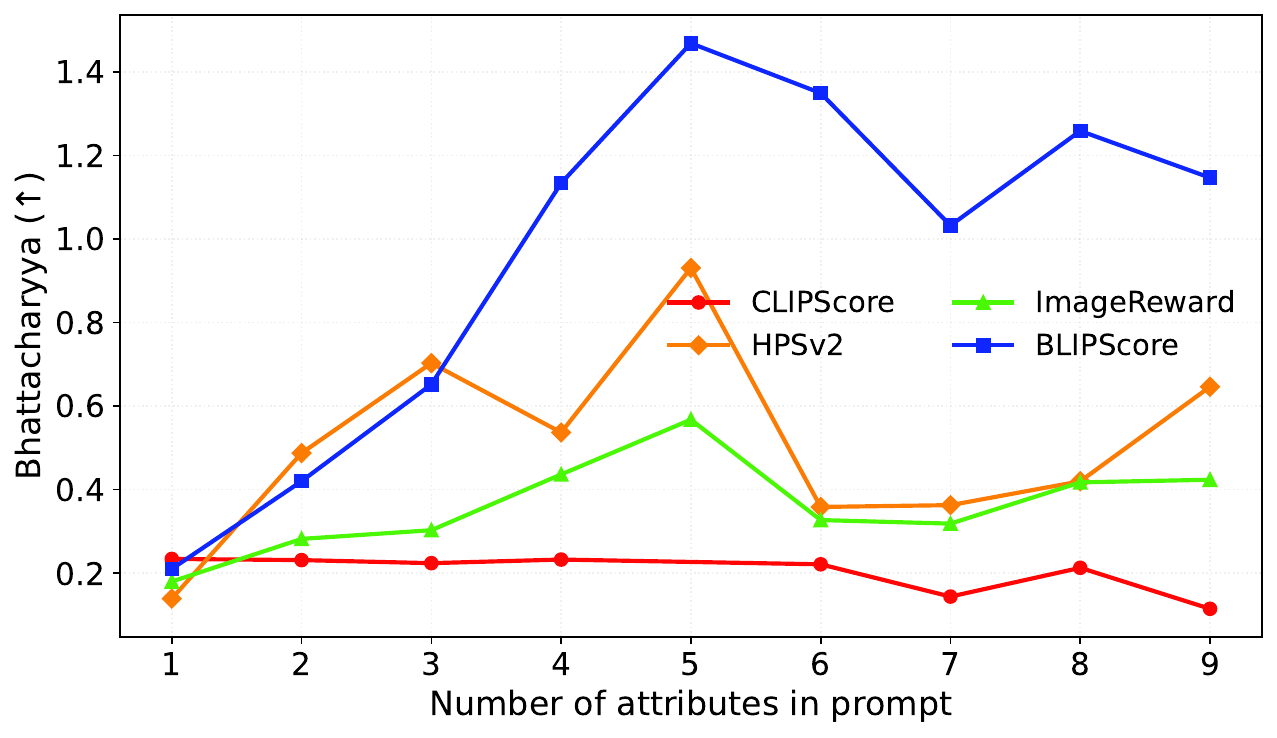}%
  }\hfill
  \subfloat[RAPzs\label{fig:supp_promptnum_bhatt_rapzs}]{%
      \includegraphics[trim=0.25cm 0.3cm 0.25cm 0.0cm, clip, width=0.48\linewidth]{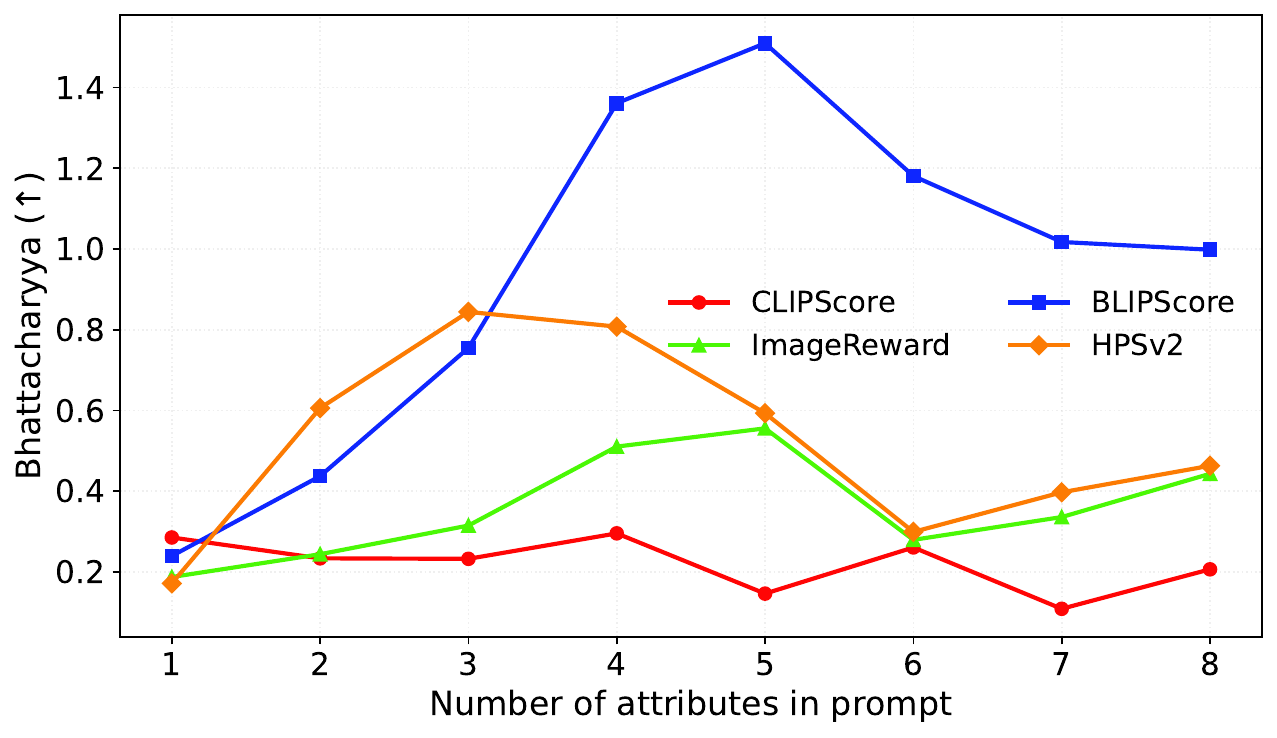}%
  }

  \vspace{0.3cm} 

  \subfloat[PETA\label{fig:supp_promptnum_bhatt_peta}]{%
      \includegraphics[trim=0.25cm 0.3cm 0.25cm 0.0cm, clip, width=0.48\linewidth]{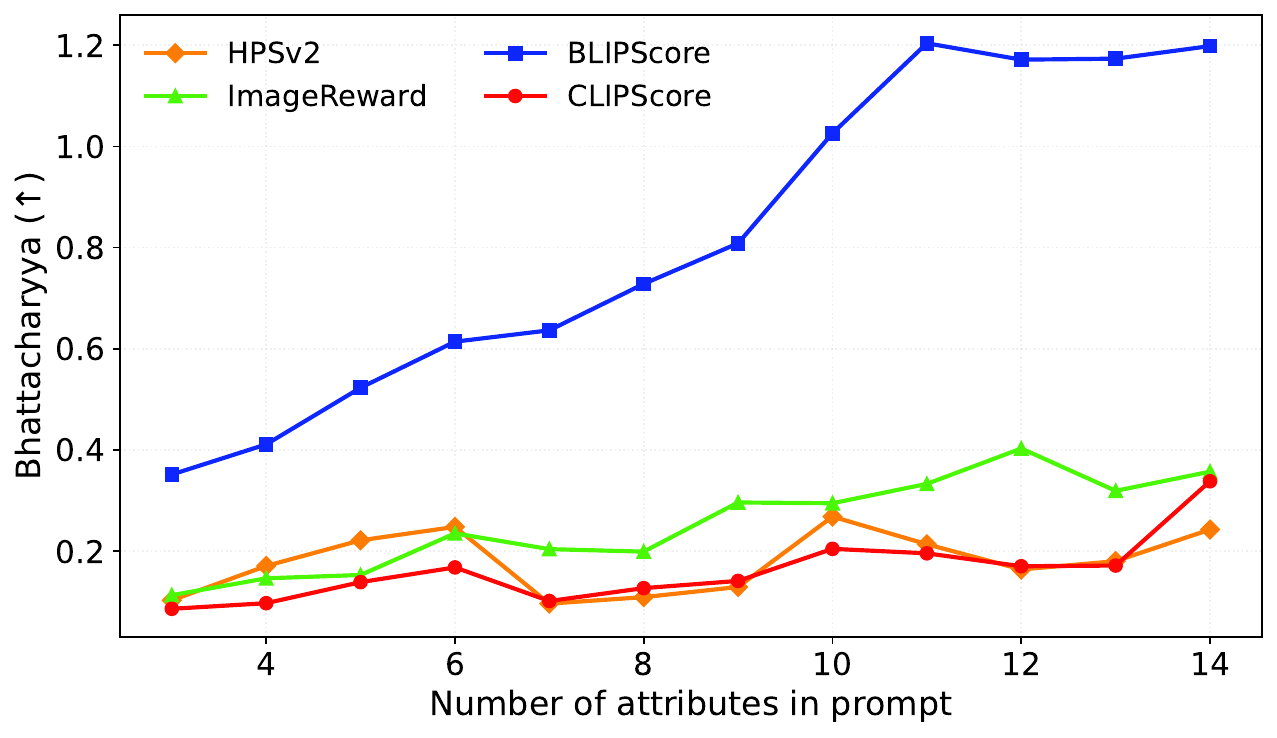}%
  }
\caption{\textbf{BHATT ($\uparrow$) vs.\ number of attributes in the prompt.} For the same images, we score the positive probes $p_{\text{pos}}$ against the fully complemented negative probes $p_{\text{neg}}$ with $\rho = 100\%$ and report the Bhattacharyya distance for four scorers as the prompt length increases.}
  \label{fig:supp_promptnum_bhatt_all}
  
\end{figure*}

\clearpage

\clearpage

\section{Extended Analysis of BLIPScore Distributions}
\label{sec:anexa_distributionsBlipScore}

As introduced in Section IV-C of the main manuscript, the effectiveness of our Bayesian autolabeling framework relies heavily on the underlying stability of the representational similarity scores. While the main text visualizes this behavior exclusively for the PETAzs dataset, this section provides the comprehensive empirical distributions for all remaining benchmarks (RAPv1, RAPv2, RAPzs, PA100K, and PETA).

Following the established experimental protocol, we plot the density of the \textsc{BLIPScore} evaluated on the testing splits for each dataset. As illustrated in 
\cref{fig:supp_promptnum_disBlipscore_test_pa100k}, 
\cref{fig:supp_promptnum_disBlipscore_test_rapv1},
\cref{fig:supp_promptnum_disBlipscore_test_rapv2}, 
\cref{fig:supp_promptnum_disBlipscore_test_rapzs}, 
\cref{fig:supp_promptnum_disBlipscore_test_peta} a pervasive structural trend is evident across all domains: the score distributions exhibit unimodal and smooth characteristics with clear separability between the positive probes ($p_{\text{pos}}$) and their fully complemented counterparts ($p_{\text{neg}}$ with $\rho=100\%$). Crucially, demonstrating this robust separability directly on test data proves that the metric's discriminative capacity generalizes effectively beyond the training calibration phase. This widespread consistency empirically validates our choice of a two-component Gaussian formulation for the generative likelihoods, confirming it as a dataset-agnostic solution for isolating valid attributes from semantic noise.

\begin{figure}[h]
  \centering
\includegraphics[trim=0.25cm 0.20cm 0.25cm 0.25cm, clip, width=0.8\linewidth]{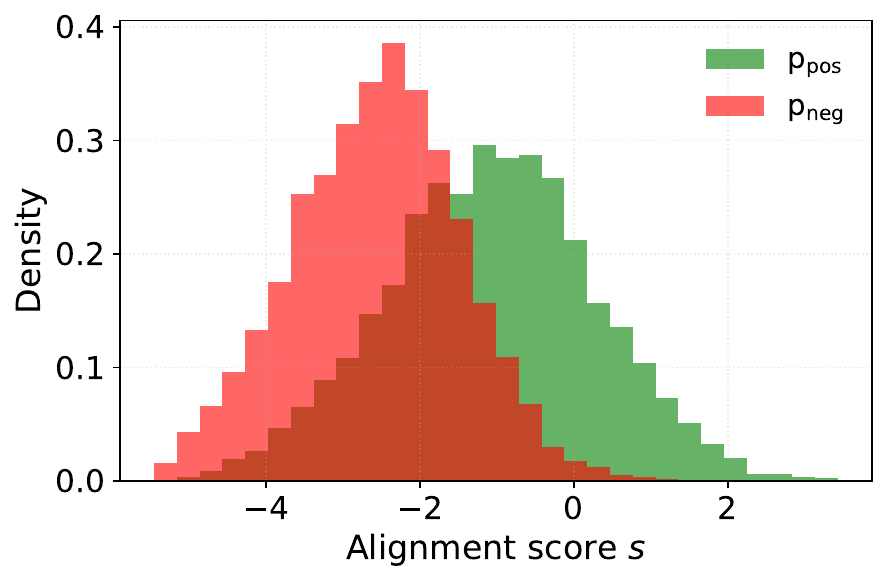}
  \caption{\textbf{PA100K:} BLIPScore density distribution on the testing set. The metric maintains strong separability between the positive ($p_{\text{pos}}$) and fully complemented negative ($p_{\text{neg}}$) probes.}
\label{fig:supp_promptnum_disBlipscore_test_pa100k}

\end{figure}

\begin{figure}[h]
  \centering
\includegraphics[trim=0.25cm 0.20cm 0.25cm 0.25cm, clip, width=0.8\linewidth]{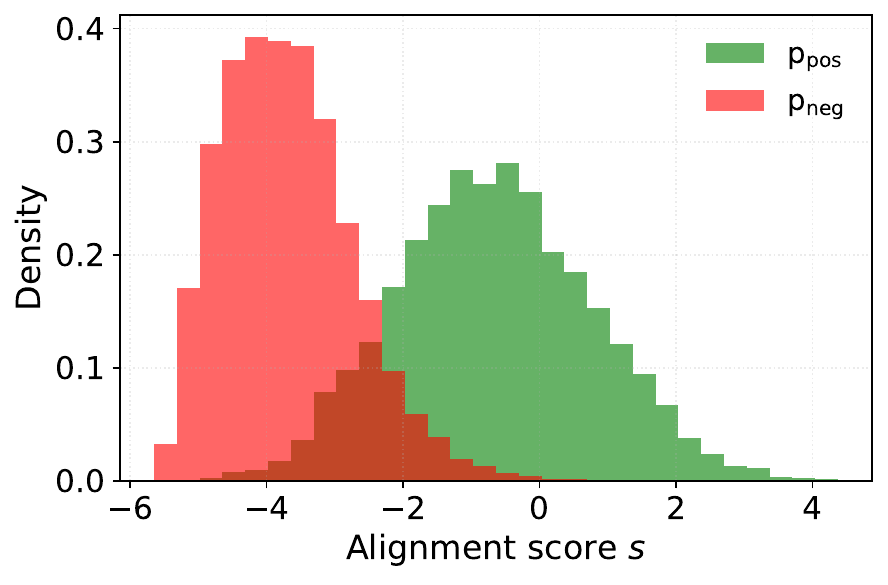}
  \caption{\textbf{RAPv1:} BLIPScore density distribution on the testing set. The metric maintains strong separability between the positive ($p_{\text{pos}}$) and fully complemented negative ($p_{\text{neg}}$) probes.}
\label{fig:supp_promptnum_disBlipscore_test_rapv1}

\end{figure}

\begin{figure}[h]
  \centering
\includegraphics[trim=0.25cm 0.20cm 0.25cm 0.25cm, clip, width=0.8\linewidth]{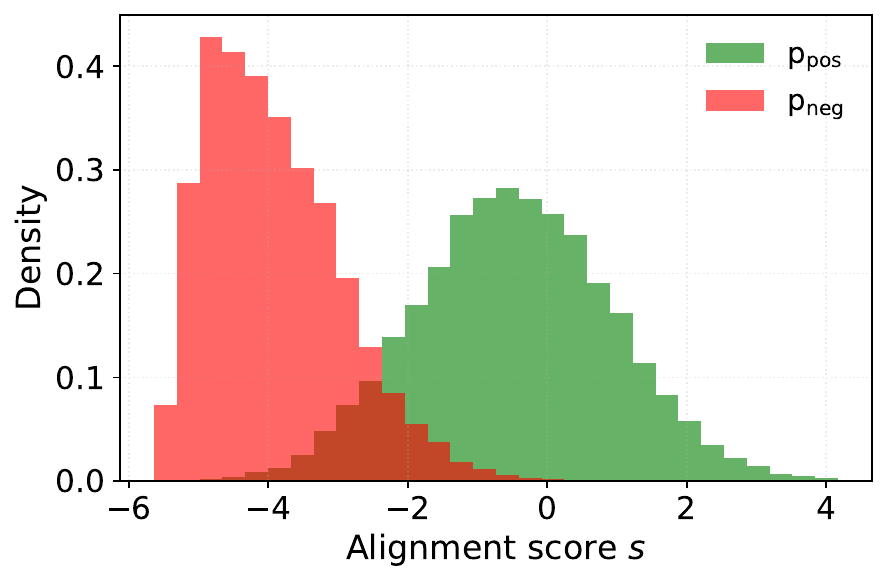}
  \caption{\textbf{RAPv2:} BLIPScore density distribution on the testing set. The metric maintains strong separability between the positive ($p_{\text{pos}}$) and fully complemented negative ($p_{\text{neg}}$) probes.}
\label{fig:supp_promptnum_disBlipscore_test_rapv2}

\end{figure}

\begin{figure}[h]
  \centering
\includegraphics[trim=0.25cm 0.20cm 0.25cm 0.25cm, clip, width=0.8\linewidth]{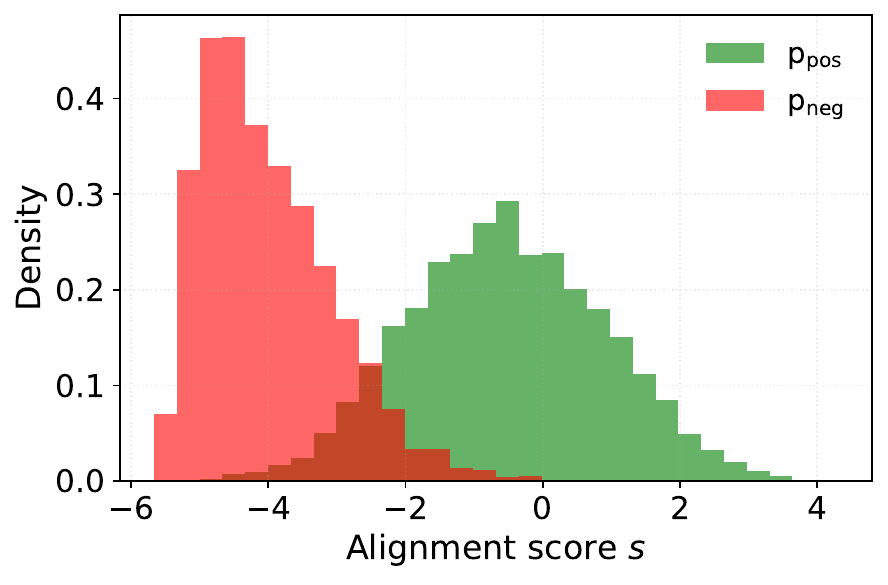}
  \caption{\textbf{RAPzs:} BLIPScore density distribution on the testing set. The metric maintains strong separability between the positive ($p_{\text{pos}}$) and fully complemented negative ($p_{\text{neg}}$) probes.}
\label{fig:supp_promptnum_disBlipscore_test_rapzs}

\end{figure}

\begin{figure}[h]
  \centering
\includegraphics[trim=0.25cm 0.20cm 0.25cm 0.25cm, clip, width=0.8\linewidth]{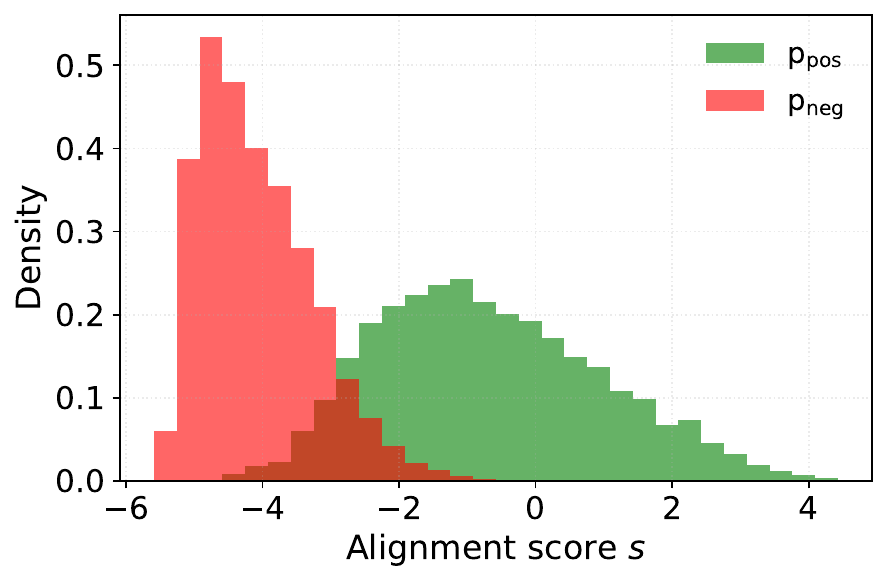}
  \caption{\textbf{PETA:} BLIPScore density distribution on the testing set. The metric maintains strong separability between the positive ($p_{\text{pos}}$) and fully complemented negative ($p_{\text{neg}}$) probes.}
\label{fig:supp_promptnum_disBlipscore_test_peta}

\end{figure}

\clearpage
\section{Detailed Autolabeling Accuracy per Attribute}
\label{sec:anexa_attributesimproved}

This appendix details how our score-based autolabeler behaves across different datasets and individual attributes. We first recap the labeling protocol: alignment scores extracted from positive probes ($p_{\text{pos}}$) versus fully complemented negative probes ($p_{\text{neg}}^{(100\%)}$) are used to train a 1D Bayesian classifier on \emph{real} images. This calibrated model is then frozen and applied to \emph{synthetic} images to generate pseudo-labels using a standard MAP decision threshold ($\tau=0.50$). 

To understand the practical capabilities and limitations of this vision-language grounding, we analyze the per-attribute accuracy of the generated pseudo-labels against the ground-truth annotations. As shown in the following tables, the inherent difficulty of fine-grained attribute recognition in generative contexts yields a moderate average attribute accuracy ($\approx 45{-}54\%$). The complete attribute-level test accuracy breakdown for all datasets is provided in \cref{tab:attr_ma_threecols_pa_peta_petazs} and \cref{tab:attr_ma_rap_threecols}.

To illustrate the decision behavior of our Bayesian filter, \cref{fig:qualitative_analysis_pa100k}, \cref{fig:qualitative_analysis_rapv2}, \cref{fig:qualitative_analysis_rapv1}, \cref{fig:qualitative_analysis_rapzs}, \cref{fig:qualitative_analysis_peta}, \cref{fig:qualitative_analysis_petazs} presents representative qualitative results across our datasets. We showcase four distinct scenarios: (a) successful attribute alignment with positive prompts, (b) successful neutralization with negative prompts, (c) failure on positive prompts (false negatives), and (d) failure on negative prompts (false positives). This comparison clarifies the filter's behavior, highlighting its effectiveness in identifying high-confidence attributes while also identifying specific cases where ambiguous visual features lead to misclassification.

\begin{table*}[t]
\centering
\caption{Per-attribute accuracy (\%) on the test split for \textbf{PA100K}, \textbf{PETA}, and \textbf{PETAzs}. Each dataset is shown in its own column; within each, we list (Attribute, Accuracy).}
\label{tab:attr_ma_threecols_pa_peta_petazs}

\scriptsize 

\setlength{\tabcolsep}{2pt} 
\renewcommand{\arraystretch}{1.05}

\begin{minipage}[t]{0.19\textwidth}
\centering \textbf{PA100K}\\[2pt]
\begin{tabular}{@{}lr@{}}
\toprule
\textbf{Attribute} & \textbf{\%} \\
\midrule
LongCoat & 95.74\\
LowerStripe & 90.30\\
Skirt\&Dress & 80.57\\
HoldObjectsInFront & 80.46\\
Shorts & 80.06\\
Back & 77.70\\
UpperSplice & 77.61\\
Trousers & 77.04\\
Backpack & 76.76\\
ShortSleeve & 73.88\\
Female & 73.65\\
Front & 72.55\\
HandBag & 72.13\\
ShoulderBag & 71.46\\
LongSleeve & 71.14\\
Side & 70.68\\
LowerPattern & 70.65\\
UpperLogo & 65.36\\
UpperPlaid & 53.93\\
\bottomrule
\end{tabular}
\end{minipage}\hfill 
\begin{minipage}[t]{0.19\textwidth}
\centering \textbf{PETA}\\[2pt]
\begin{tabular}{@{}lr@{}}
\toprule
\textbf{Attribute} & \textbf{\%} \\
\midrule
footwearYellow & 99.08\\
hairOrange & 98.26\\
lowerBodyPlaid & 98.03\\
lowerBodyOrange & 97.85\\
lowerBodyGreen & 97.41\\
footwearRed & 96.97\\
lowerBodyRed & 96.79\\
footwearPink & 96.63\\
hairGreen & 96.50\\
footwearPurple & 96.02\\
lowerBodyThinStripes & 95.99\\
footwearGreen & 95.58\\
upperBodyRed & 95.24\\
footwearOrange & 95.24\\
upperBodyLogo & 94.97\\
lowerBodyPurple & 94.97\\
accessoryKerchief & 94.78\\
hairBlue & 94.68\\
lowerBodyBrown & 94.46\\
upperBodyGreen & 94.30\\
upperBodyTshirt & 94.29\\
hairPink & 94.12\\
hairWhite & 94.09\\
accessoryHeadphone & 93.93\\
footwearBrown & 93.92\\
carryingOther & 93.82\\
upperBodyYellow & 93.58\\
carryingBackpack & 93.41\\
upperBodyJacket & 93.27\\
upperBodyOrange & 93.21\\
lowerBodyShorts & 92.96\\
lowerBodyYellow & 92.82\\
footwearBlue & 92.79\\
carryingNothing & 92.52\\
footwearSneaker & 92.47\\
upperBodyWhite & 92.47\\
hairBrown & 92.44\\
footwearSandals & 92.35\\
upperBodyPlaid & 92.26\\
lowerBodySuits & 92.25\\
hairBald & 92.03\\
upperBodyThinStripes & 92.03\\
lowerBodyLongSkirt & 91.91\\
carryingPlasticBags & 91.91\\
footwearWhite & 91.79\\
footwearShoes & 91.76\\
personalMale & 91.71\\
\bottomrule
\end{tabular}
\end{minipage}\hfill 
\begin{minipage}[t]{0.19\textwidth}
\centering \textbf{PETA}\\[2pt]
\begin{tabular}{@{}lr@{}}
\toprule
\textbf{Attribute} & \textbf{\%} \\
\midrule
hairShort & 91.67\\
upperBodyPink & 91.66\\
hairLong & 91.64\\
upperBodyPurple & 91.64\\
carryingBabyBuggy & 91.64\\
lowerBodyWhite & 91.52\\
lowerBodyJeans & 91.44\\
lowerBodyGrey & 91.28\\
footwearBlack & 91.05\\
hairYellow & 90.99\\
personalFemale & 90.92\\
carryingFolder & 90.89\\
lowerBodyTrousers & 90.84\\
hairBlack & 90.84\\
upperBodyBlack & 90.80\\
footwearLeatherShoes & 90.66\\
footwearGrey & 90.60\\
upperBodyGrey & 90.57\\
lowerBodyHotPants & 90.57\\
upperBodyOther & 90.47\\
hairGrey & 90.43\\
upperBodyBlue & 90.29\\
lowerBodyCapri & 90.28\\
lowerBodyBlack & 90.26\\
lowerBodyFormal & 90.22\\
footwearStocking & 90.02\\
lowerBodyShortSkirt & 89.94\\
accessorySunglasses & 89.27\\
lowerBodyBlue & 89.24\\
upperBodyFormal & 89.23\\
accessoryMuffler & 89.14\\
carryingShoppingTro & 87.96\\
upperBodyVNeck & 87.50\\
accessoryHat & 87.40\\
accessoryNothing & 87.16\\
carryingSuitcase & 86.62\\
carryingLuggageCase & 85.90\\
lowerBodyPink & 85.48\\
upperBodyCasual & 83.95\\
upperBodyBrown & 83.87\\
carryingUmbrella & 83.49\\
accessoryHairBand & 83.22\\
carryingMessengerBag & 82.77\\
lowerBodyCasual & 74.78\\
hairRed & 72.29\\
hairPurple & 47.70\\
\bottomrule
\end{tabular}
\end{minipage}\hfill 
\begin{minipage}[t]{0.19\textwidth}
\centering \textbf{PETAzs }\\[2pt]
\begin{tabular}{@{}lr@{}}
\toprule
\textbf{Attribute} & \textbf{\%} \\
\midrule
footwearOrange & 100.00\\
hairPurple & 99.59\\
lowerBodyRed & 98.71\\
lowerBodyPurple & 98.47\\
footwearPurple & 98.34\\
lowerBodyYellow & 98.32\\
lowerBodyGreen & 98.25\\
footwearYellow & 98.11\\
hairWhite & 98.06\\
footwearRed & 97.96\\
upperBodyThickStripes & 97.82\\
lowerBodyPlaid & 97.36\\
footwearBlue & 97.35\\
hairBald & 97.25\\
lowerBodyOrange & 97.06\\
footwearGreen & 96.88\\
footwearPink & 96.88\\
lowerBodyShorts & 96.82\\
accessoryKerchief & 96.75\\
lowerBodyCasual & 96.63\\
upperBodyGreen & 96.49\\
upperBodyPurple & 96.42\\
carryingNothing & 96.26\\
upperBodyOrange & 96.00\\
hairBlue & 95.95\\
upperBodyRed & 95.69\\
upperBodyLogo & 95.64\\
lowerBodyThinStripes & 95.62\\
accessorySunglasses & 95.24\\
upperBodyBlack & 95.20\\
personalMale & 94.99\\
footwearShoes & 94.95\\
upperBodyPink & 94.58\\
hairShort & 94.58\\
accessoryMuffler & 94.43\\
carryingUmbrella & 94.39\\
lowerBodyBlue & 94.36\\
lowerBodyJeans & 94.34\\
upperBodyTshirt & 94.21\\
upperBodyLongSleeve & 94.10\\
lowerBodySuits & 94.07\\
lowerBodyWhite & 94.07\\
hairGrey & 94.03\\
lowerBodyLongSkirt & 94.01\\
upperBodyOther & 93.99\\
hairBlack & 93.95\\
footwearBlack & 93.88\\
upperBodyShortSleeve & 93.82\\
lowerBodyTrousers & 93.70\\
upperBodySweater & 93.41\\
\bottomrule
\end{tabular}
\end{minipage}\hfill 
\begin{minipage}[t]{0.19\textwidth}
\centering \textbf{PETAzs}\\[2pt]
\begin{tabular}{@{}lr@{}}
\toprule
\textbf{Attribute} & \textbf{\%} \\
\midrule
hairLong & 93.34\\
upperBodyBlue & 93.33\\
upperBodyYellow & 93.27\\
footwearLeatherShoes & 93.26\\
upperBodyJacket & 93.20\\
carryingOther & 93.19\\
footwearWhite & 93.07\\
carryingBackpack & 93.06\\
lowerBodyBlack & 93.04\\
footwearSandals & 92.98\\
carryingPlasticBags & 92.88\\
upperBodyCasual & 92.81\\
upperBodyFormal & 92.50\\
upperBodyWhite & 92.45\\
personalFemale & 92.45\\
upperBodySuit & 92.35\\
lowerBodyShortSkirt & 92.12\\
footwearSneaker & 92.02\\
footwearStocking & 92.01\\
carryingBabyBuggy & 91.87\\
footwearGrey & 91.85\\
accessoryHat & 91.78\\
hairPink & 91.67\\
upperBodyThinStripes & 91.58\\
accessoryNothing & 91.41\\
upperBodyGrey & 91.03\\
lowerBodyGrey & 90.81\\
carryingMessengerBag & 90.73\\
lowerBodyFormal & 90.71\\
lowerBodyCapri & 90.51\\
hairOrange & 90.14\\
footwearBrown & 90.09\\
carryingFolder & 89.92\\
hairYellow & 89.41\\
accessoryHeadphone & 89.07\\
carryingSuitcase & 89.04\\
upperBodyVNeck & 89.00\\
upperBodyPlaid & 88.63\\
hairRed & 88.57\\
lowerBodyBrown & 87.53\\
footwearBoots & 87.51\\
hairGreen & 86.96\\
upperBodyBrown & 86.49\\
accessoryHairBand & 86.11\\
lowerBodyHotPants & 84.77\\
upperBodyNoSleeve & 84.75\\
carryingLuggageCase & 84.65\\
lowerBodyPink & 79.78\\
carryingShoppingTro & 72.92\\
hairBrown & 64.27\\
\bottomrule
\end{tabular}
\end{minipage}

\end{table*}

\begin{table*}[t]
\centering
\caption{Per-attribute accuracy (\%) on the test split for \textbf{RAPv1}, \textbf{RAPv2}, and \textbf{RAPzs}. Each dataset occupies one column; within each, we list (Attribute, Accuracy).}
\label{tab:attr_ma_rap_threecols}
\small
\setlength{\tabcolsep}{4pt}
\renewcommand{\arraystretch}{1.02}
\begin{tabular}{@{}c c c@{}}

\begin{minipage}[t]{0.32\textwidth}
\centering \textbf{RAPv1}\\[2pt]
\begin{tabular}{@{}p{0.68\linewidth}r@{}}
\toprule
\textbf{Attribute} & \textbf{\%} \\
\midrule
ub-ShortSleeve & 99.71\\
action-CarrybyHand & 96.10\\
action-Calling & 95.33\\
hs-BaldHead & 95.08\\
ub-Vest & 94.43\\
hs-BlackHair & 93.99\\
shoes-Casual & 93.70\\
attach-HandTrunk & 93.57\\
ub-SuitUp & 93.47\\
hs-Glasses & 93.08\\
hs-Muffler & 91.95\\
ub-Sweater & 91.83\\
shoes-Sport & 91.46\\
lb-Jeans & 91.11\\
ub-Tight & 90.84\\
attach-PaperBag & 90.55\\
ub-TShirt & 90.50\\
shoes-Boots & 89.85\\
ub-Shirt & 89.74\\
hs-LongHair & 89.56\\
attach-PlasticBag & 88.99\\
lb-LongTrousers & 88.88\\
attach-HandBag & 88.27\\
ub-Jacket & 88.09\\
Female & 87.95\\
lb-TightTrousers & 87.94\\
attach-Backpack & 87.80\\
action-Talking & 87.76\\
shoes-Cloth & 87.74\\
action-CarrybyArm & 87.74\\
action-Gathering & 87.52\\
shoes-Leather & 87.14\\
lb-Skirt & 86.74\\
lb-Dress & 86.32\\
attach-SingleShoulderBag & 85.94\\
lb-ShortSkirt & 84.44\\
attach-Other & 84.16\\
action-Pusing & 82.78\\
ub-Cotton & 81.82\\
action-Pulling & 81.45\\
hs-Hat & 81.29\\
attach-Box & 77.62\\
action-Holding & 74.79\\
\bottomrule
\end{tabular}
\end{minipage}

&

\begin{minipage}[t]{0.32\textwidth}
\centering \textbf{RAPv2}\\[2pt]
\begin{tabular}{@{}p{0.68\linewidth}r@{}}
\toprule
\textbf{Attribute} & \textbf{\%} \\
\midrule
action-Pulling & 98.28\\
ub-Vest & 96.57\\
ub-ShortSleeve & 95.80\\
shoes-Other & 95.53\\
hs-Glasses & 95.13\\
shoes-Casual & 95.04\\
hs-BaldHead & 94.17\\
lb-Dress & 94.11\\
lb-Jeans & 93.72\\
ub-Sweater & 93.44\\
ub-SuitUp & 93.12\\
ub-Others & 93.04\\
ub-TShirt & 92.74\\
attachment-HandBag & 92.71\\
ub-Tight & 92.60\\
shoes-Sports & 92.37\\
hs-LongHair & 92.15\\
attachment-ShoulderBag & 91.97\\
action-Calling & 91.76\\
attachment-Box & 91.68\\
attachment-Backpack & 91.63\\
ub-Shirt & 91.42\\
attachment-Other & 91.40\\
attachment-HandTrunk & 91.15\\
shoes-Boots & 91.09\\
hs-BlackHair & 90.88\\
lb-LongTrousers & 90.74\\
Femal & 90.68\\
attachment-PlasticBag & 90.55\\
attachment-PaperBag & 90.32\\
action-Holding & 89.90\\
shoes-Cloth & 89.85\\
action-Gathering & 89.70\\
action-Pushing & 89.48\\
action-Talking & 88.78\\
shoes-Leather & 88.72\\
ub-Jacket & 88.65\\
lb-ShortSkirt & 88.26\\
lb-TightTrousers & 88.01\\
lb-Skirt & 87.04\\
action-CarryingByHand & 86.97\\
action-CarryingByArm & 86.82\\
hs-Hat & 85.21\\
ub-Cotton & 81.64\\
action-Other & 81.21\\
\bottomrule
\end{tabular}
\end{minipage}

&

\begin{minipage}[t]{0.32\textwidth}
\centering \textbf{RAPzs}\\[2pt]
\begin{tabular}{@{}p{0.68\linewidth}r@{}}
\toprule
\textbf{Attribute} & \textbf{\%} \\
\midrule
action-Pulling & 100.00\\
action-Other & 100.00\\
ub-ShortSleeve & 97.22\\
attachment-Other & 97.15\\
lb-Dress & 97.12\\
ub-Vest & 96.68\\
hs-Glasses & 96.33\\
attachment-HandTrunk & 96.23\\
hs-BaldHead & 95.65\\
ub-Others & 95.65\\
ub-SuitUp & 95.09\\
lb-Jeans & 94.36\\
attachment-PaperBag & 93.92\\
ub-Shirt & 93.72\\
attachment-HandBag & 93.57\\
attachment-Box & 93.51\\
hs-LongHair & 93.45\\
attachment-Backpack & 93.36\\
ub-Sweater & 93.28\\
shoes-Casual & 93.20\\
shoes-Boots & 93.08\\
attachment-PlasticBag & 92.74\\
hs-BlackHair & 92.52\\
Femal & 92.38\\
ub-TShirt & 92.34\\
action-CarryingByHand & 92.29\\
action-Gathering & 91.90\\
action-Calling & 91.87\\
action-Talking & 91.81\\
attachment-ShoulderBag & 91.46\\
shoes-Sports & 91.07\\
lb-TightTrousers & 89.87\\
action-Holding & 89.76\\
shoes-Leather & 89.50\\
lb-LongTrousers & 88.95\\
action-Pushing & 88.08\\
ub-Tight & 87.99\\
ub-Jacket & 87.66\\
shoes-Cloth & 86.44\\
lb-ShortSkirt & 86.36\\
action-CarryingByArm & 85.87\\
hs-Hat & 85.05\\
ub-Cotton & 84.66\\
lb-Skirt & 84.47\\
shoes-Other & 48.08\\
\bottomrule
\end{tabular}
\end{minipage}

\\
\end{tabular}
\end{table*}

\clearpage

\begin{figure*}[t]
  \centering
  
  \subfloat[\footnotesize Success: High-confidence alignment with the positive prompt and negative prompt]{%
      \includegraphics[trim=0.25cm 0.2cm 0.25cm 0.0cm, clip, width=0.48\linewidth]{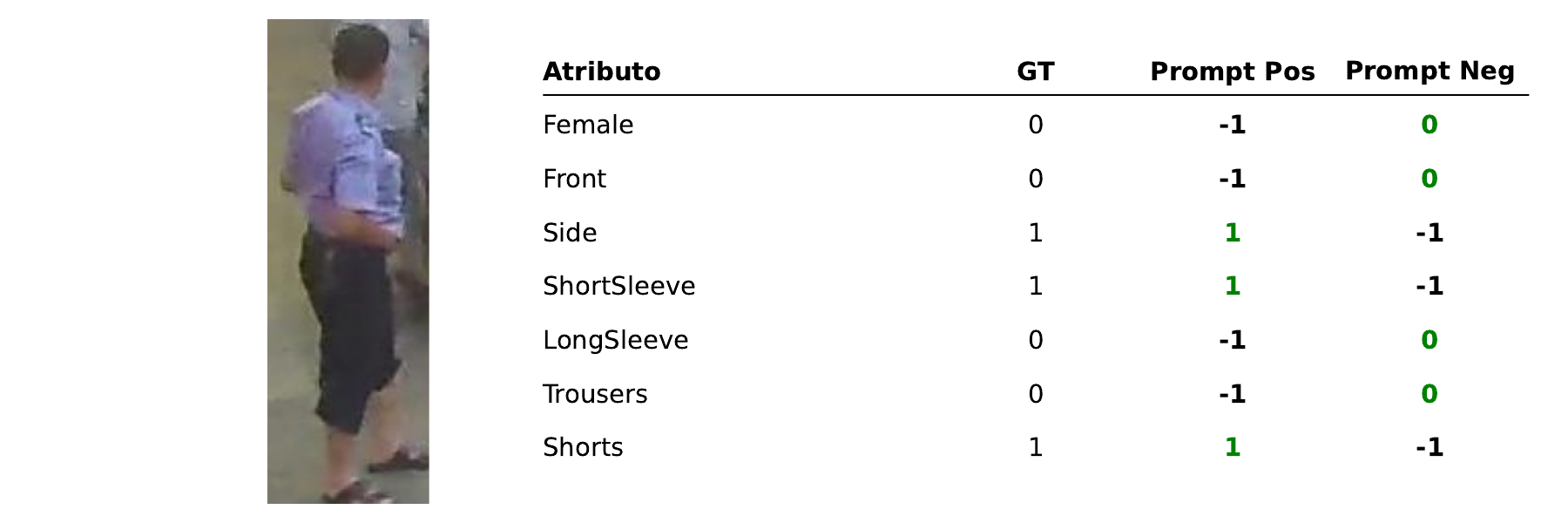}%
  }\hfill
  \subfloat[\footnotesize Success: High-confidence alignment with the positive prompt and negative prompt]{%
      \includegraphics[trim=0.25cm 0.2cm 0.25cm 0.0cm, clip, width=0.48\linewidth]{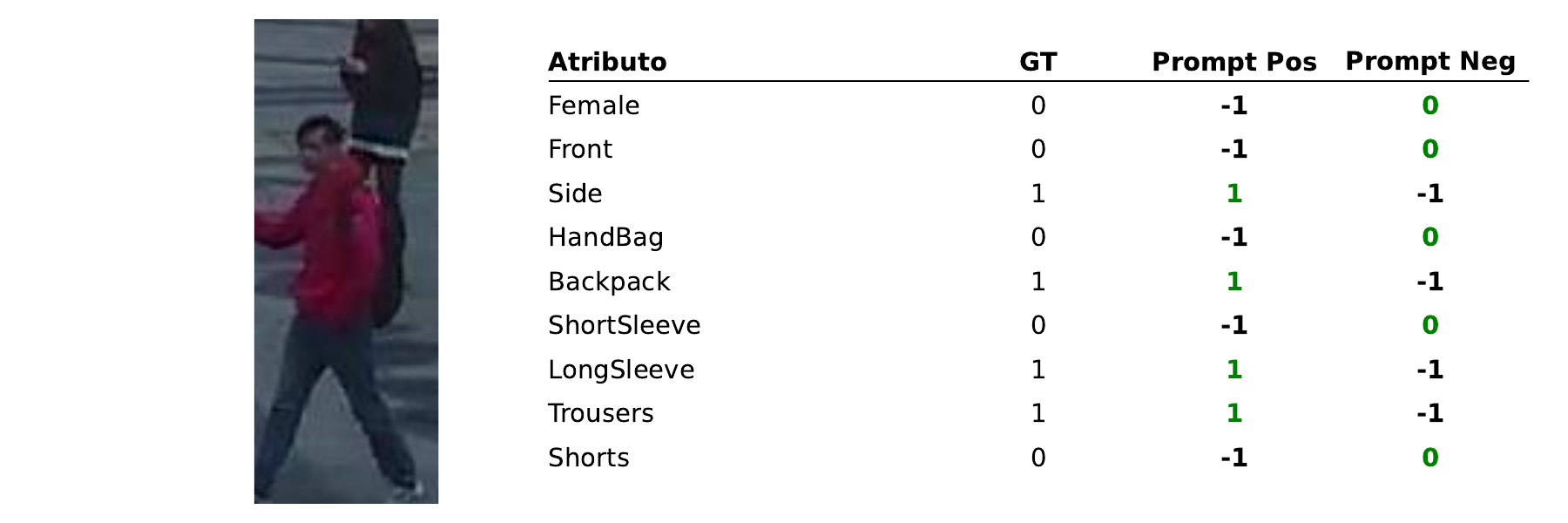}%
  }

  \subfloat[\footnotesize Failure: High-confidence alignment with the positive prompt]{%
      \includegraphics[trim=0.25cm 0.2cm 0.25cm 0.0cm, clip, width=0.48\linewidth]{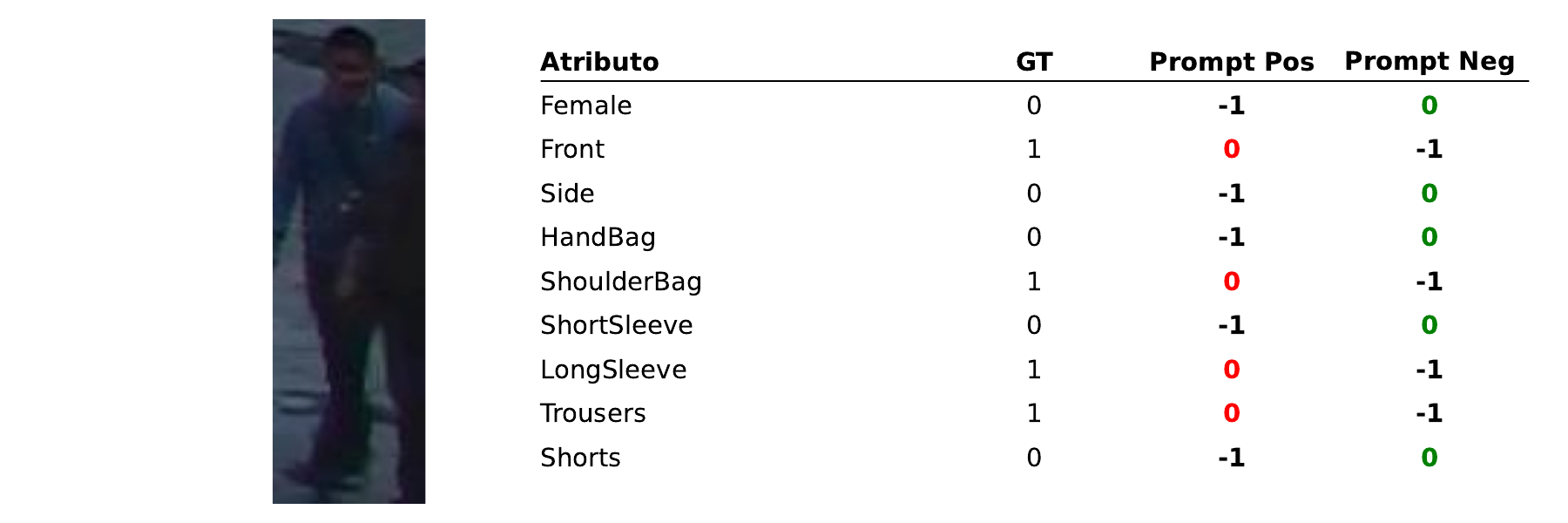}%
  }\hfill
  \subfloat[\footnotesize Failure: High-confidence alignment with the negative prompt]{%
      \includegraphics[trim=0.25cm 0.2cm 0.25cm 0.0cm, clip, width=0.48\linewidth]{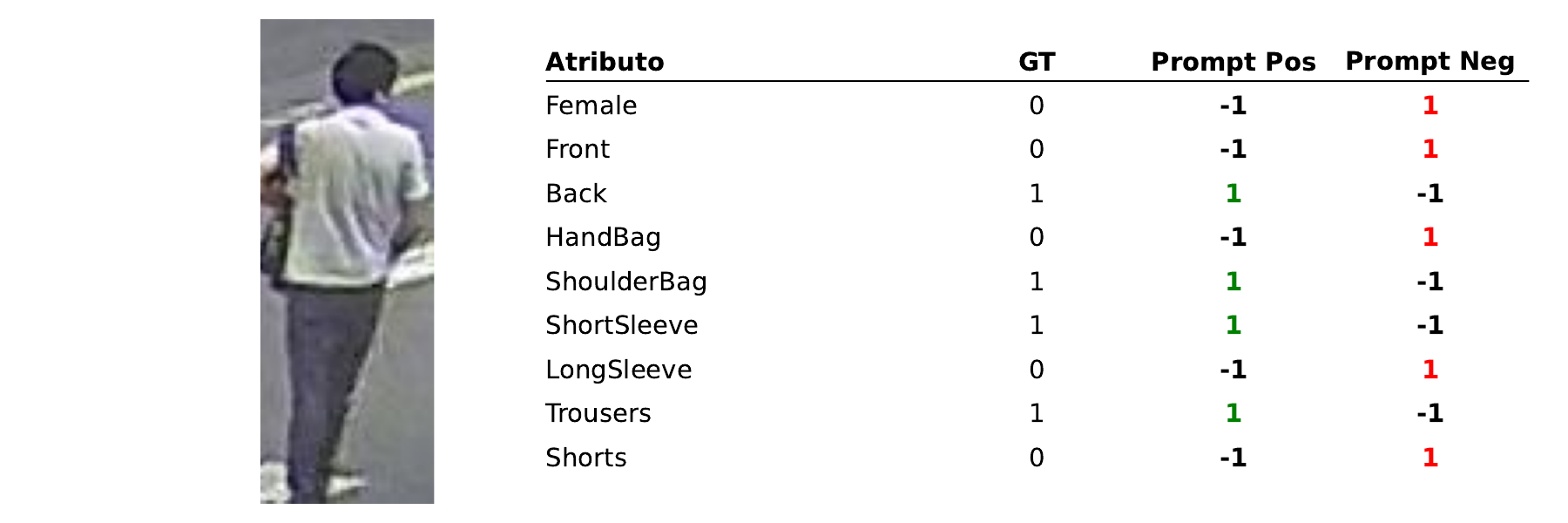}%
  }
  \caption{Qualitative performance analysis of the ReSAGE-PAR Bayesian autolabeling framework on the PA100K dataset. We showcase representative success cases (a, b) and failure cases (c, d) to illustrate the filter's decision behavior. For each example, we display the ground truth (GT), alongside the predictions derived from the positive prompt (Prompt Pos) and the negative prompt (Prompt Neg). In both prompt representations, attributes not explicitly present at the prompt are masked as $-1$.}
\label{fig:qualitative_analysis_pa100k}
\end{figure*}

\begin{figure*}[t]
  \centering

 \subfloat[\footnotesize Success: High-confidence alignment with the positive prompt and negative prompt]{%
      \includegraphics[trim=0.25cm 0.2cm 0.25cm 0.0cm, clip, width=0.48\linewidth]{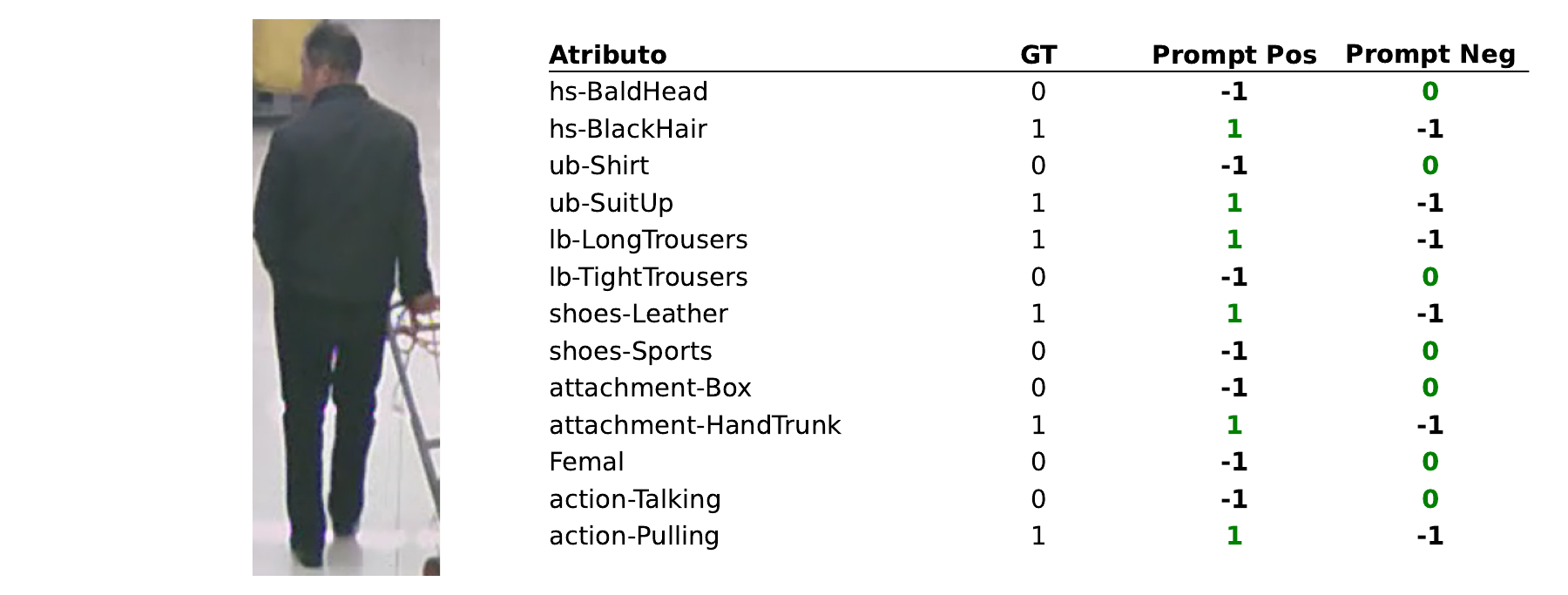}%
  }\hfill
  \subfloat[\footnotesize Success: High-confidence alignment with the positive prompt and negative prompt]{%
      \includegraphics[trim=0.25cm 0.2cm 0.25cm 0.0cm, clip, width=0.48\linewidth]{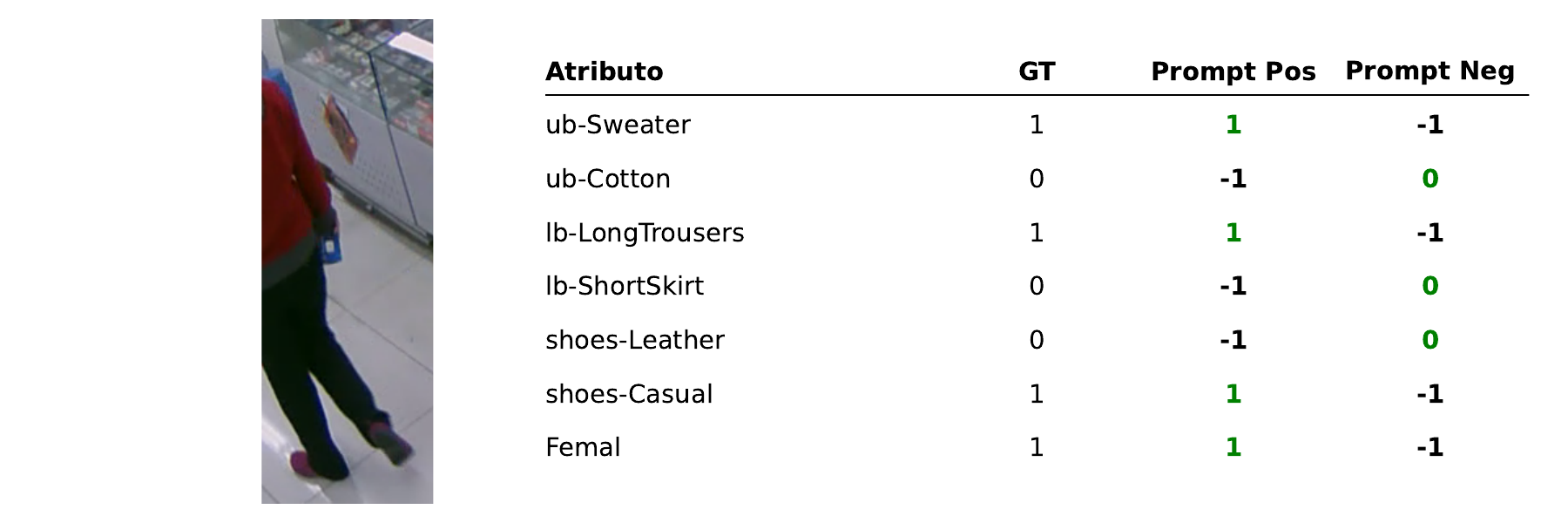}%
  }

    \subfloat[\footnotesize Failure: High-confidence alignment with the positive prompt]{%
      \includegraphics[trim=0.25cm 0.2cm 0.25cm 0.0cm, clip, width=0.48\linewidth]{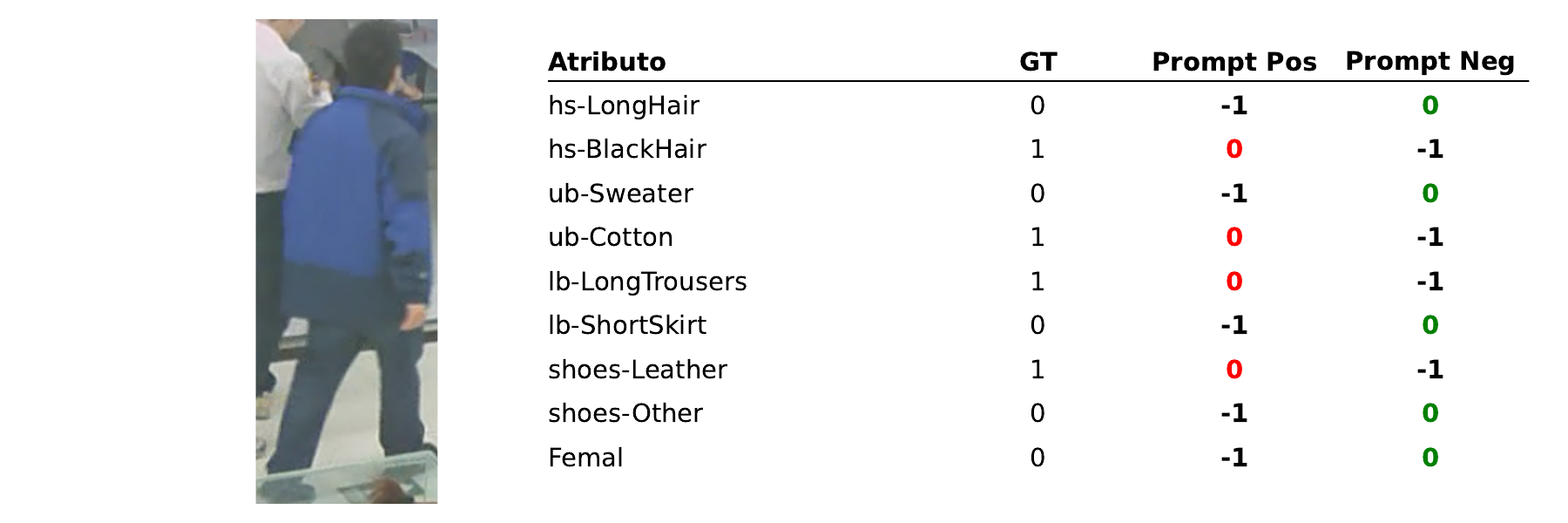}%
  }\hfill
  \subfloat[\footnotesize Failure: High-confidence alignment with the negative prompt]{%
      \includegraphics[trim=0.25cm 0.2cm 0.25cm 0.0cm, clip, width=0.48\linewidth]{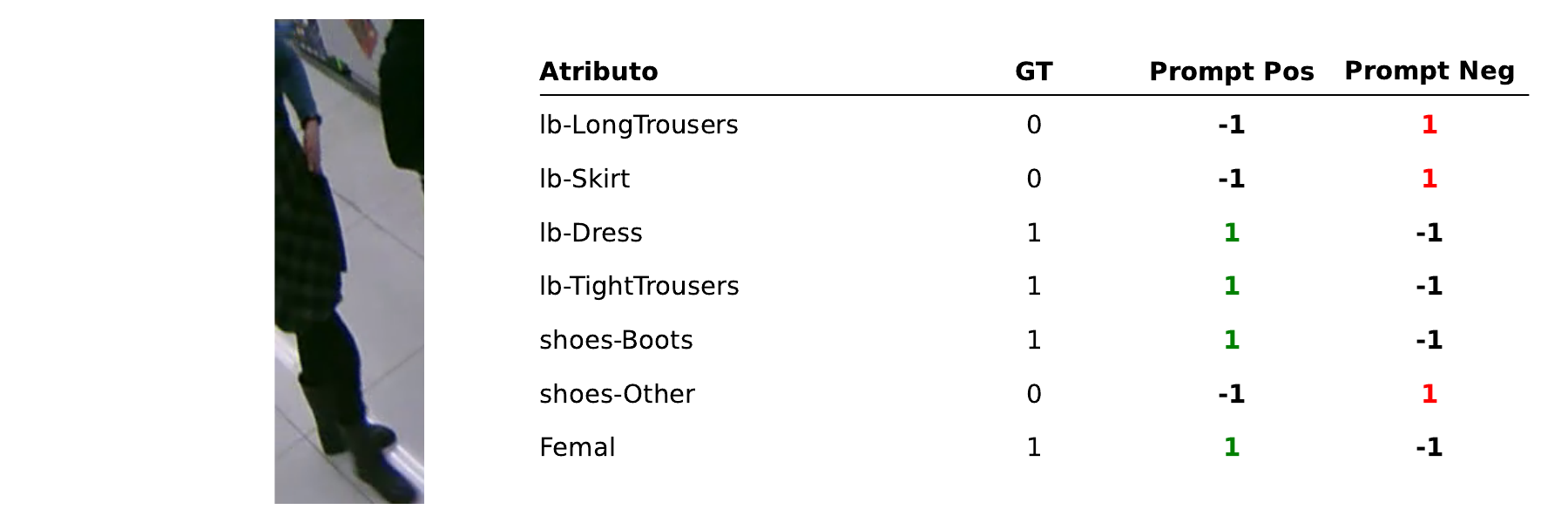}%
  }
  \caption{Qualitative performance analysis of the ReSAGE-PAR Bayesian autolabeling framework on the RAPv2 dataset. We showcase representative success cases (a, b) and failure cases (c, d) to illustrate the filter's decision behavior. For each example, we display the ground truth (GT), alongside the predictions derived from the positive prompt (Prompt Pos) and the negative prompt (Prompt Neg). In both prompt representations, attributes not explicitly present at the prompt are masked as $-1$.}
  \label{fig:qualitative_analysis_rapv2}
\end{figure*}

\begin{figure*}[t]
  \centering
  
 \subfloat[\footnotesize Success: High-confidence alignment with the positive prompt and negative prompt]{%
      \includegraphics[trim=0.25cm 0.2cm 0.25cm 0.0cm, clip, width=0.48\linewidth]{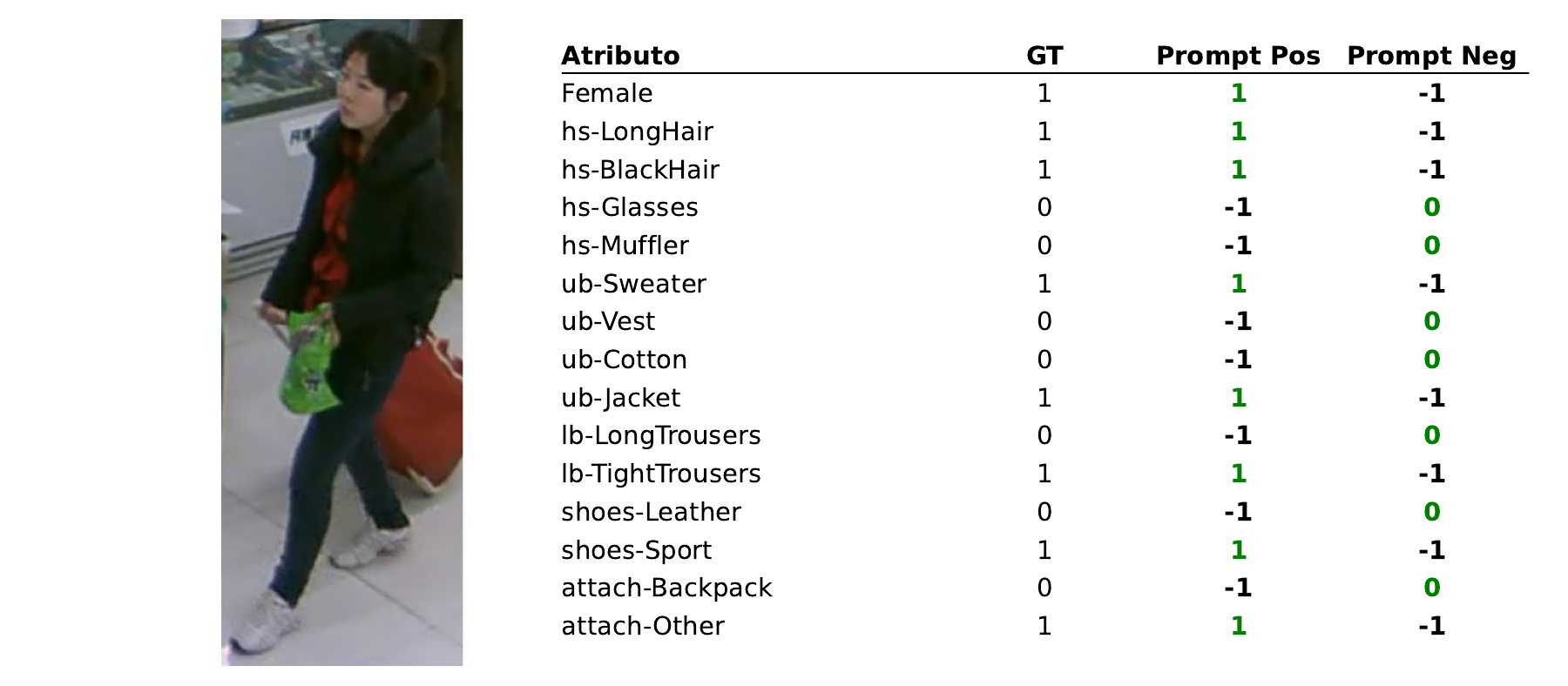}%
  }\hfill
  \subfloat[\footnotesize Success: High-confidence alignment with the positive prompt and negative prompt]{%
      \includegraphics[trim=0.25cm 0.2cm 0.25cm 0.0cm, clip, width=0.48\linewidth]{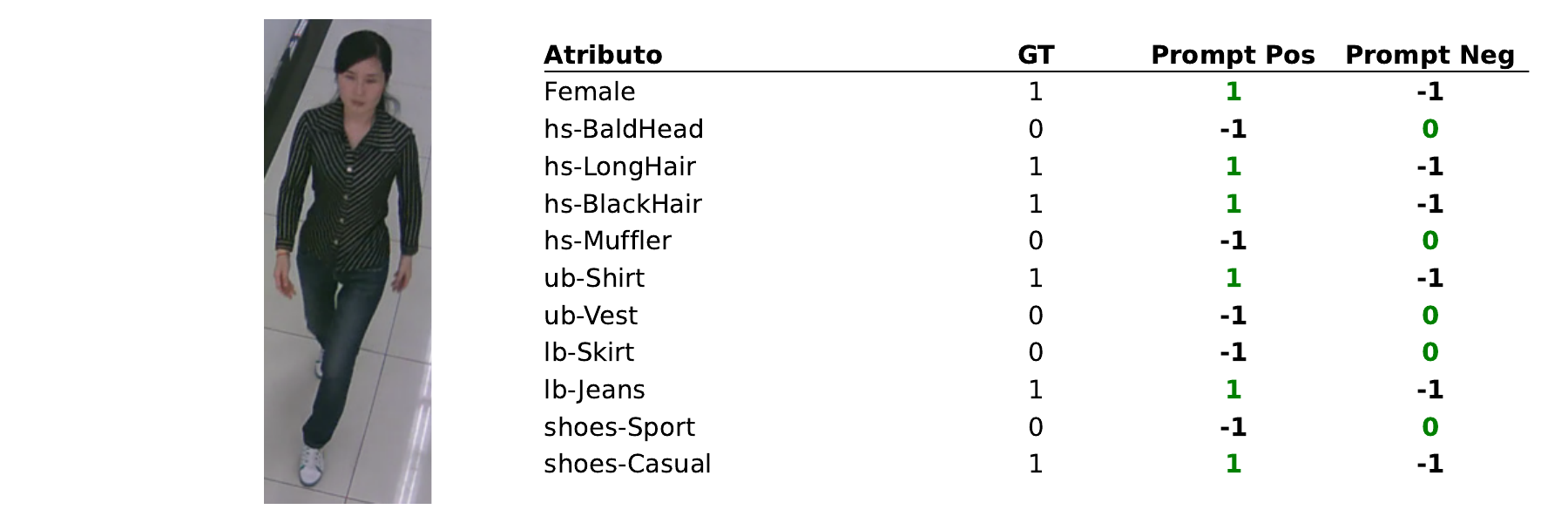}%
  }

    \subfloat[\footnotesize Failure: High-confidence alignment with the positive prompt]{%
      \includegraphics[trim=0.25cm 0.2cm 0.25cm 0.0cm, clip, width=0.48\linewidth]{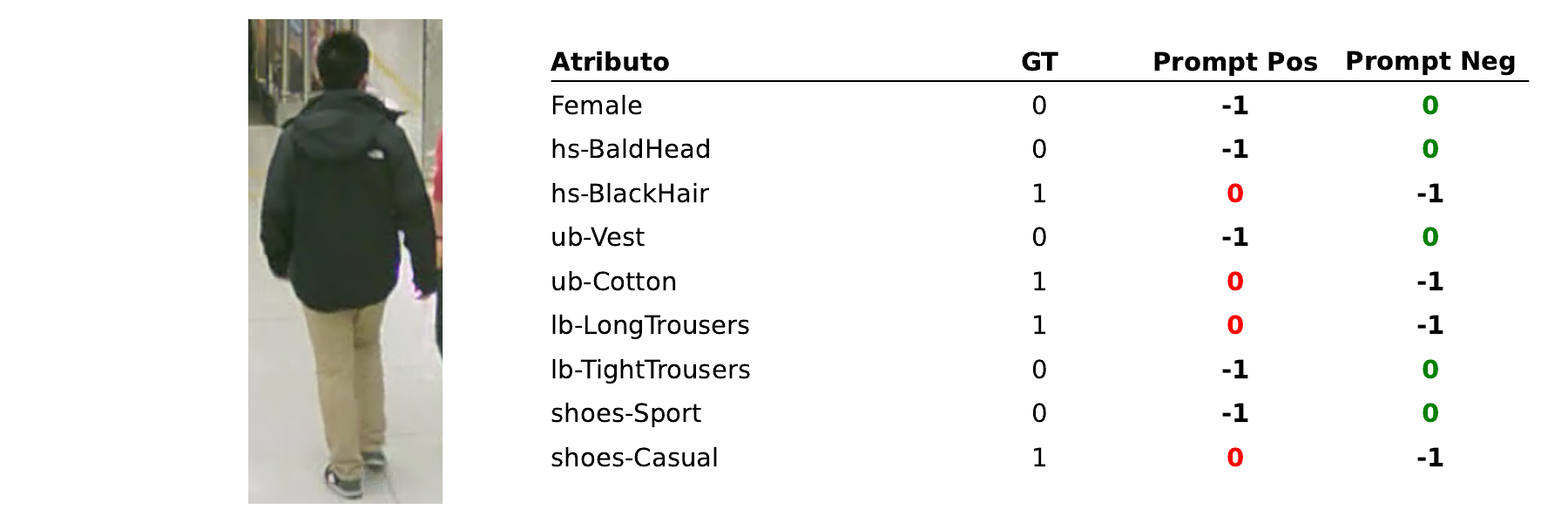}%
  }\hfill
  \subfloat[\footnotesize Failure: High-confidence alignment with the negative prompt]{%
      \includegraphics[trim=0.25cm 0.2cm 0.25cm 0.0cm, clip, width=0.48\linewidth]{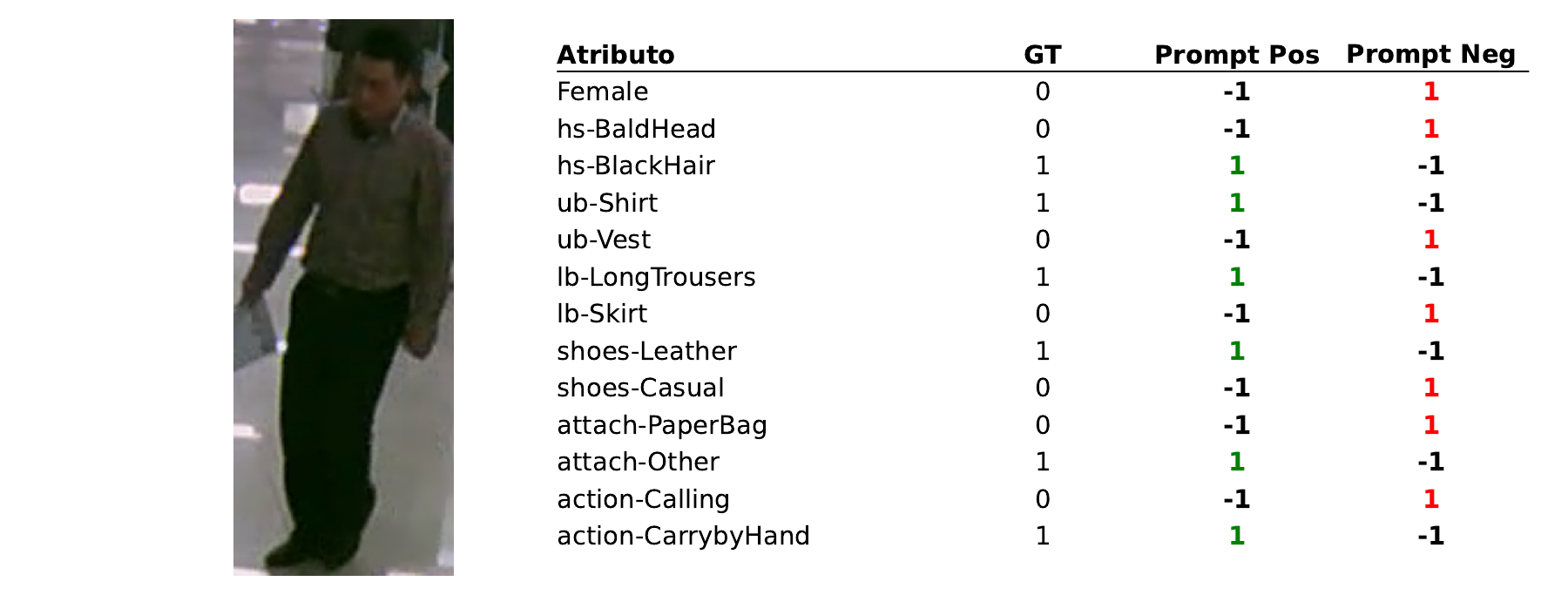}%
  }
  \caption{Qualitative performance analysis of the ReSAGE-PAR Bayesian autolabeling framework on the RAPv1 dataset. We showcase representative success cases (a, b) and failure cases (c, d) to illustrate the filter's decision behavior. For each example, we display the ground truth (GT), alongside the predictions derived from the positive prompt (Prompt Pos) and the negative prompt (Prompt Neg). In both prompt representations, attributes not explicitly present at the prompt are masked as $-1$.}
  \label{fig:qualitative_analysis_rapv1}
\end{figure*}

\begin{figure*}[t]
  \centering
  
 \subfloat[\footnotesize Success: High-confidence alignment with the positive prompt and negative prompt]{%
      \includegraphics[trim=0.25cm 0.2cm 0.25cm 0.0cm, clip, width=0.48\linewidth]{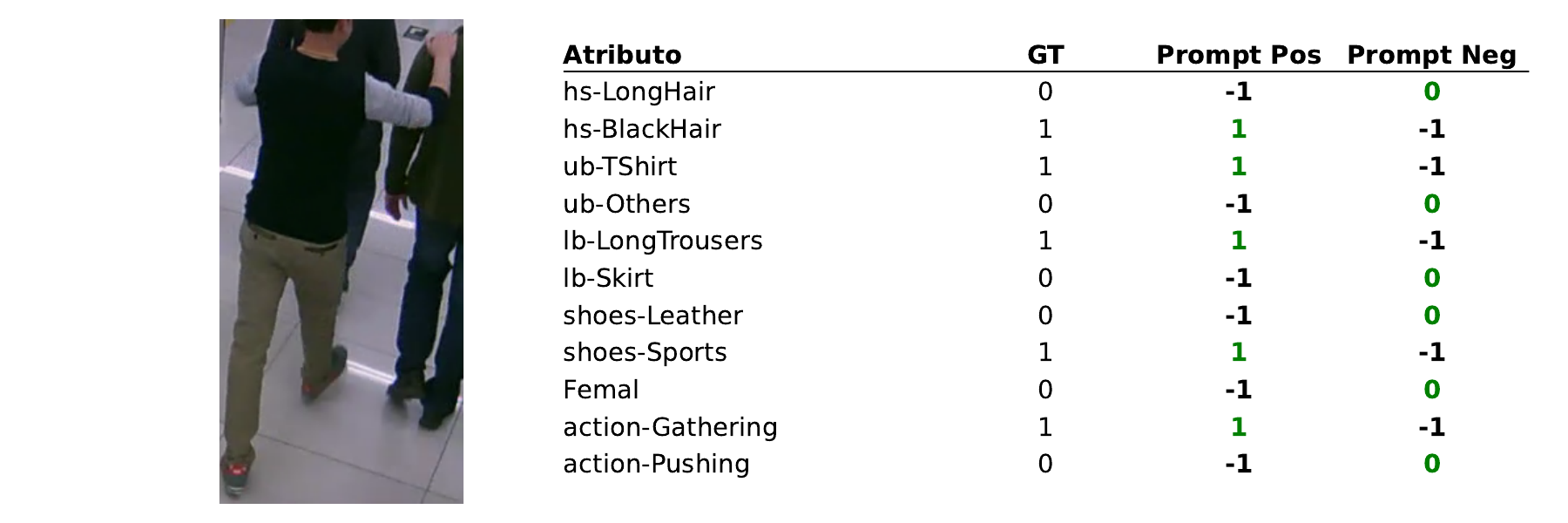}%
  }\hfill
  \subfloat[\footnotesize Success: High-confidence alignment with the positive prompt and negative prompt]{%
      \includegraphics[trim=0.25cm 0.2cm 0.25cm 0.0cm, clip, width=0.48\linewidth]{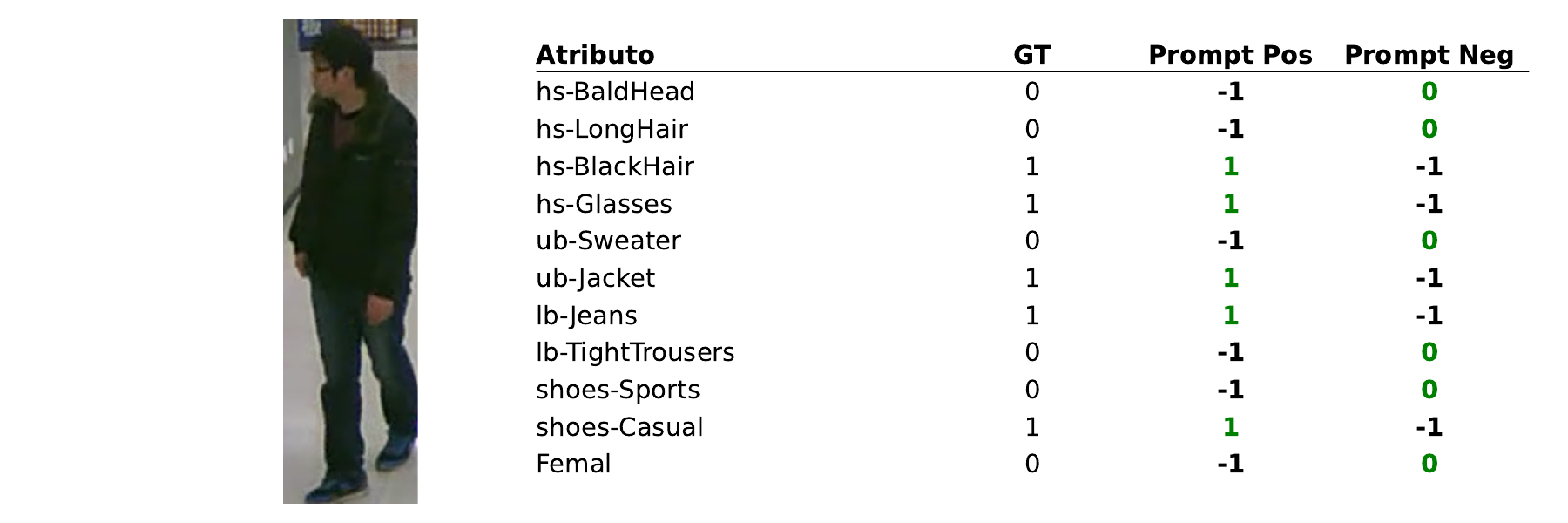}%
  }

    \subfloat[\footnotesize Failure: High-confidence alignment with the positive prompt]{%
      \includegraphics[trim=0.25cm 0.2cm 0.25cm 0.0cm, clip, width=0.48\linewidth]{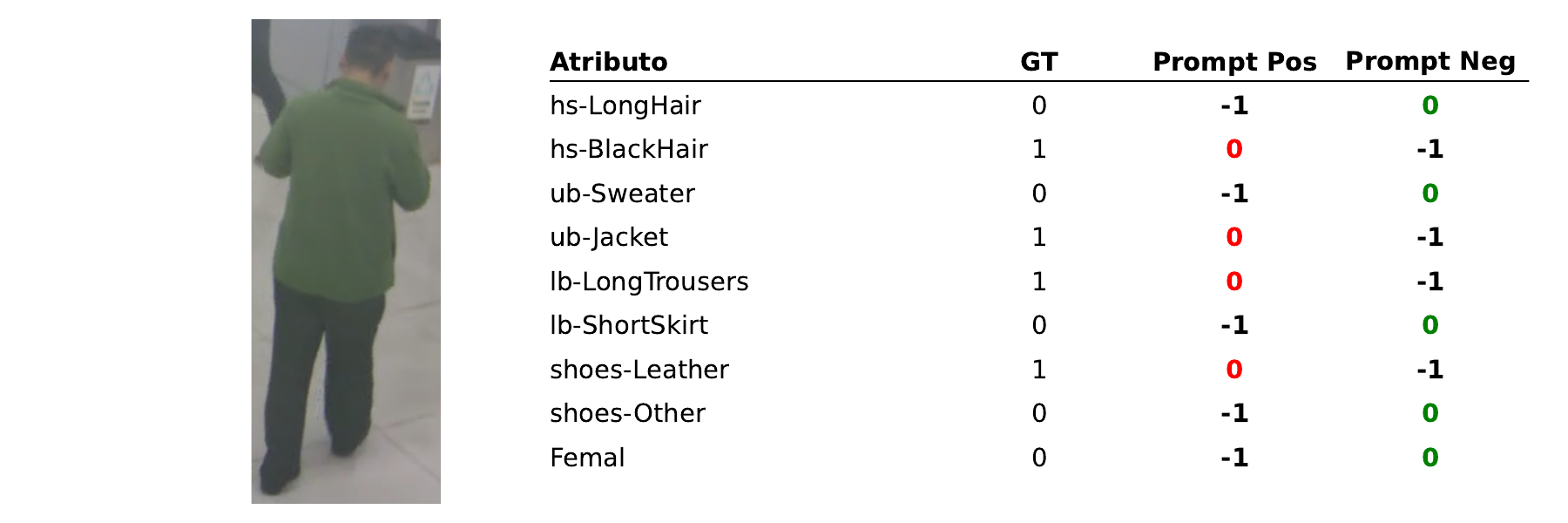}%
  }\hfill
  \subfloat[\footnotesize Failure: High-confidence alignment with the negative prompt]{%
      \includegraphics[trim=0.25cm 0.2cm 0.25cm 0.0cm, clip, width=0.48\linewidth]{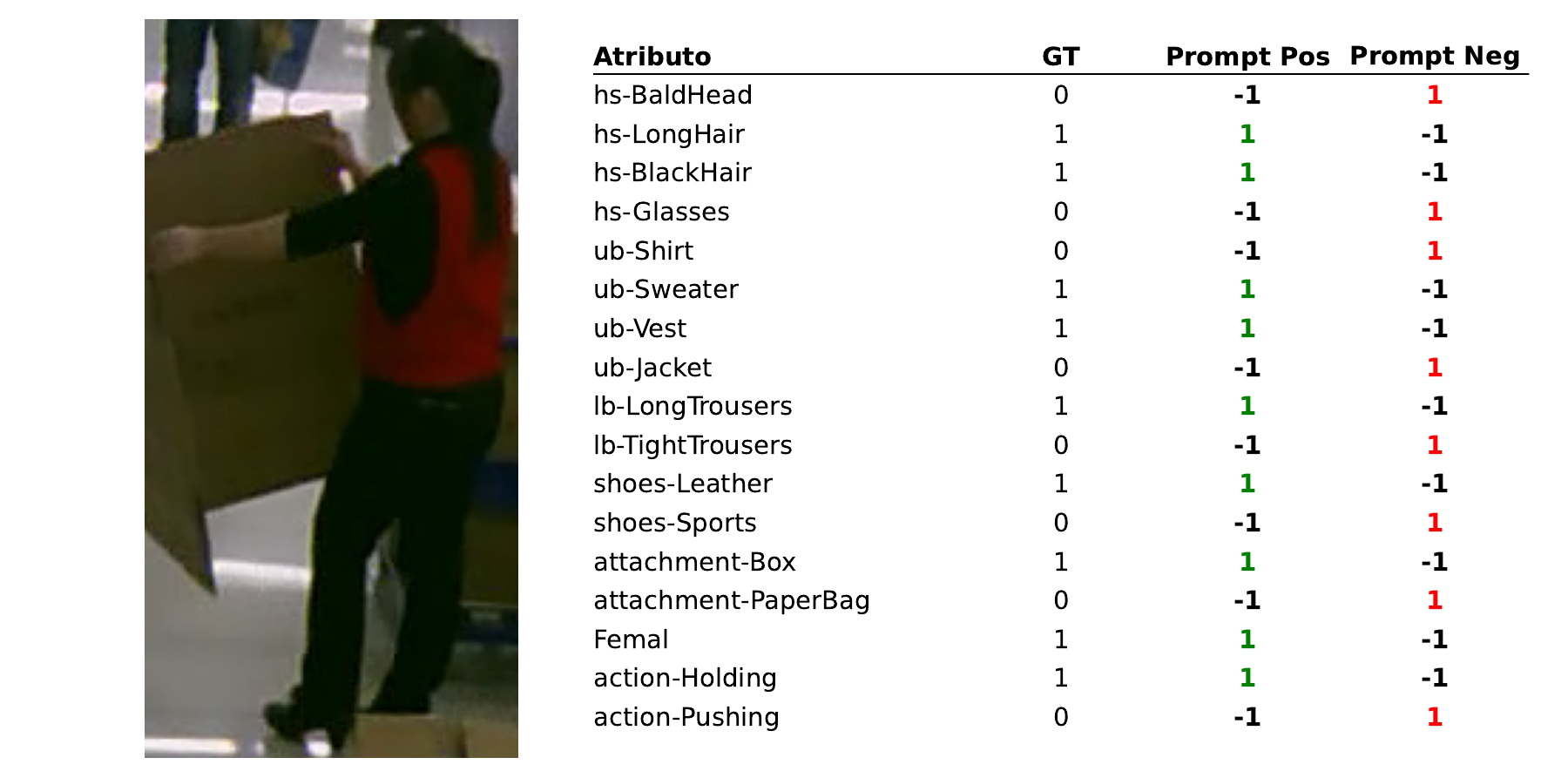}%
  }
  \caption{Qualitative performance analysis of the ReSAGE-PAR Bayesian autolabeling framework on the RAPzs dataset. We showcase representative success cases (a, b) and failure cases (c, d) to illustrate the filter's decision behavior. For each example, we display the ground truth (GT), alongside the predictions derived from the positive prompt (Prompt Pos) and the negative prompt (Prompt Neg). In both prompt representations, attributes not explicitly present at the prompt are masked as $-1$.}
\label{fig:qualitative_analysis_rapzs}
\end{figure*}

\begin{figure*}[t]
  \centering
  
 \subfloat[\footnotesize Success: High-confidence alignment with the positive prompt and negative prompt]{%
      \includegraphics[trim=0.25cm 0.2cm 0.25cm 0.0cm, clip, width=0.48\linewidth]{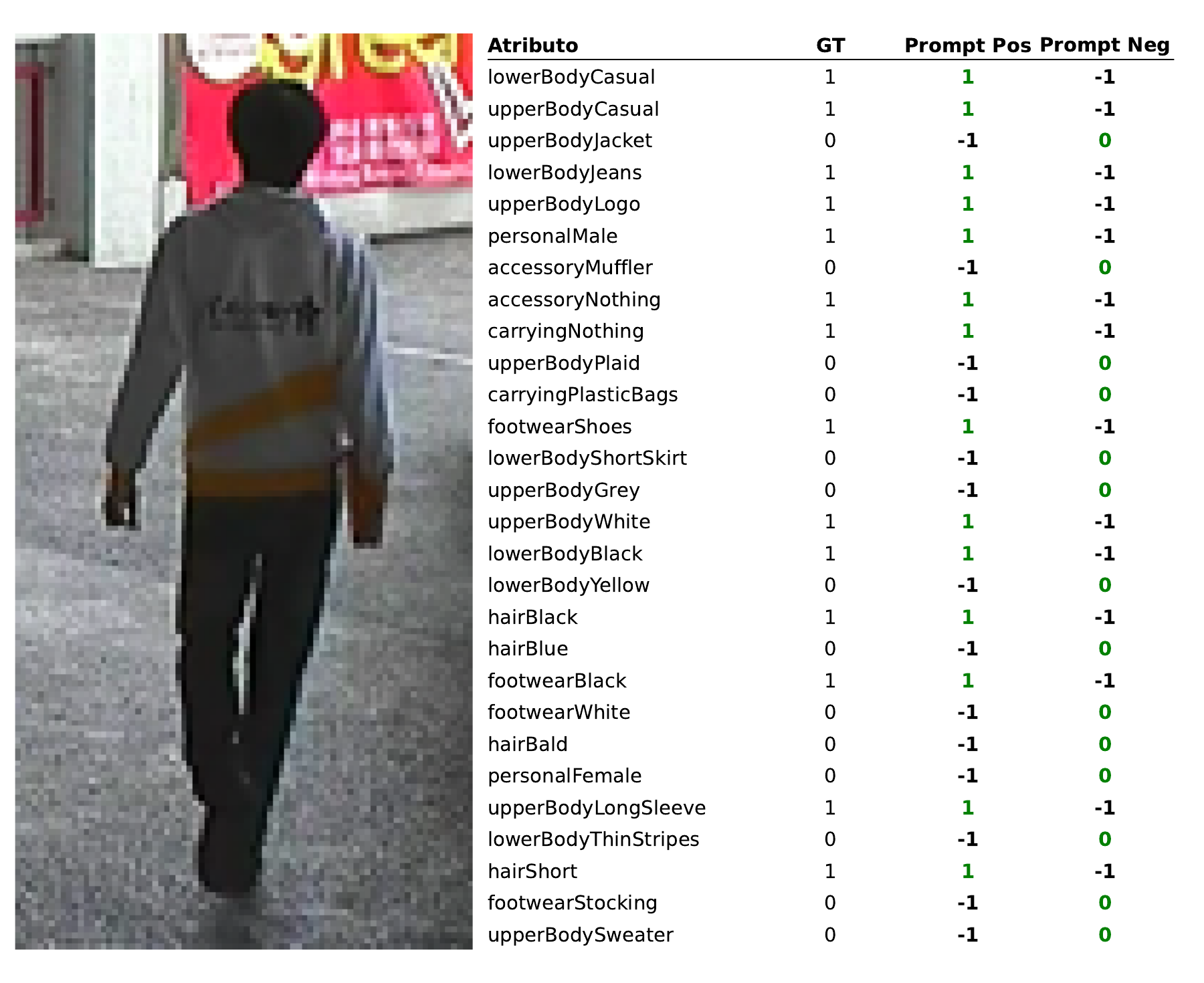}%
  }\hfill
  \subfloat[\footnotesize Success: High-confidence alignment with the positive prompt and negative prompt]{%
      \includegraphics[trim=0.25cm 0.2cm 0.25cm 0.0cm, clip, width=0.48\linewidth]{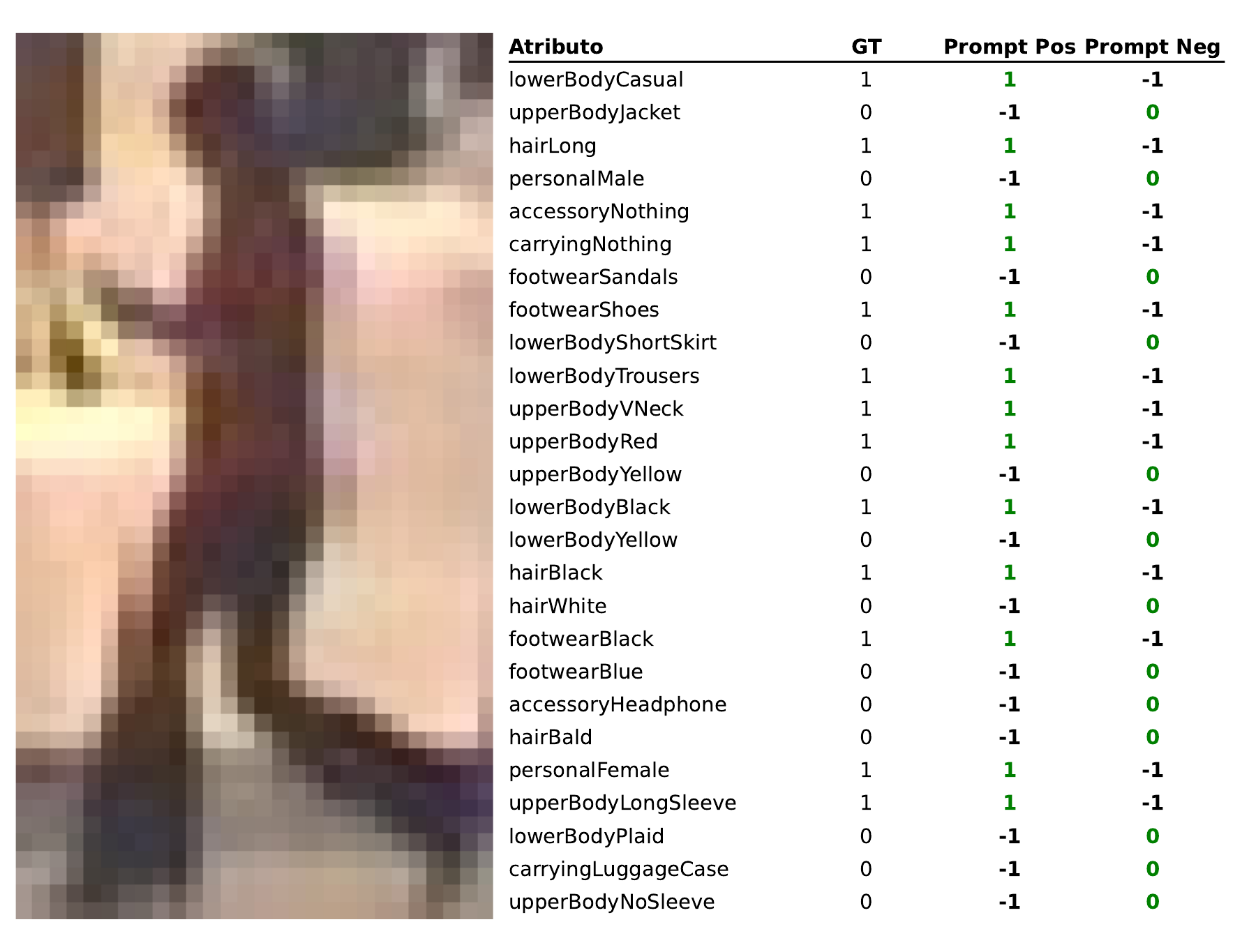}%
  }

    \subfloat[\footnotesize Failure: High-confidence alignment with the positive prompt]{%
      \includegraphics[trim=0.25cm 0.2cm 0.25cm 0.0cm, clip, width=0.48\linewidth]{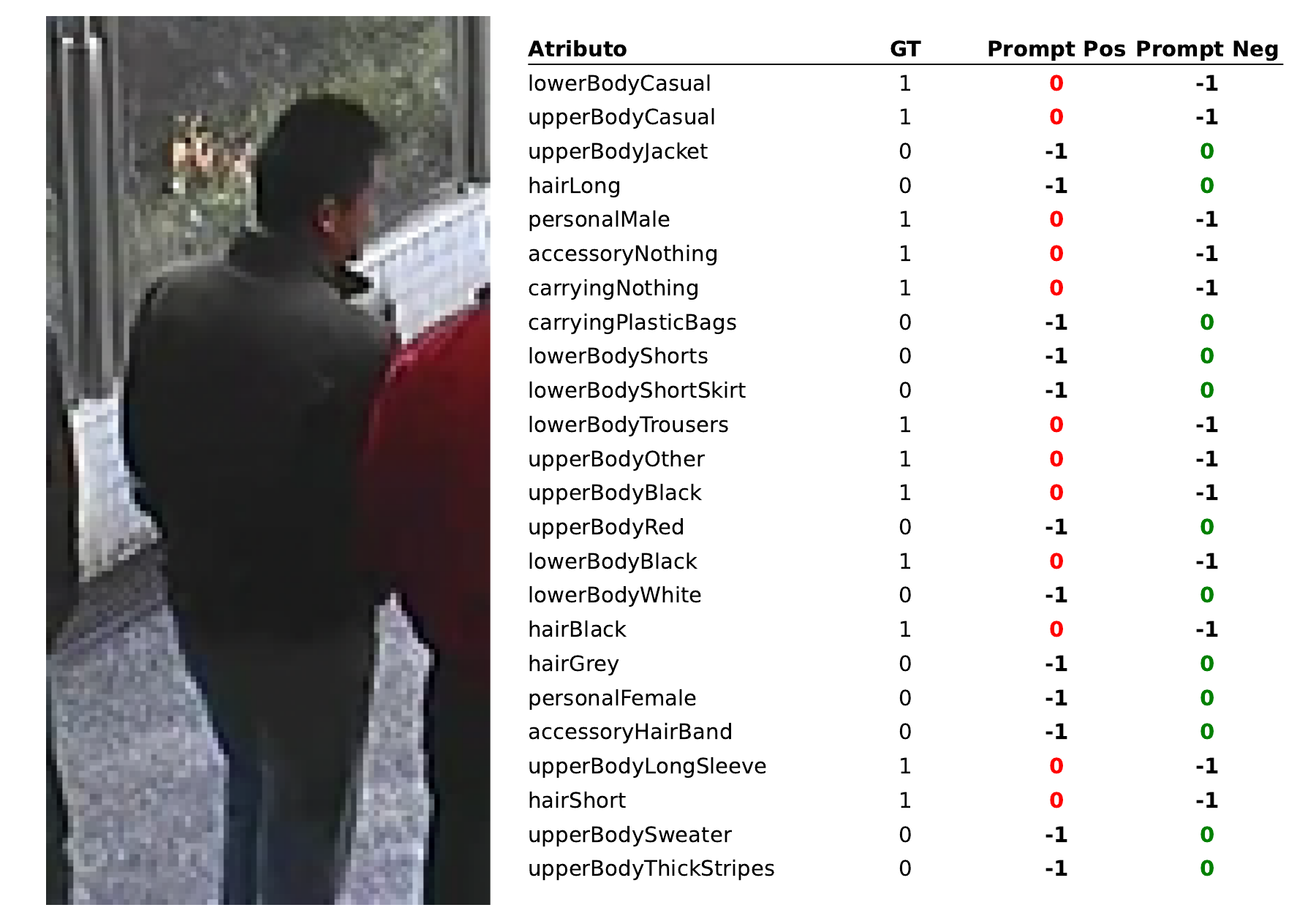}%
  }\hfill
  \subfloat[\footnotesize Failure: High-confidence alignment with the negative prompt]{%
      \includegraphics[trim=0.25cm 0.2cm 0.25cm 0.0cm, clip, width=0.48\linewidth]{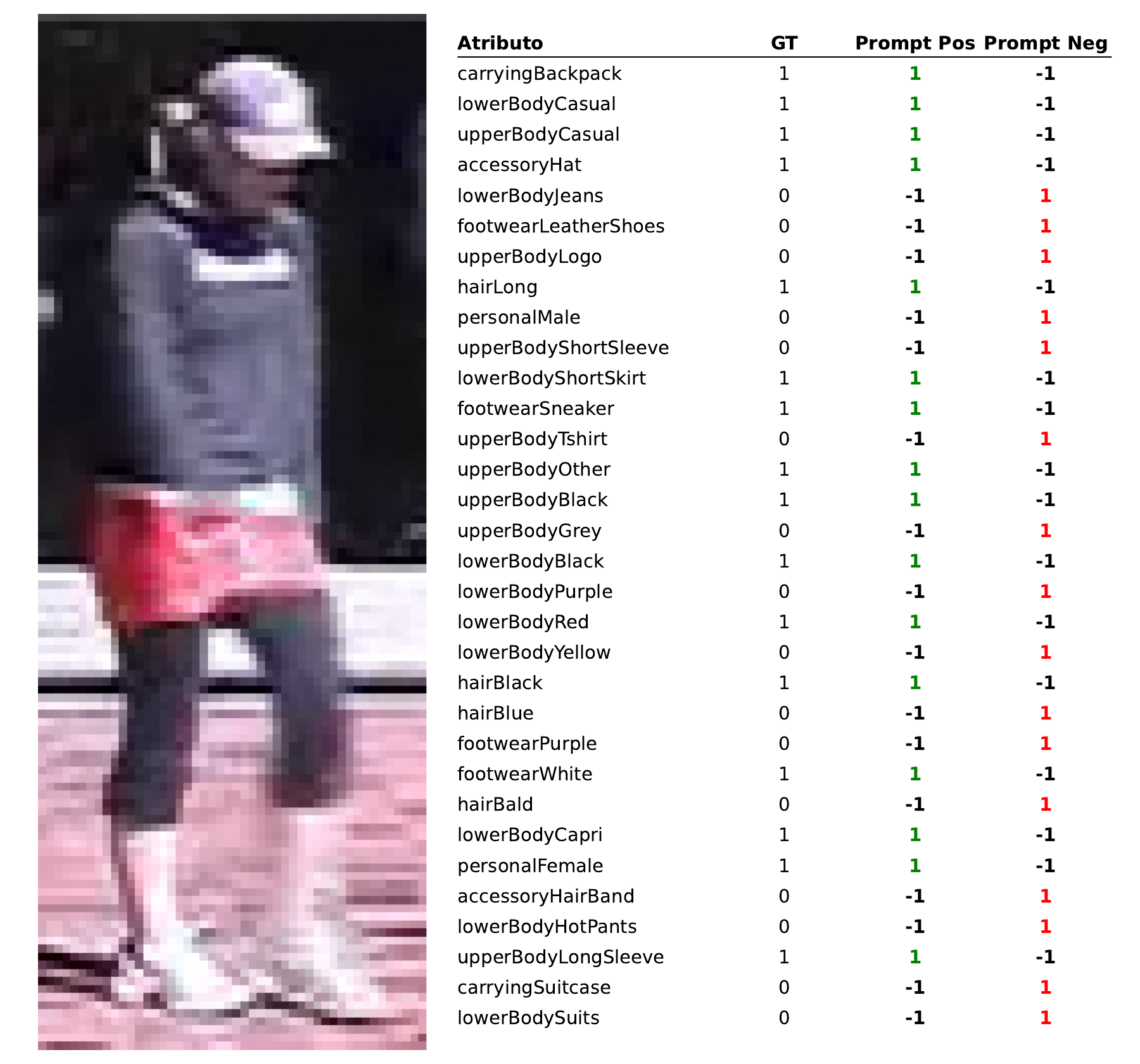}%
  }
  \caption{Qualitative performance analysis of the ReSAGE-PAR Bayesian autolabeling framework on the PETAzs dataset. We showcase representative success cases (a, b) and failure cases (c, d) to illustrate the filter's decision behavior. For each example, we display the ground truth (GT), alongside the predictions derived from the positive prompt (Prompt Pos) and the negative prompt (Prompt Neg). In both prompt representations, attributes not explicitly present at the prompt are masked as $-1$.}
\label{fig:qualitative_analysis_petazs}
\end{figure*}

\begin{figure*}[t]
  \centering
  
 \subfloat[\footnotesize Success: High-confidence alignment with the positive prompt and negative prompt]{%
      \includegraphics[trim=0.25cm 0.2cm 0.25cm 0.0cm, clip, width=0.48\linewidth]{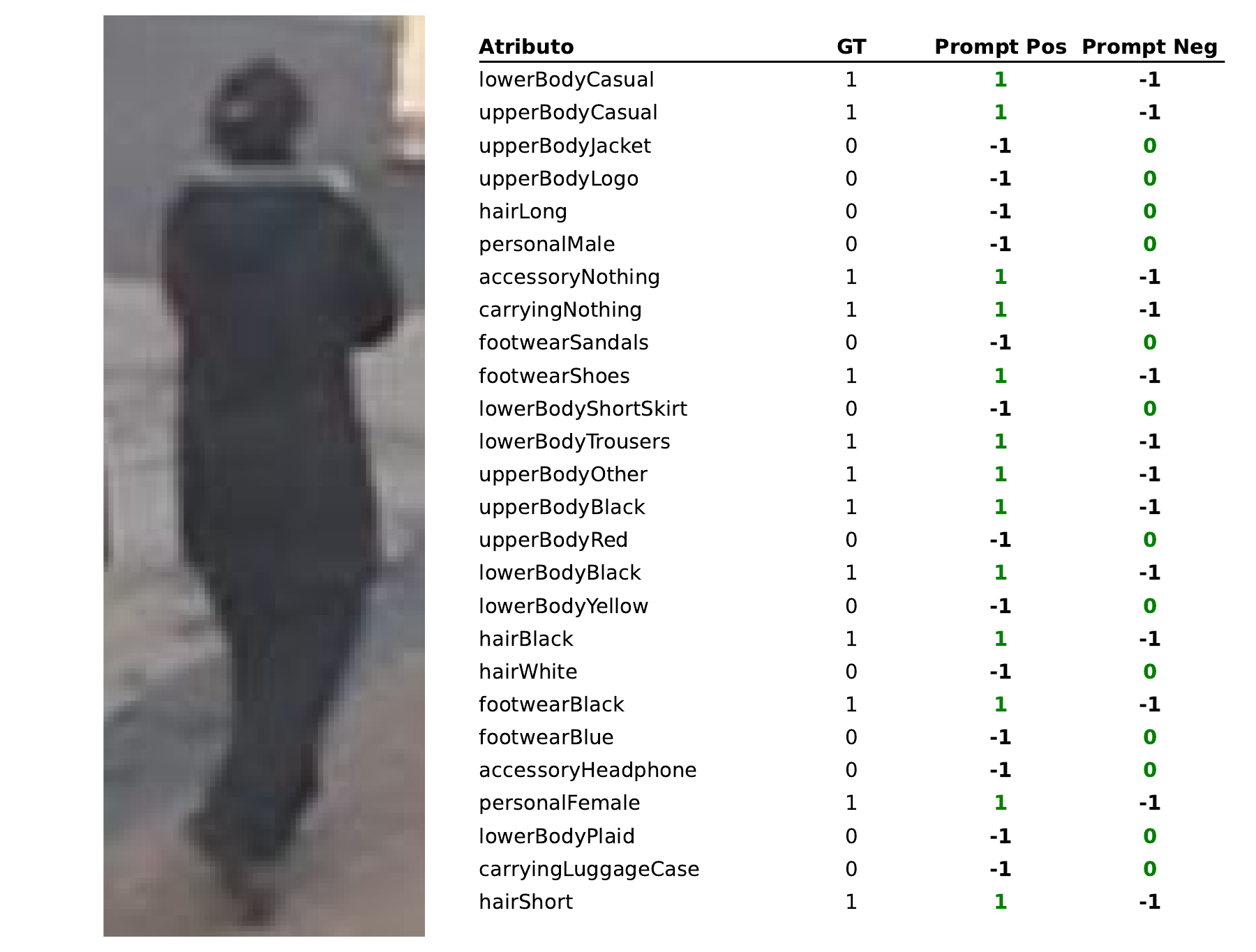}%
  }\hfill
  \subfloat[\footnotesize Success: High-confidence alignment with the positive prompt and negative prompt]{%
      \includegraphics[trim=0.25cm 0.2cm 0.25cm 0.0cm, clip, width=0.48\linewidth]{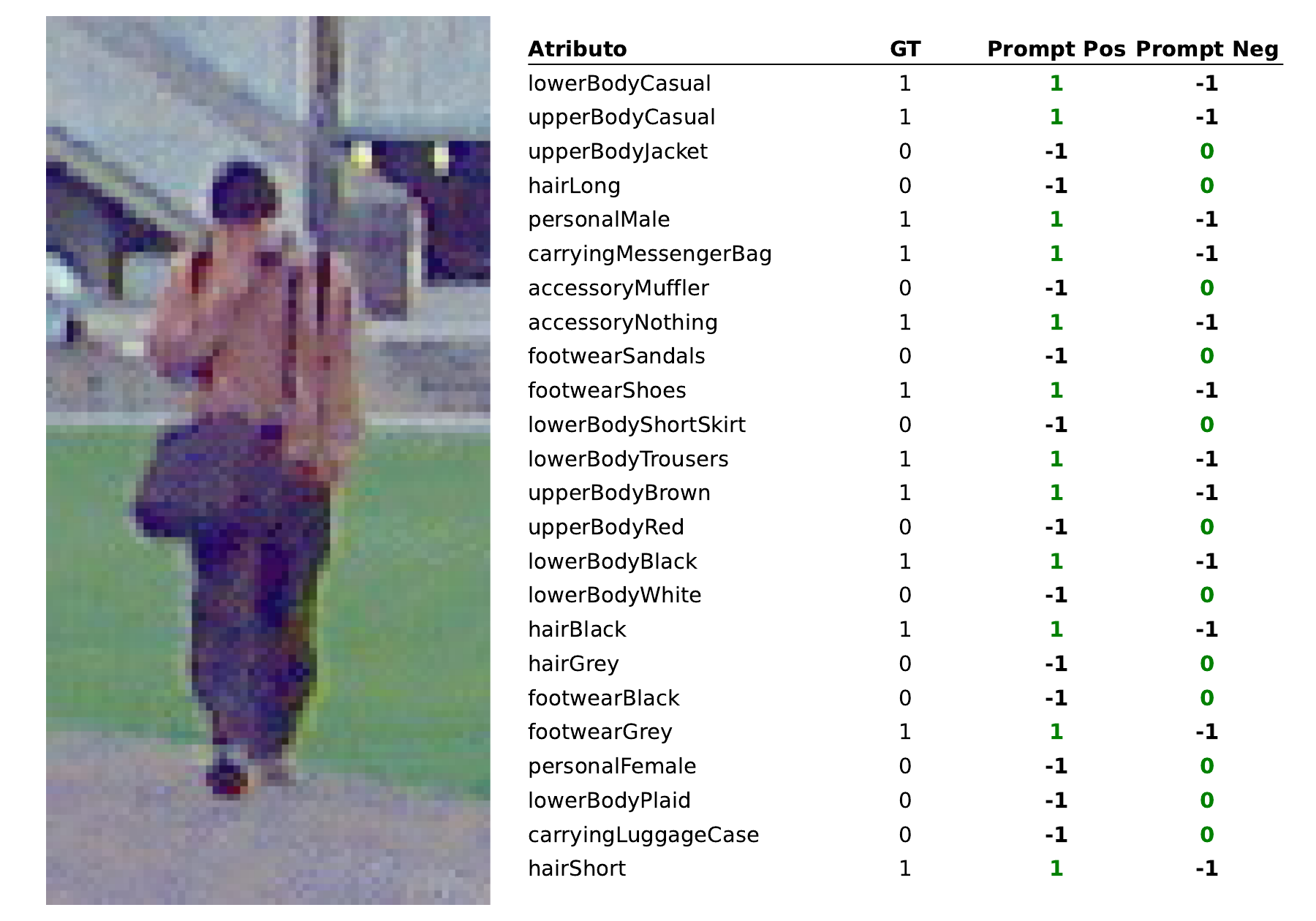}%
  }

    \subfloat[\footnotesize Failure: High-confidence alignment with the positive prompt]{%
      \includegraphics[trim=0.25cm 0.2cm 0.25cm 0.0cm, clip, width=0.48\linewidth]{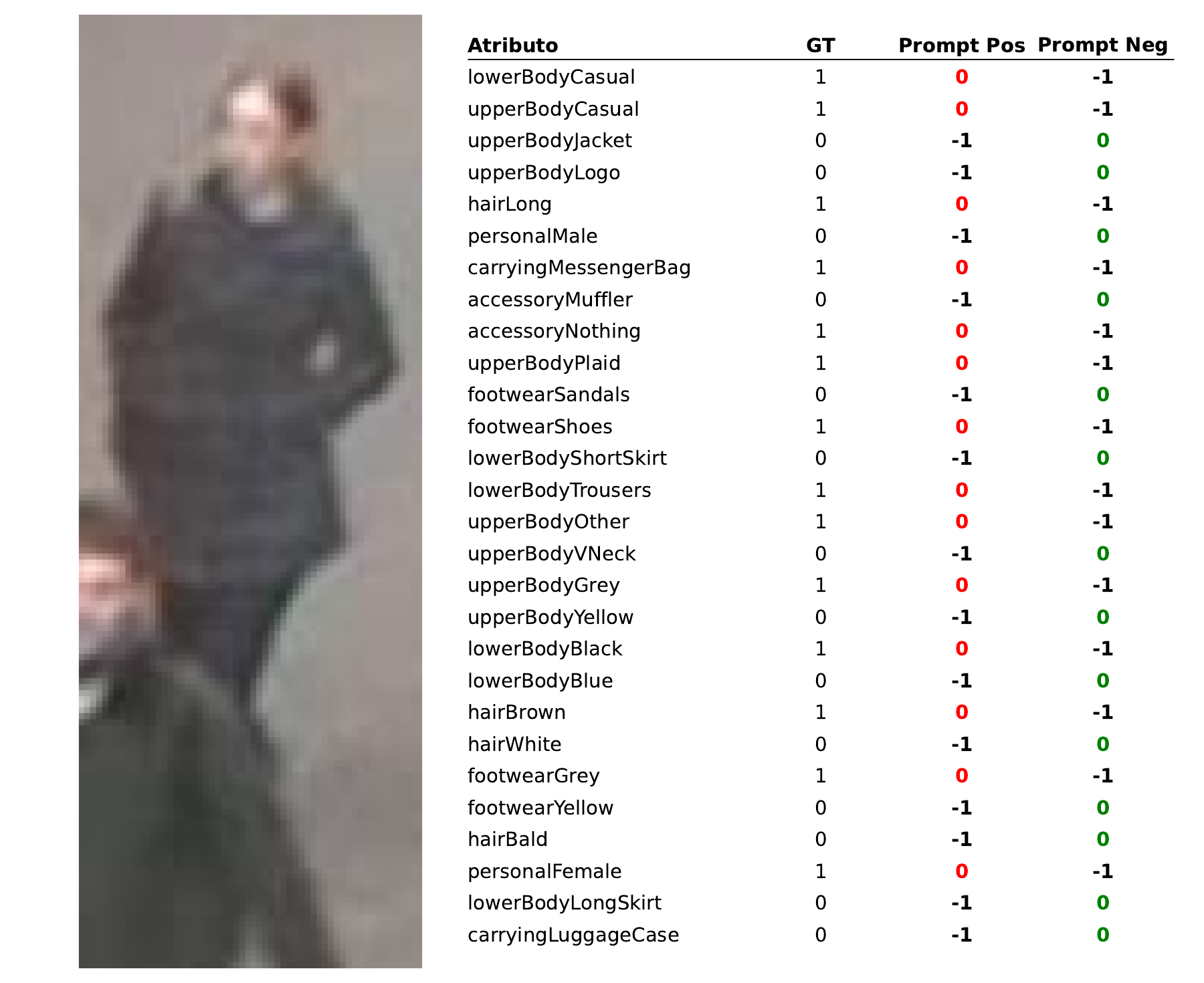}%
  }\hfill
  \subfloat[\footnotesize Failure: High-confidence alignment with the negative prompt]{%
      \includegraphics[trim=0.25cm 0.2cm 0.25cm 0.0cm, clip, width=0.48\linewidth]{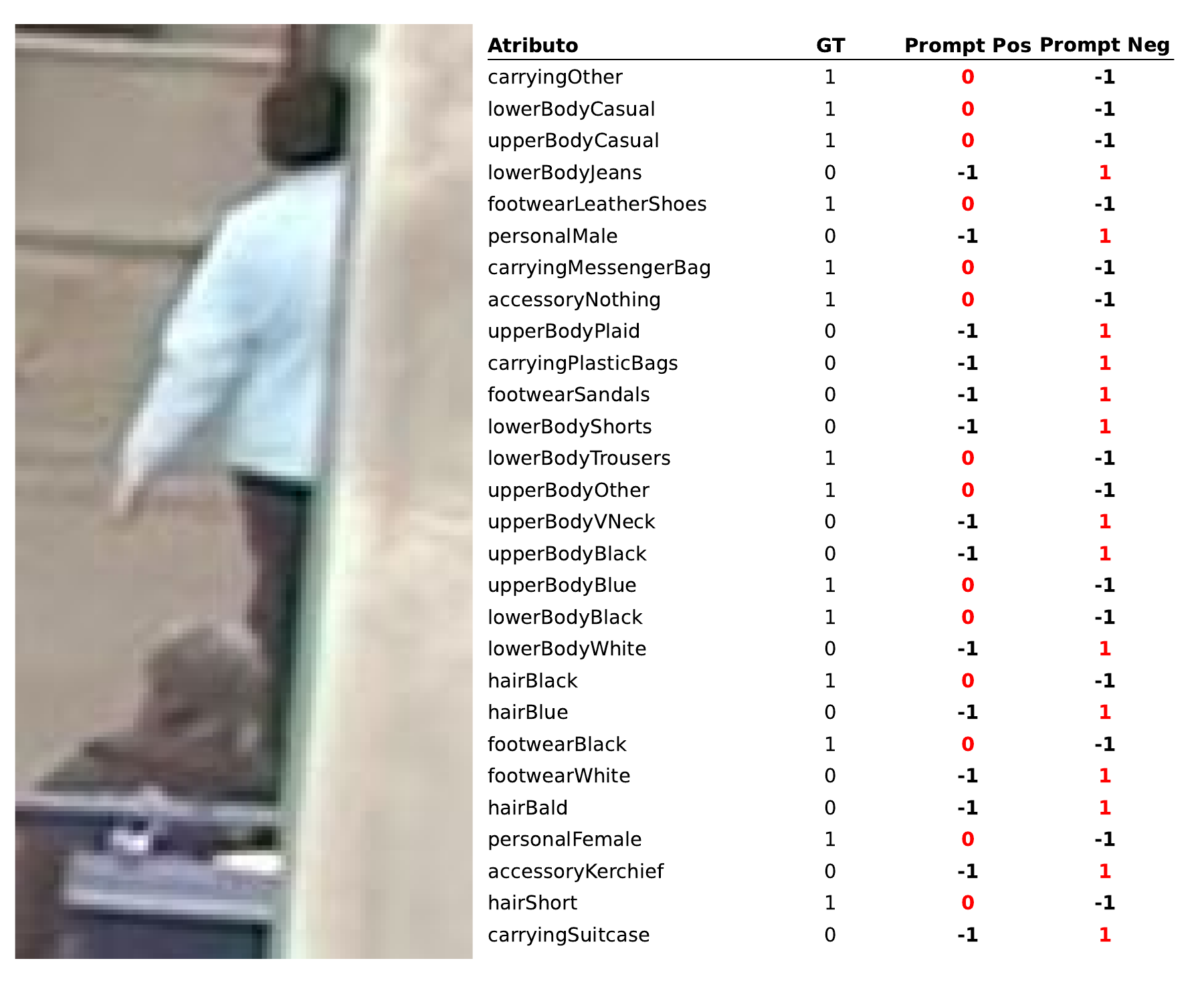}%
  }
  \caption{Qualitative performance analysis of the ReSAGE-PAR Bayesian autolabeling framework on the PETA dataset. We showcase representative success cases (a, b) and failure cases (c, d) to illustrate the filter's decision behavior. For each example, we display the ground truth (GT), alongside the predictions derived from the positive prompt (Prompt Pos) and the negative prompt (Prompt Neg). In both prompt representations, attributes not explicitly present at the prompt are masked as $-1$.}
\label{fig:qualitative_analysis_peta}
\end{figure*}
\clearpage
\clearpage
\section{Extended Threshold Sensitivity Analysis}
\label{sec:anexa_thresholdAnalysis}

As discussed in Section IV-C of the main manuscript, the selection of the decision threshold $\tau$ dictates the trade-off between retaining valid generated attributes and filtering out semantic noise during the autolabeling process. While the primary threshold sensitivity analysis was illustrated using the PETAzs dataset, this section extends the evaluation to the remaining benchmarks. For each dataset, we display the empirical posterior probability distributions $p = P(\text{aligned} \mid s)$ on the testing splits, specifically comparing the density of aligned vs not aligned scores. As shown in \cref{fig:supp_rapv1_thresholds,fig:supp_rapv2_thresholds,fig:supp_rapzs_thresholds,fig:supp_peta_thresholds}, the optimal operating point identified in the main text ($\tau = 0.50$) generalizes exceptionally well across the RAP and PETA families, consistently retaining a high fraction of true positive samples (between 87\% and 94\%) while effectively minimizing the false positive rate. \cref{fig:supp_pa100k_thresholds} illustrates the behavior on PA100K, where the inherent scarcity of annotated attributes slightly increases the classification difficulty, yet $\tau = 0.50$ remains a robust threshold for strict noise filtering. Overall, these extended results corroborate that the default threshold of $\tau = 0.50$ (indicated by the purple line in the density plots) provides a robust, dataset-agnostic baseline for the ReSAGE-PAR framework, yielding high precision (around 90\%) across most benchmarks and a reasonable operational boundary (around 70\%) for the more challenging PA100K dataset.

\vspace{-0.4cm}

\begin{figure}[h]
  \centering
\includegraphics[trim=0.25cm 0.20cm 0.25cm 0.25cm, clip, width=0.8\linewidth]{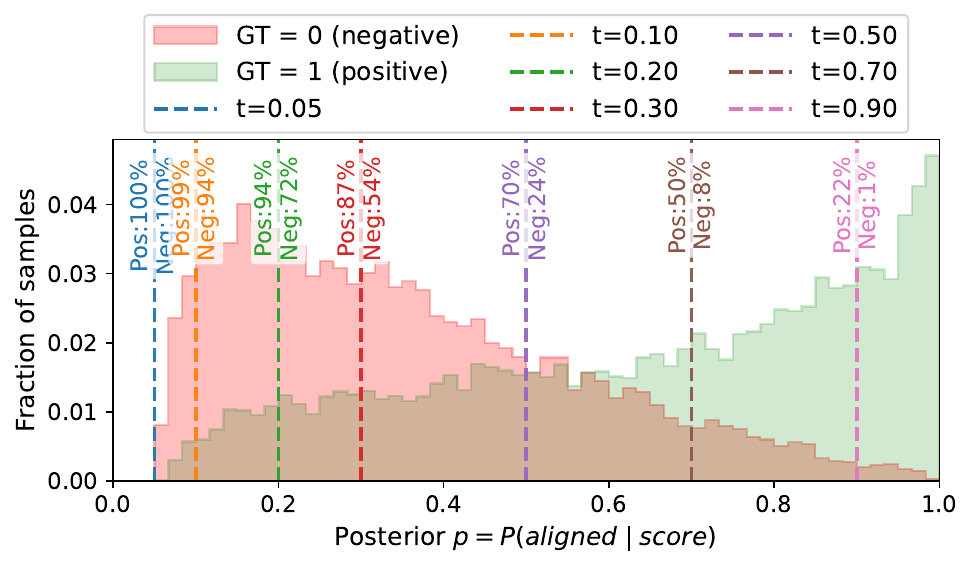}
  \caption{Posterior probabilities $p = P(\text{aligned} \mid s)$ on the PA100K testing split. Compares the density of aligned vs not aligned scores. The vertical line marks the decision threshold $\tau$, showing the retention of valid attributes against the acceptance of complemented ones.}
\label{fig:supp_pa100k_thresholds}

\end{figure}

\vspace{-0.4cm}

\begin{figure}[h]
  \centering
\includegraphics[trim=0.25cm 0.20cm 0.25cm 0.25cm, clip, width=0.8\linewidth]{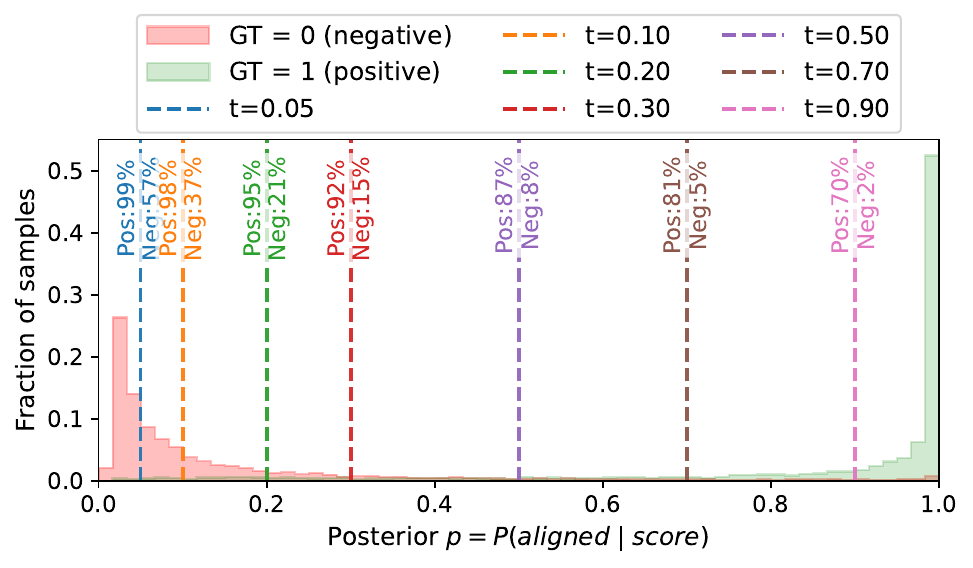}
  \caption{Posterior probabilities $p = P(\text{aligned} \mid s)$ on the RAPv1 testing split. Compares the density of aligned vs not aligned scores. The vertical line marks the decision threshold $\tau$, showing the retention of valid attributes against the acceptance of complemented ones.}
\label{fig:supp_rapv1_thresholds}

\end{figure}

\vspace{-0.4cm}

\begin{figure}[h]
  \centering
\includegraphics[trim=0.25cm 0.20cm 0.25cm 0.25cm, clip, width=0.8\linewidth]{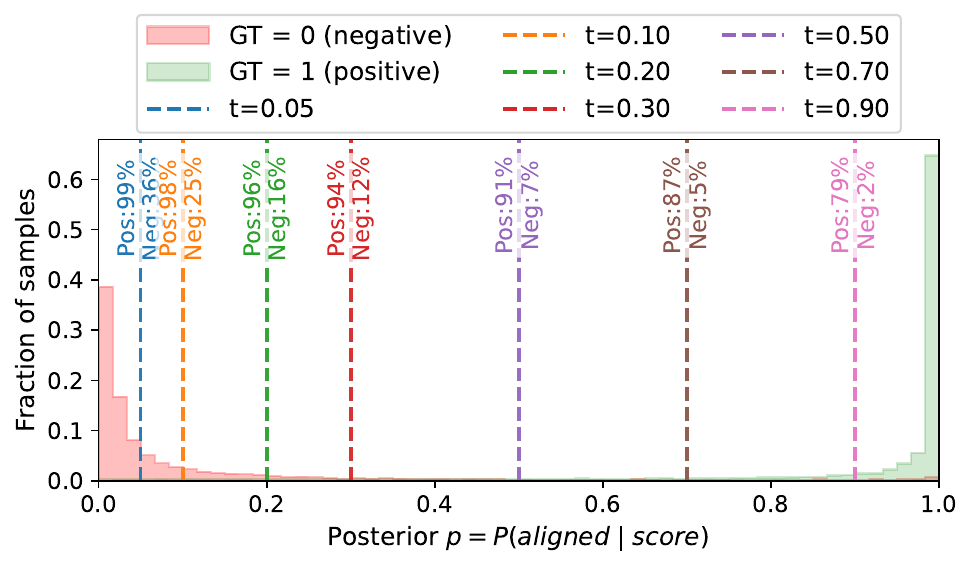}
  \caption{Posterior probabilities $p = P(\text{aligned} \mid s)$ on the RAPv2 testing split. Compares the density of aligned vs not aligned scores. The vertical line marks the decision threshold $\tau$, showing the retention of valid attributes against the acceptance of complemented ones.}
\label{fig:supp_rapv2_thresholds}

\end{figure}

\vspace{-0.4cm}

\begin{figure}[h]
  \centering
\includegraphics[trim=0.25cm 0.20cm 0.25cm 0.25cm, clip, width=0.8\linewidth]{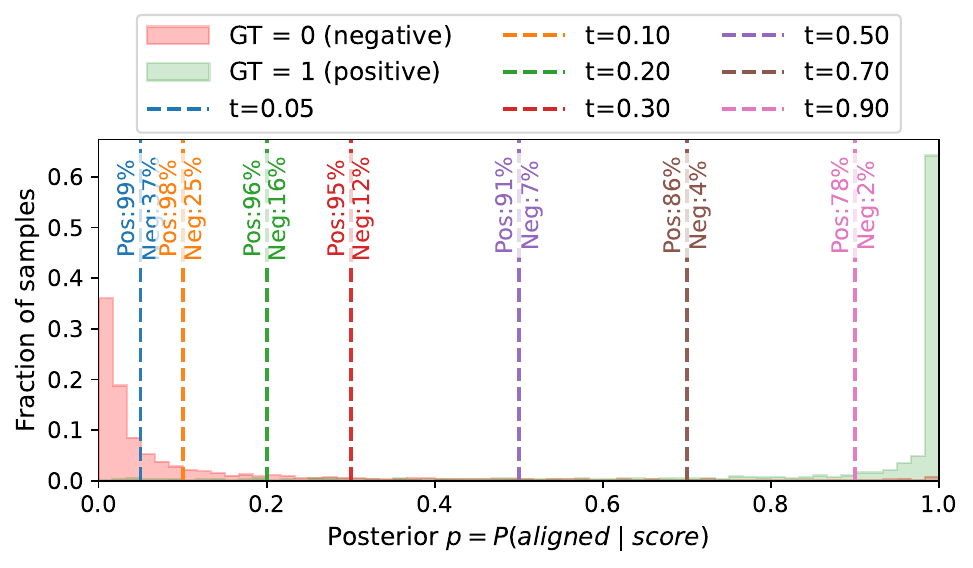}
  \caption{Posterior probabilities $p = P(\text{aligned} \mid s)$ on the RAPzs testing split. Compares the density of aligned vs not aligned scores. The vertical line marks the decision threshold $\tau$, showing the retention of valid attributes against the acceptance of complemented ones.}
\label{fig:supp_rapzs_thresholds}

\end{figure}

\vspace{-0.4cm}

\begin{figure}[h]
  \centering
\includegraphics[trim=0.25cm 0.20cm 0.25cm 0.25cm, clip, width=0.8\linewidth]{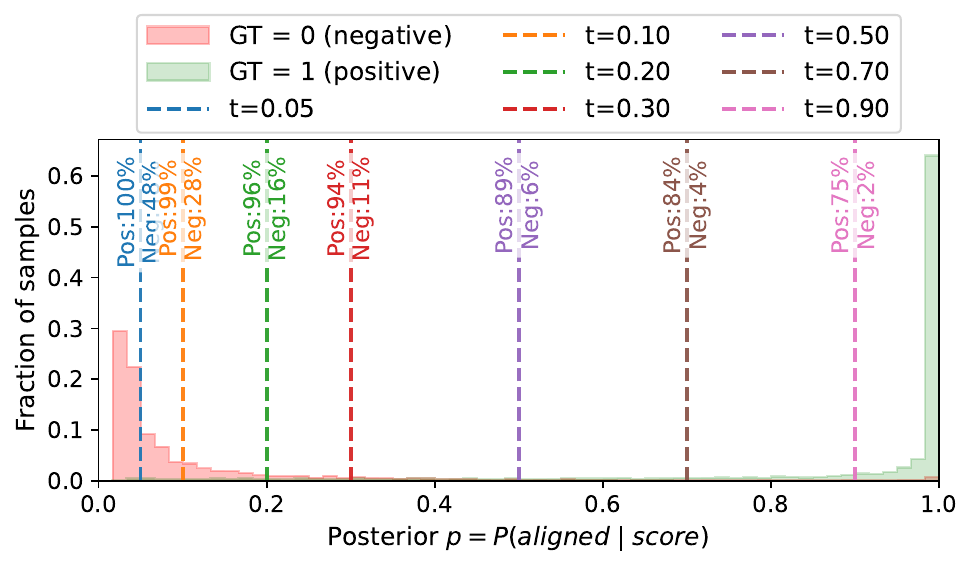}
  \caption{Posterior probabilities $p = P(\text{aligned} \mid s)$ on the PETA testing split. Compares the density of aligned vs not aligned scores. The vertical line marks the decision threshold $\tau$, showing the retention of valid attributes against the acceptance of complemented ones.}
\label{fig:supp_peta_thresholds}

\end{figure}

\clearpage}


\end{document}